\documentclass{article}

\usepackage{arxiv}

\usepackage[utf8]{inputenc} 
\usepackage[T1]{fontenc}    
\usepackage{hyperref}       
\usepackage{url}            
\usepackage{booktabs}       
\usepackage{amsfonts}       
\usepackage{amsmath}
\usepackage{amstext}
\usepackage{enumitem}
\usepackage{caption}
\usepackage{cite}
\usepackage{color}
\usepackage{colortbl}
\usepackage{epsfig}
\usepackage{eurosym} 
\usepackage{emptypage}
\usepackage{fancyhdr}
\usepackage{graphicx}
\usepackage{listings}
\usepackage{multirow}
\usepackage{microtype}      
\usepackage{nicefrac}       
\usepackage[intoc,prefix]{nomencl}
\usepackage{lipsum}
\usepackage{subcaption}
\usepackage{tabularx}
\usepackage{tikz,pgfplots,pgf}
\usetikzlibrary{matrix,shapes,arrows,positioning}
\usepackage{units}
\usepackage{upgreek}
\usepackage{microtype}      

\title{On the application of Physically-Guided Neural Networks with Internal Variables to Continuum Problems}

\author{
Jacobo Ayensa-Jim\'enez \\
  Mechanical Engineering Department;\\
  Arag\'on Institute of Engineering Research (I3A);\\
  University of Zaragoza;\\
  Mariano Esquillor, S/N,  50018 Zaragoza, Spain. \\
  \texttt{jacoboaj@unizar.es} \\
\And
Mohamed H. Doweidar \\
  Mechanical Engineering Department;\\
  Escuela de Ingenier\'ia y Arquitectura (EINA);\\
  University of Zaragoza;\\
  Mar\'ia de Luna S/N, E. Betancourt, 50018 Zaragoza, Spain. \\
  \texttt{mohamed@unizar.es} \\
\And
Jose A. Sanz-Herrera \\
  Mechanical Engineering Department;\\
  School of Engineering;\\
  University of Sevilla;\\
  Camino de los Descubrimientos, S/N, 41092 Sevilla, Spain. \\
  \texttt{jsanz@us.es} \\
\And
Manuel Doblaré\thanks{Corresponding author} \\
  Mechanical Engineering Department;\\
  Arag\'on Institute of Engineering Research (I3A);\\
  University of Zaragoza;\\
  Mariano Esquillor, S/N,  50018 Zaragoza, Spain. \\
  \texttt{mdoblare@unizar.es} \\
}

\pgfplotsset{compat=1.16}
\begin{document}

\newcommand\norm[1]{\left\lVert#1\right\rVert}
\newcommand{\bs}[1]{\boldsymbol{#1}}
\newcommand{\ssf}[1]{\mathsf{#1}}
\tikzset{%
  every neuron/.style={
    circle,
    draw,
    minimum size=0.5cm
  },
  neuron missing/.style={
    draw=none, 
    scale=3,
    text height=0.2cm,
    execute at begin node=\color{black}$\vdots$
  },
}

\maketitle

\tableofcontents

\begin{abstract}
Predictive Physics has been historically based upon the development of mathematical models that describe the evolution of a system under certain external stimuli and constraints. The structure of such mathematical models relies on a set of physical hypotheses that are assumed to be fulfilled by the system within a certain range of environmental conditions.  A new perspective is now raising that uses physical knowledge to inform the data prediction capability of artificial neural networks. 

A particular extension of this data-driven approach is Physically-Guided Neural Networks with Internal Variables (PGNNIV). In this context, universal physical laws are used as constraints in the neural network, in such a way that some neuron values can be interpreted as internal state variables of the system. This endows the network with unraveling capacity, as well as better predictive properties such as faster convergence, fewer data needs and additional noise filtering. Besides, only observable data are used to train the network, and the internal state equations may be extracted as a result of the training processes, so there is no need to make explicit the particular structure of the internal state model, while getting consistent solutions with Physics.

We extend here this new methodology to continuum physical problems, showing again its predictive and explanatory capacities when only using measurable values in the training set. Moreover, we show that the mathematical operators developed for image analysis in deep learning approaches can be used and extended to consider standard functional operators in continuum Physics, thus establishing a common framework for both.

The methodology presented demonstrates its ability to discover the internal constitutive state equation for some problems, including heterogeneous and nonlinear features, while maintaining its predictive ability for the whole dataset coverage, with the cost of a single evaluation.
\end{abstract}


\clearpage

\section{Introduction} \label{secc::introduction}

In the last years, our capacity to collect data has increased at an unprecedented rate \cite{atzori2010, manyika2011}. Today, billions of sensors, transducers or videocameras get data from physical systems, while the advent of the Internet of Things will hugely increase the amount and variety of such data \cite{atzori2010}. This ability is progressively moving us from the parametric regression approach used in model-based Science to non-parametric regression methods, typical of Artificial Intelligence and, in particular, of Artificial Neural Networks (ANN)\cite{maggiora1992computational}, a technique that has had an impressive success in many fields \cite{stulp2015many, hoffmann2019benchmarking, Hill2006, Aneshensel2013, Raghupathi2014, Lopez-Moreno2014, Krizhevsky2012, purwins2019deep}. This new paradigm may be interpreted as a certain return from the well-established hybrid inductive-deductive approach, to the original pure inductive method that only uses direct regressions between input and output, without imposing any additional hypothesis on the mathematical structure of the regression model \cite{minto2018}. However, this move back is being helped by the huge amount and variety of the data available, the power of data preprocessing techniques, especially for non-structured data, the increasing computer power with dedicated capacities \cite{LeCun2019} and a new generation of software tools that simplify and optimize the construction of the networks, the training process, and, finally, the validation of the NN-based model, such as TensorFlow and Keras \cite{gulli2017deep,geron2019hands}, Theano \cite{bergstra2011theano,bastien2012theano} or Pytorch \cite{paszke2017automatic,paszke2019pytorch}.

This latter approach still faces, however, some important drawbacks in its application to physical sciences. For example, scientific data are biased by centuries of knowledge\cite{berry2011, gould1981, kitchin2014big}. Also, spurious correlations between them may be unnoticed by AI methods \cite{kitchin2014, kitchin2014big}, so a blind algorithm without any additional information may lead to wrong predictions. Also, many times, we lack of sufficient data, in terms of quantity, variety and quality, to extract the characteristic features in problems that often involve a large number of variables that interact in a complex manner. As a consequence, we can expect poor extrapolation capacity of such models or, alternatively, overfitting behavior \cite{Xue2019}. Finally, a physically-based model is also useful to get new information by interpreting its structure, parameters, and mathematical properties, being this the reason for the important efforts made to whitening the black-box way of working of current machine-learning predictive algorithms \cite{xu2019explainable}. All these characteristics, with especial emphasis on the latter, have delayed in Physics and Engineering the success that data-driven applications have achieved in other domains.

Despite these current limitations, data-based models, helped by artificial intelligence techniques, have started to gain more and more relevance in predictive Physics.  The term coined for this new  paradigm is Physically-Guided Data Science (PGDS) \cite{ayensa2019unsupervised, karpatne2017theory, raissi2017physics, raissi2019physics, li2019combination}.  A straightforward application of these techniques is dynamic data-driven systems (DDS) \cite{darema2004dynamic, peherstorfer2015dynamic, kirchdoerfer2016data, kutz2016dynamic, ayensa2018new}. The DDS idea is improving the predictive capability of physical models by adding information from the experimental data. As it is well-known, in any physical system, two types of state variables may be identified: i) observable (measurable) ones, that can be obtained directly from physical sensors such as position, temperature or forces; ii) internal non-observable (not directly measurable) variables, that integrate locally other observable magnitudes and depend on the particular internal structure of the system, that is condensed and therefore lost in this integration process. This homogenization procedure is required in any "averaged" theory, established at scales higher than the one of quantum mechanics and first principles. In general, these internal state variables (e.g. stresses, plastic strains, damage, etc.) depend upon the whole time-history of the system, and collect the internal changes in the microstructure. Since they are not directly measurable, the only way of relating them with the former observable ones is by means of physical experiments, sufficiently simplified to allow assuming a certain internal state of the system (e.g. uniform distribution of stresses in the central section of a sample under uniaxial tension). The results obtained are extrapolated to more general conditions by additional assumptions (internal state models). In the DDS approach, the experimental results are used directly, without appealing to that latter extrapolation procedure, thus avoiding using any state model with its inherent assumptions and associated errors. Of course, the counterpart is the need of big amounts of experimental data under sufficiently varied conditions. This is costly and does not avoid the need for simplified experimental designs and the implicit assumptions to approximate in them the values of the internal state variables, since they are not directly measurable.

An opposite perspective is integrating physical knowledge into data science models to inform and improve the data prediction capability of neural networks, that is, to constrain the prediction domain of the standard data model coming from pure data-treatment to fulfill some physical constraints. This idea has been applied to different examples in \cite{karpatne2017physics, raissi2017physics, raissi2019physics, lu2019deepxde, long2019pde, bar2019unsupervised, haghighat2020deep}, while a classification of research topics was enumerated in \cite{karpatne2017theory}. However, in all these works, the physical information was introduced directly as relations between the input and output layers. Only in \cite{raissi2019physics, lu2019deepxde} a first attempt was made to provide the network with some explanatory capacity by adding as output some of the parameters associated with the internal state model. A particular idea in this framework is using the physical universal laws to inform the neural network in such a way that it is then possible to associate some neuron values to internal state variables (observable or not). This idea, named as Physically-Guided Neural Networks with Internal Variables (PGNNIV), was introduced in a previous paper \cite{ayensajimenez2020identification}. In such method, the equations of evolution (physical principles) are treated as constraints between neuron values in the NN, while the network adopts a particular topology imposed by the Physics. Moreover, the internal state equations, that represent the averaged behavior of the internal structure of the system, are directly derived from the NN outcome. This latter characteristic endows the methodology with explanatory capacity. The beauty and power of this idea is that only observable data are used while the internal variables may be predicted by the NN itself, thus relaxing as much as we want the internal state model.

In particular, in continuum Physics (deformable solids and fluid mechanics, electromagnetism, energy and mass transport problems, etc.) the universal physical principles (NN constraints) are written in terms of partial differential equations that are currently computationally solved by means of numerical methods (finite elements, boundary elements, finite volumes, meshless methods and a long etcetera) \cite{ruas2016, larsson2009}. These use a previous discretization step in space, driving to an algebraic, in general non-linear, system, that is then solved by means of standard matrix manipulation. Also, for time-dependent (evolution) problems, another discretization step in time is required to transform the time-continuum problem into a discrete one (alternatively automatic differentiation may be used \cite{raissi2019physics}). This includes the selection of a suitable time integrator (i.e. Euler, Multi-step, Runge-Kutta, among many others) \cite{larsson2009}. In the PGNNIV approach, solving this problem is straightforward by working with the discretized version of the problem, assigning an internal neuron to each nodal variable. 

There are several reasons that supports the interest of using PGNNIV in continuum Physics: on one hand, the multiple problems of interest, in many disciplines, that are expressed in this framework. Secondly, the increasing use of images and videos as the main method for data provision in engineering and healthcare, for example, which is equivalent to having a continuous distribution of data in space, previously discretized (pixels in 2D images and voxels in 3D ones), and also in time (time frames in a video). In third place, and following this trend, in the last years, there has been a tremendous effort in treating images and in developing AI tools to make predictions from such images (e.g. convolutional neural networks - CNN-) \cite{yoo2015deep, rawat2017deep, mccann2017convolutional, anwar2018medical, pujari2018}. Finally, the mathematical operators acting on the image can be extended to consider the standard differential operators in continuum Physics, thus allowing PGNNIV to leverage all current possibilities in image treatment to predict the evolution of a physical system (predictive capacity of PGNNIV) as well as to extract information on its structure (explanatory capacity of PGNNIV). Therefore, the similarity in the mathematical language of image treatment and image-based NN predictions (convolutional filters) and of discretized continuous physical problems (discretized differential operators) is so high that makes it plausible to think of a common framework for both, which would allow to take profit of the tools available in both sides to improve the other.

The objective of this work is, therefore, to extend the PGNNIV methodology to continuum problems, showing its predictive capacity to get the input-output relation in a physical system from a sufficient set of data, as well as its unraveling (explanatory) ability to extract knowledge on the system internal structure. This is always performed considering the constraints imposed by Physics, and using only observable (measurable) variables in the training set (here related to continuous distributions of data values in a spatial and/or time domain). Also, the similarities between PDEs and CNN are highlighted and described in detail.

The structure of the paper is as follows: first, in section \ref{sec::problem}, we introduce the problem to solve and the main objectives. Then, we present in section \ref{sec::methods} the general methodology to study continuum physics under the PGNNIV framework, that is, we reformulate the mathematical foundations of continuum physics in the domain of artificial neural networks. Both the fields and the operators are recast in a standard ANN language, as it is TensorFlow. Next, we present in section \ref{sec::validation} several validation examples where the methodology is fully illustrated and its performance is analyzed when dealing with heterogeneous and nonlinear problems. The predictive and explanatory capacity of the methodology is here revealed. Later, in section \ref{sec::numerical_experiments}, we perform several numerical experiments to demonstrate the performance and the main features of the methodology and how it depends on the training dataset size and noise level and the deep learning structure. Finally, we finish the paper with a discussion and the main conclusions of the work.

\section{Problem statement.} \label{sec::problem}

\paragraph{The physical problem (PP).} Let us consider a certain physical problem defined by a set of (possibly nonlinear) partial differential equations which can usually be split into two main groups, universal physical laws and constitutive equations:

\begin{subequations}\label{eq::eq_fundamental}
\begin{align}
    \mathcal{F}[\bs{u},\bs{v}] &= \bs{f} \label{eq::eq_fundamental1}\\
    \mathcal{H}[\bs{u},\bs{v}] &= \bs{0} \label{eq::eq_fundamental2}
\end{align}
\end{subequations}

Eq. (\ref{eq::eq_fundamental}) must be completed with appropriate initial and/or boundary conditions to make the problem well-posed:

\begin{equation} \label{eq::eq_fundamental_bc}
    \mathcal{G}[\bs{u},\bs{v},\bs{g}] = 0
\end{equation}

where $\bs{u}$ and $\bs{v}$ are unknown tensor fields and $\bs{f}$ and $\bs{g}$ are other tensor fields assumed to be known.

The (also known) functionals $\mathcal{F}$ and $\mathcal{G}$ represent the whole set of universal laws associated with the problem in hands, as well as particular geometrical and environmental constraints, whereas $\mathcal{H}$ represents all (possibly unknown) internal state or constitutive relations of the problem. Splitting of the problem in these two sets of partial differential equation systems also drives to the distinction between two kinds of fields: the essential measurable fields, $\bs{u}$, and the internal state fields, $\bs{v}$, that are particular to each continuum based field theory. In many contexts, the underlying theory is formulated such that Eq. (\ref{eq::eq_fundamental2}) may be expressed in the form $\bs{v} = \mathcal{H}(\bs{u})$. For instance, in solid mechanics, $\mathcal{F}$ encodes mass, momentum and energy conservation equations, while $\mathcal{H}$ expresses the material-dependent constitutive relations referred to stresses and strains (or displacements). In this context, $\bs{u}$ is the displacement field and $\bs{v}$ represents stresses and any other internal variable associated with the constitutive framework established (plastic strains, plastic multipliers, stress rates...). To fix ideas, and in the case of linear elasticity theory, we have:

\begin{subequations}\label{eq::eq_example}
\begin{align}
    \bs \nabla \cdot  \bs{\sigma}  &= \rho\bs{b} \\
    \bs{\varepsilon} &= \frac{1}{2}\left(\bs \nabla \otimes \bs{u} + \bs{u} \otimes \bs \nabla\right) \\
    \bs{\sigma} - \bs{C}:\bs{\varepsilon}&= \bs{0}
\end{align}
\end{subequations}

with boundary conditions:

\begin{subequations}\label{eq::eq_example_bc}
\begin{align}
    \bs{u} &= \bar{\bs{u}}. \quad \mathrm{in} \quad \Gamma_D \\
    \bs{\sigma} \cdot \boldsymbol{n}  &= \bar{\bs{t}}, \quad \mathrm{in} \quad \Gamma_N
\end{align}
\end{subequations}

The two first equations in (\ref{eq::eq_example}) are the equilibrium and kinematics expressions (the former derived from universal laws: the variation of linear and angular momenta, while the latter is a mere definition of the strains in terms of the displacements). On the contrary, the third one is the constitutive relation, which is postulated as a linear relationship between stress and strain. This example may be enriched when considering finite strains and displacements, and nonlinear/inelastic materials\cite{bonet2016nonlinear}, keeping the structure of Eq. (\ref{eq::eq_fundamental}). The general mixed boundary constraints given by (\ref{eq::eq_example_bc}) may be considered as special cases of more general functional operators. For instance $\mathcal{G}[\bs{u}]$ may be defined as $\mathcal{G}[u](\bs{x}) =  \int_\Omega \left((\bs{u}(\bs{y})-\bar{\bs{u}}(\bs{y})\right)\delta(\bs{x}-\bs{y}) \, d\bs{y}, \forall \bs{x} \in \Gamma_D$ and $\mathcal{G}[u](\bs{x}) = \bs{0}, \forall \bs{x} \notin \Gamma_D$. For this particular problem, $\bs{u}$ is the essential measurable field and $\bs{\sigma}$ is an internal field, which identifies with $\bs{v}$ in Eq. (\ref{eq::eq_fundamental}), and  arising from classical field theory. Finally, $\bs{b}$ corresponds to the stimulus $\bs{f}$, and $\bar{\bs{u}}$ and $\bar{\bs{t}}$ are the known values of the respective Dirichlet and Neumann prescribed boundary conditions identified with $\bs{g}$.

To solve numerically the physical problem (\ref{eq::eq_fundamental}) with boundary conditions (\ref{eq::eq_fundamental_bc}), it has to be previously discretized both in space and time by means of one of the many discretization available techniques available \cite{larsson2009}, such as the Finite Difference Method \cite{langtangen1999computational}, Finite Element Method \cite{zienkiewicz1971finite}, or other spectral techniques \cite{boyd2001chebyshev}. Once discretized, the physical problems writes:

\begin{subequations}\label{eq::eq_fudamental_d}
\begin{align}
    \bs{F}(\bs{u},\bs{v}) &= \bs{f} \label{eq::eq_fundamental_d1}\\
    \bs{H}(\bs{u},\bs{v}) &= \bs{0} \label{eq::eq_fundamental_d2}
\end{align}
\end{subequations}

with boundary conditions:

\begin{equation} \label{eq::eq_fudamental_bc_d}
    \bs{G}(\bs{u},\bs{v},\bs{g}) = 0
\end{equation}

where now $\bs{u}$, $\bs{v}$ are unknown vectors of dimension $n$ (number of degrees of freedom of the problem) containing all nodal values, $\bs{f}$ and $\bs{g}$ are known vectors of dimension $n$ and $n'$ (number of prescribed degrees of freedom at the boundary), respectively, and $\bs{F}$, $\bs{G}$ and $\bs{H}$ are corresponding array-valued functions. As commented before, often, we replace Eq. (\ref{eq::eq_fundamental_d2}) by $\bs{v} = \bs{H}(\bs{u})$. Note that in this discretized version of the problem, the continuous position label $\bs{x}$ is replaced by a discrete index $j$. Similarly, the fields $\bs{u}$, $\bs{v}$, $\bs{f}$ and $\bs{g}$ are replaced by their discrete counterparts, $\bs{u}=(u_1,\ldots,u_n)$, $\bs{v}=(u_1,\ldots,u_n)$, $\bs{f}=(f_1,\ldots,f_n)$ and $\bs{g}=(g_1,\ldots,g_{n'})$.

\paragraph{The Data Science problem (DSP).} Alternatively to the physical problem, we can think in a \emph{Data-Science Problem} consisting in a collection of input-output data $\mathcal{D} = \{\bs{x}^i,\bs{y}^i\}_{i=1,\ldots,N}$ and whose goal is to learn the underlying implicit relationship $\bs{y} = \bs{F}(\bs{x})$, or, in other words, to build an estimate $\tilde{\bs{y}}$ from $\bs{x}$. To face this problem, Deep Learning (DL) applied to regression problems arises as one possibility, so it is possible to set-up a DL model relating $\bs{x}$ and $\bs{y}$ that may be expressed as $\tilde{\bs{y}} = \mathsf{Y}(\bs{x})$ (note that the use of sans serif notation indicates that there is a DL model relating these two variables). 

Once the error $\bs{e} = \tilde{\bs{y}} - \bs{y}$ is defined, the construction of the model $\mathsf{Y}$ is performed by solving a minimization problem. We usually define a cost function related to the norm of the error for the whole learning dataset $\mathcal{D}$, for instance $\mathrm{CF} = \sum_{i=1}^N\|\bs{e}^i\|^2$, where $\bs{e}^i = \mathsf{Y}(\bs{x}^i) - \bs{y}^i$.

\section{Methodology} \label{sec::methods}

\subsection{Coupling physics and data science problems}

\paragraph{General recipe.} In order to link both the Physics-based and Data Science problems we build a physically-based
neural network with internal variables (PGNNIV), following the three steps described below. Additional details on the
PGNNIV methodology may be found in \cite{ayensajimenez2020identification}:

\begin{enumerate} 
\item \textbf{Selection of appropriate input and output variables:} Since we consider that they come from any type of data-capturing sensor or device, we require that these variables have to be measurable, so they are a subset of variables $\bs{u}$, $\bs{f}$ and $\bs{g}$ of the problem. Of course, the output variables are always the variables that the engineer or scientist want to predict. For notation purposes, we write $\bs{x} = \bs{I}(\bs{u},\bs{f},\bs{g})$ and $\bs{y}=\bs{O}(\bs{u},\bs{f},\bs{g})$.
\item \textbf{Physical constraints:} The physics of the problem given by Eq. (\ref{eq::eq_fundamental_d1}) is now supplied to the NN as constraints on some prescribed layers (PILs). However, this  equation includes the internal variables $\bs{v}$, so for the problem to make sense, we have to include Eq. (\ref{eq::eq_fundamental_d2}) as an element of the network. For example, by defining $\bs{u} = \bs{H}(\bs{v})$ or $\bs{v} = \bs{H}(\bs{u})$.
\item \textbf{Model relaxation:} Since the interest of this methodology is both \emph{to predict} new values of the variable $\bs{y}$ and \emph{to unravel} the constitutive model $\bs{H}$, this latter is generally only partially known. That is, we may know partly its functional structure, or some of the associated parameters. Therefore, the model is replaced by a subnetwork, for instance, $\bs{v} = \mathsf{H}(\bs{u})$. Some guidelines to the set-up of this $\mathsf{H}$ are given hereafter. 
\end{enumerate}

It is important to note that these three steps have to be balanced and consistent: each model relaxation has to be complemented with the definition of supplementary output variables and/or the addition of physical constraints in order to learn both the output variables and the constitutive model. Classical simulation (by means of the NN) is recovered as a limit case when the model is not relaxed at all (the constitutive relation is assumed to be known) and no learning on its structural parameters is required. In this case, the state model is ``exactly'' imposed as an internal constraint. On the other end, the standard Data Science appears as another particular case of this methodology, when no physical information is introduced to the network, and the constitutive model is considered totally relaxed.

\paragraph{PGNNIV formulation.} Now, all ingredients of the PGNNIV approach have been already set-up: we have a predictive input-output neural network, that we call the Reduced Order Modelling (ROM) network, with appropriate physical constraints acting on some prescribed internal layers (PILs). Renaming all the constraints associated with the known physics of the problem as $\bs{R}$, that is, the whole set of relations in $\bs{F}$ and $\bs{G}$ and, if desired, part of the relations in $\bs{H}$ or any other knowledge on the system, it is possible to write the PGNNIV problem as:

\begin{equation} \label{eq::optimization}
\begin{aligned} 
\bs{y} &= \mathsf{Y}(\bs x); \;\;\; \bs{v} = \mathsf{H}(\bs{u})  \\
& \text{s. t.}  \; \; \; \;  \bs{x} = \bs{I}(\bs u, \bs f, \bs g)  \\
& \quad \quad  \; \;  \bs{y} = \bs{O}(\bs u, \bs f, \bs g)  \\
& \quad \quad  \; \;  \bs{R}(\bs u, \bs v, \bs f, \bs g) = \bs{0} 
\end{aligned}
\end{equation}

or, equivalently, following the standard approach in NN, a cost function is minimized, including now the physical constraints as penalty terms. This is equivalent to consider a physically augmented neural network where the new output variables are identically equal to zero and are related to the internal neuron layers through the predefined relations. If we define  $e = \norm{\bs{e}}$ and $\pi_j = \norm{\bs{R}_j}$, this leads to the minimization of the cost function:

\begin{equation} \label{eq::optimization_penalty1}
    \mathrm{CF} = \frac{1}{N}\sum_{i=1}^N \left((e^i)^2 + \sum_{j=1}^rc_j (\pi^i_j)^2\right)
\end{equation}

where the upper index $i$ indicates that the values of $e$ and $\pi_j$ are associated with the sample $i$ of the dataset $\mathcal{D}$, and $c_j$ are penalty coefficients. Using a standard data science notation, we can write:

\begin{equation} \label{eq::optimization_penalty2}
    \mathrm{CF} = \mathrm{MSE}(e) + \sum_{j=1}^rc_j \mathrm{MSE}(\pi_j)
\end{equation}

Note that the coefficients $c_j$, $j=1,\ldots,r$, are numerical parameters of the optimization procedure, so they are new metaparameters of the neural network. Eq (\ref{eq::optimization_penalty2}) may be written in a more compact form as:

\begin{equation} \label{eq::optimization_penalty3}
    \mathrm{CF} = \sum_{j=0}^rc_j \mathrm{MSE}(\pi_j)
\end{equation}
just by adding a penalty coefficient to the loss term $c_0$ and defining $\pi_0 = e$.

\subsection{Data and field description} \label{secc::data_fields}

In continuum physical problems, a time-dependent tensor field is a point-dependent magnitude $\bs{f} = \bs{f}(\bs{x},t)$ indexed therefore by the point coordinate $\bs{x}$ and the time $t$. In many problems in continuum Physics, we deal with tensor fields that, once discretized, are represented by arrays of appropriate dimension. For example, the time dependent ($l$-covariant, $m$-contravariant) tensor field $\bs{f}$ is represented by the multi-indexed array $\bs{F}$, where $F^{i_1,\ldots,i_l,k_1,k_2,k_3,k_4}_{j_1,\ldots,j_m} = F^{i_1,\ldots,i_l}_{j_1,\ldots,j_m}(x_{k_1},y_{k_2},z_{k_3},t_{k_4})$. Note that we have used the contravariant indexes for referring the spatial coordinates and time (voxels and time frames when referring to a particular image or video).

The TensorFlow framework \cite{abadi2016tensorflow} is particularly suitable for working with data associated with a physical discretized field. Indeed, a tensor field $\bs{f}$ is represented in TensorFlow notation by a multiarray tensor $\mathtt{f}$. When that tensor field varies among samples of a given dataset, the tensor rank is expanded to take this into consideration. For instance, the value of a two dimensional discretized displacement field at a given time $t_k = k \Delta t$, $\bs{f}(\bs{x},t_k)$, for a given sampled value $i$, is represented by $\mathtt{f}[i,\cdot,\cdot,k]$, that is a 2nd-rank tensor, where the symbol ``$\cdot$'' represents the two spatial indexes as a whole.

One main feature of TensorFlow is that it allows working with both data-independent and data-dependent tensors. Data-independent tensors are unalterable over the learning dataset, whereas data-dependent tensors depend on the considered sampled. For instance, in continuum mechanics, the displacement field $\bs{u}$ and the stress field $\bs{\sigma}$ are dependent on the boundary conditions, $\bs{g}$, that may change with the input of each problem. On the contrary, the elastic tensor, even if we consider the material as heterogeneous, so that such tensor is spatially-dependent, is constant for any possible input stimulus (internal forces and boundary conditions). Depending on the selection of the input and output variables and the specific conditions of the problem, the different fields involved are either constant or variable fields. For instance, for the linear elastic problem given by Eqs. (\ref{eq::eq_example}) with boundary conditions (\ref{eq::eq_example_bc}) we have at least two possibilities:

\begin{itemize}
    \item The boundary conditions are fixed and we want to learn the displacement field $\bs{u}$ from the external forces $\bs{b}$. In that case, $\bar{\bs{u}}$, $\bs{n}$, $\bar{\bs{t}}$ and $\bs{C}$ are data-independent tensor fields whereas $\bs{\varepsilon}$, $\bs{\sigma}$ and $\bs{u}$ are data-dependent fields.
    \item The boundary conditions (displacements and/or normal tractions at the boundaries) are taken as input and we want to learn the displacement field $\bs{u}$ when the external forces are known and fixed. In that case, $\bs{b}$, $\bs{n}$ and $\bs{C}$ are data-independent and $\bar{\bs{u}}$, $\bar{\bs{t}}$, $\bs{\varepsilon}$, $\bs{\sigma}$ and $\bs{u}$ are data-dependent fields.
\end{itemize}

\subsection{Operator description}

\subsubsection{General considerations and notations}

The operator $\mathcal{F}$, on a vector field $\bs{u}$, acts, after discretization, as a vectorial function $\bs{F}$. This is directly reframed in the TensorFlow language by defining a function relating two multiarray tensors. Briefly, a functional relationship $\bs{u} = \mathcal{F}(\bs{v})$ is first discretized into a $n$-variables function $\bs{v} = \bs{F}(\bs{u})$, which in turns is expressed in TensorFlow as a tensor relationship $\mathtt{v} = F(\mathtt{u})$. Nonetheless, there are two fundamental observation that has to be mentioned:

\begin{enumerate}
    
    \item First, most of the operators acting in the formulation of the continuum physics are either (i) functions acting over the field values or (ii) linear functional operators. Even more, almost all linear operators involved in the formulation of the continua are local operators. TensorFlow is a framework that seems to have been developed for that purpose, as both cases of operators may be seen as convolution filters. 
    
    The first case, that is an operator such that $v_i = f(u_i)$, may be expanded using convolution filters in the TensorFlow framework into a multilayer perceptron. For example, if $\bs{v}$ is a 2D vector field, then, $v^{i_1,k_1,k_2}$ is a rank 3 tensor:
    
    \begin{equation} \label{eq::operator1}
        \underbrace{\mathtt{v_0}}_{[2,n_x,n_y]} \underbrace{\longrightarrow}_{[2,m_1,n_x,n_y]} \underbrace{\mathtt{v_1}}_{[m_1,n_x,n_y]} \underbrace{\longrightarrow}_{[m_1,m_2,n_x,n_y]} \cdots \underbrace{\longrightarrow}_{[m_{k-1},2,n_x,n_y]} \underbrace{\mathtt{v}_k}_{[2,n_x,n_y]} \\
    \end{equation}
    where $\mathtt{v_0}=\mathtt{u}$ and $\mathtt{v_k}=\mathtt{v}$. $k$ is the number of hidden layers and $m_1,\ldots,m_k$ are the number of neurons at layer $k$. The universal approximation theorem \cite{cybenko1989approximations,hornik1991approximation,lu2017expressive,hanin2017universal} guarantees that every regular enough function $f$ may be approximated by multilayer perceptrons so this approximation makes sense.
    
    The second case, that is local linear operators, may be reframed in the TensorFlow framework using convolution filters of a given size. If $\bs{v} = \bs{F}(\bs{u})$ is a local linear operator relating a rank $k$ mutiarray and a rank $k'$ multiarray, both representing 2D spatial fields:
    
    \begin{equation} \label{eq::operator2}
        \underbrace{\mathtt{u}}_{[\underbrace{2,\ldots,2}_k,n_x,n_y]} \underbrace{\longrightarrow}_{[\underbrace{2,\ldots,2}_k,\underbrace{2,\ldots,2}_{k'},n_x,n_y]} \underbrace{\mathtt{v}}_{[\underbrace{2,\ldots,2}_{k'},n_x',n_y']}
    \end{equation}
    
    It is important to note that, as the considered operator is local, the spatial field $\bs{v}$ is undefined at some values close to the boundaries, so $n_x' \neq n_x$ and $n_y' \neq n_y$.
    
    To summarize, all common operators in physical problems may be framed in terms of artificial network structures using convolutional filters or multilayer perceptrons including proper activation functions, in order to take into account possible nonlinearities.
    
    \item Among the data-independent tensors, we distinguish between constant (non-trainable) and variable (trainable) tensors. Once we have fixed the physical problem and decided which is the input-output relation that has to be learned, the selection of the role of each operator tensor is natural: when a tensor is involved in a known operator, such as the ones related with $\bs{F}$ and $\bs{G}$ functions, it is a TensorFlow constant tensor and is denoted with a star. One particular example is the tensor associated with the derivation operator, $\mathtt{D}^*$. If the tensor is associated with an unknown relationship, such as the ROM network $\mathsf{Y}$ or the model network $\mathsf{H}$, the tensor is a variable tensor. An example is the (possibly heterogeneous) elastic tensor $\mathtt{C}$. 
\end{enumerate}

\subsubsection{Brief taxonomy of operators}

To illustrate the introduced concepts above, we particularize the general ideas above to a brief taxonomy of different operators found in continuum physical problems. This discussion is not intended to be exhaustive and complete, but showing on the suitability of the PGNNIV formulation to handle a very wide range of operators.

\paragraph{Common linear differential operators.}

With the presented framework, all differential operators can be cast as pre-defined filters acting on field tensors. A (discretized) differential operator is a function $\bs{D}$ transforming one multiarray tensor into another. For linear differential operators, $\mathcal{D}$ are linear. Consequently, they are encoded as known constant tensors $\mathtt{D}^*$. To fix ideas, let us consider the equilibrium equation in solid mechanics (infinitesimal theory):

\begin{equation} \label{eq::equilibrium_intrinsic}
    \boldsymbol \nabla \cdot \sigma = \rho \boldsymbol{b} 
\end{equation}

Eq. (\ref{eq::equilibrium_intrinsic}) involves covariant derivative, as $\mathrm{div}$ is a linear differential operator defined for a two contravariant tensor. If $\boldsymbol{f}=\mathrm{div}(\boldsymbol{X})$, $f^i=X^{ij}_{|j}$, where $|$ represents the covariant derivative. In a coordinate representation, the covariant derivative is expressed for the considered tensor as $X^{ij}_{|j}= X^{ij}_{,j}  +\Gamma^i_{jk}X^{kj} + \Gamma^j_{jk}X^{ik}$ where $\Gamma^k_{ij}$ are the Christoffel symbols that, for the Levi-Civita connection, are defined in terms of the metric tensor $\bs{\mathfrak{g}}$, and satisfies the following linear equation $\mathfrak{g}_{kl}\Gamma^k_{ij} = \frac{1}{2}\left(\mathfrak{g}_{jl,i}+ \mathfrak{g}_{li,j}- \mathfrak{g}_{ij,l}\right)$. Now, as $\bs{\mathfrak{g}}$ encodes the geometry of the problem, the only ingredient to reframe Eq. (\ref{eq::equilibrium_intrinsic}) to the multiarray framework is to select a discretization of the common, one dimensional, derivative operator $\partial_i$ as a tensor operator. For instance, let us consider a two-dimensional problem. $\partial_1$, may be defined using first-order finite difference approximation as:

\begin{equation} \label{eq::partial_i}
    \left[\partial_1 F\right]^{\bullet,kl}_\bullet = \frac{1}{l} F^{\bullet,k+1\, l}_\bullet -\frac{1}{l} F^{\bullet,k\, l}_\bullet
\end{equation}
so the tensorial expression of $\partial_1$ is $\left[\partial_i\right]_{mn}^{rs}= -1/l$ if $r=m,s=n$, $\left[\partial_i\right]_{mn}^{rs}= 1/l$ if $r=m+1,s=n$ and $\left[\partial_i\right]_{mn}^{rs}= 0$ otherwise. Note that $\partial_i$  is a convolutional filter, given in a planar 2D representation by the kernel stencil (moving from left to right and from bottom to top):

\begin{equation}
    D_1 = \begin{bmatrix}
    0 & 0 & 0 \\
    0 & -1/l & 1/l \\
    0 & 0 & 0
    \end{bmatrix}
\end{equation}

Analogously, the tensorial representation of $\partial_2$ is $\left[\partial_i\right]_{mn}^{rs}= -1/l$ if $r=m,s=n$, $\left[\partial_i\right]_{mn}^{rs}= 1/l$ if $r=m,s=n+1$ and $\left[\partial_i\right]_{mn}^{rs}= 0$ otherwise and its planar representation is:

\begin{equation}
    D_2 = \begin{bmatrix}
    0 & 1/l & 0 \\
    0 & -1/l & 0 \\
    0 & 0 & 0
    \end{bmatrix}
\end{equation}

If we go back to the equilibrium equation, the tensor $\bs \sigma$ associated with the stress field is represented as $\sigma^{ij,kl}$ and the tensor $\bs{b}$ associated with the external forces field per unit mass is represented as $b^{i,kl}$ so the divergence operator may be expressed as a tensor $\mathtt{D}^*$ so that:

\begin{equation} \label{eq::div_components}
    \rho b^{i,rs} = D^{rs*}_{j,kl}\sigma^{ij,kl}
\end{equation}

In particular, let us consider that there is no curvature, $\Gamma^k_{ij}=0$. The non zero components of the tensor are $D^{\bullet,11}_{1,11}=D^{\bullet,11}_{2,11}=-1/l$  and $D^{\bullet,11}_{1,12}=D^{\bullet,11}_{2,12}=1/l$ for a two-by-two grid.

Recall that all differential operators may be reframed as convolutional filters in the spatial slots. This has important consequences from a practical point of view:

\begin{itemize}
\item Operator tensors are sparse in the discretization dimensions (that are those of greater dimensionality). This allows sparse-based algebra and storage, resulting in high performance computations and less demanding requirements.
\item Differential operators may be easily built and used in standard neural network software codes and tools, such as TensorFlow, although some care must be taken in indexing.
\end{itemize}

In summary, all differential operators involved in the fundamental balance equations in Continuum Physics (universal laws) may be encapsulated in this tensor framework, provided we have established two main ingredients: the space geometry ($\bs{\mathfrak{g}}$) and a given discretization rule for differentiation ($\delta$).

\paragraph{Constitutive models.}

Constitutive models (or internal state equations) define the internal state (in general, non-measurable) variables of the problem in terms of the essential (measurable) ones \cite{ayensajimenez2020identification}. They can be written in a general case as:

\begin{equation} \label{eq::const_1}
    \bs{v}= \mathcal{H}(\bs{u})
\end{equation} 
where $\bs{v}$ is the set of internal variable fields and $\bs{u}$ is the set of essential variable fields (for instance, stresses and displacements in continuum mechanics, macroscopic -$\boldsymbol{D},\boldsymbol{H}$- and microscopic -$\boldsymbol{E},\boldsymbol{B}$- intensity of electromagnetic fields in electromagnetism...) and $\mathcal{H}$ must be interpreted as a functional (e.g. $\bs{\sigma} = \bs{H}(\bs{u}) = \frac{1}{2}\bs{C}:\left(\bs{\nabla} \otimes \bs{u} + \bs{u} \otimes \bs{\nabla}\right)$ in linear elasticity).

Once discretized, Eq. (\ref{eq::const_1}) is expressed as:
 
 \begin{equation} \label{eq::const_2}
    \bs{v} = \bs{H}(\bs{u})
\end{equation} 
where now  $\bs{v}$ and $\bs{u}$ are the tensor fields associated to the nodal field values and $\bs{H}$ is a (in general nonlinear) mapping between tensors.

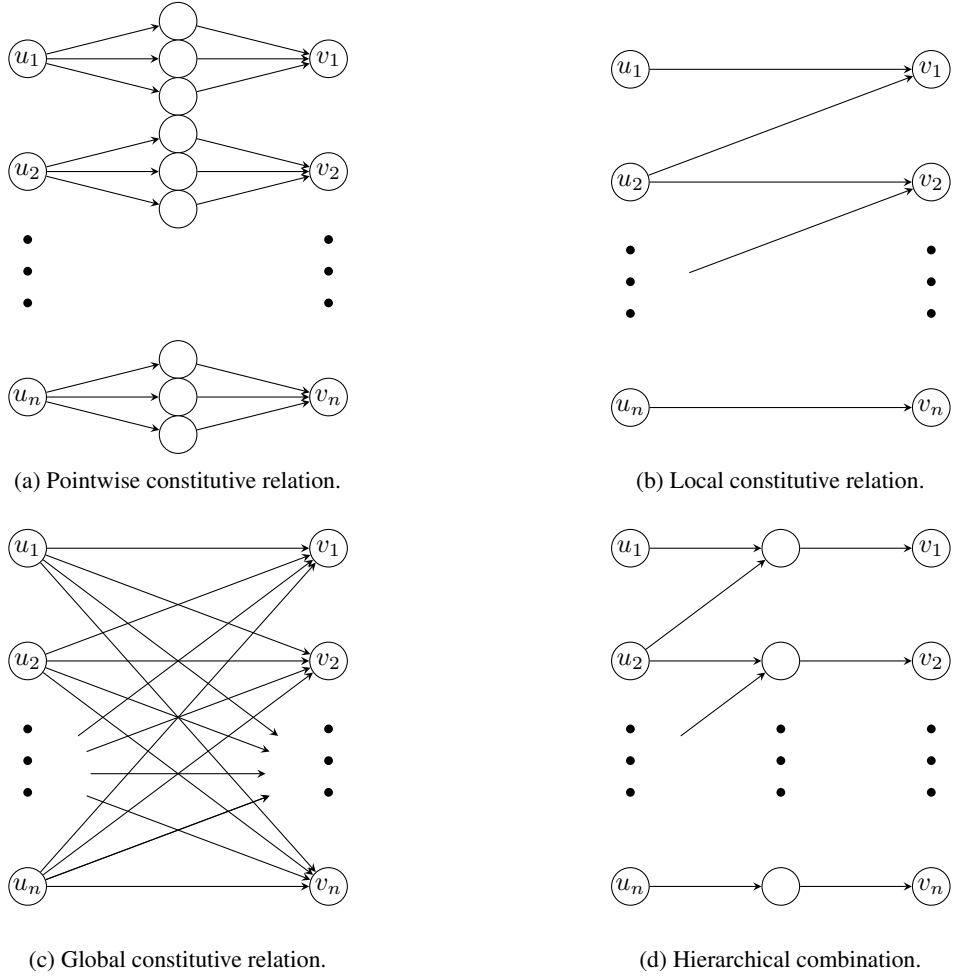
\begin{figure}
    \begin{subfigure}[b]{.48\textwidth}
    \centering
        \begin{tikzpicture}[x=2cm, y=1cm, >=stealth]
            \foreach \m [count=\y] in {1,2,missing,3}
              \node [every neuron/.try, neuron \m/.try] (input-\m) at (0,4.5-\y*1.5) {};
            \foreach \m [count=\y] in {1,2,3,4,5,6}
              \node [every neuron/.try, neuron \m/.try,fill=white] (hidden-\m) at (1,4-\y*0.5) {};
            \foreach \m [count=\y] in {7,8,9}
              \node [every neuron/.try, neuron \m/.try,fill=white] (hidden-\m) at (1,-0.5-\y*0.5) {};
            \foreach \m [count=\y] in {1,2,missing,3}
              \node [every neuron/.try, neuron \m/.try,fill=white] (output-\m) at (2,4.5-\y*1.5) {};
            \node [] at (input-1) {$u_1$};
            \node [] at (input-2) {$u_2$};
            \node [] at (input-3) {$u_n$};
            \node [] at (output-1) {$v_1$};
            \node [] at (output-2) {$v_2$};
            \node [] at (output-3) {$v_n$};
            
            
            \foreach \i in {1,2,3}
                \draw [->] (input-1) -- (hidden-\i);
           \foreach \i in {1,2,3}
                \draw [->] (hidden-\i) -- (output-1);
            \foreach \i in {4,5,6}
                \draw [->] (input-2) -- (hidden-\i);
            \foreach \i in {4,5,6}
                \draw [->] (hidden-\i) -- (output-2);
            \foreach \i in {7,8,9}
                \draw [->] (input-3) -- (hidden-\i);
            \foreach \i in {7,8,9}
                \draw [->] (hidden-\i) -- (output-3);
            
        \end{tikzpicture}
        \caption{Pointwise constitutive relation.}
        \label{fig::net_1}
    \end{subfigure}
    \begin{subfigure}[b]{.48\textwidth}
    \centering
        \begin{tikzpicture}[x=2cm, y=1cm, >=stealth]
            
            \foreach \m [count=\y] in {1,2,missing,3}
              \node [every neuron/.try, neuron \m/.try] (input-\m) at (0,4.5-\y*1.5) {};
            \foreach \m [count=\y] in {1,2,missing,3}
              \node [every neuron/.try, neuron \m/.try,fill=white] (output-\m) at (2,4.5-\y*1.5) {};
            \node [] at (input-1) {$u_1$};
            \node [] at (input-2) {$u_2$};
            \node [] at (input-3) {$u_n$};
            \node [] at (output-1) {$v_1$};
            \node [] at (output-2) {$v_2$};
            \node [] at (output-3) {$v_n$};
            
            
            \draw [->] (input-1) -- (output-1);
            \draw [->] (input-2) -- (output-1);
            \draw [->] (input-2) -- (output-2);
            \draw [->] (input-missing) -- (output-2);
            \draw [->] (input-3) -- (output-3);
            
            \node [] at (1,-2) {};
            
        \end{tikzpicture}
        \caption{Local constitutive relation.}
        \label{fig::net_2}
    \end{subfigure} \vspace{0.4cm} \\
    \begin{subfigure}[b]{.48\textwidth}
    \centering
        \begin{tikzpicture}[x=2cm, y=1cm, >=stealth]
        
            \foreach \m [count=\y] in {1,2,missing,3}
              \node [every neuron/.try, neuron \m/.try] (input-\m) at (0,4.5-\y*1.5) {};
            \foreach \m [count=\y] in {1,2,missing,3}
              \node [every neuron/.try, neuron \m/.try,fill=white] (output-\m) at (2,4.5-\y*1.5) {};
                 \node [] at (input-1) {$u_1$};
            \node [] at (input-2) {$u_2$};
            \node [] at (input-3) {$u_n$};
            \node [] at (output-1) {$v_1$};
            \node [] at (output-2) {$v_2$};
            \node [] at (output-3) {$v_n$};
            
            \foreach \i in {1,missing,2,3}
                \foreach \j in {1,2,missing,3}
                    \draw [->] (input-\i) -- (output-\j);
            \draw [->] (input-3) -- (output-missing);
            \node [] at (1,-2) {};
            
        \end{tikzpicture}
        \caption{Global constitutive relation.}
        \label{fig::net_3}
    \end{subfigure}
    \begin{subfigure}[b]{.48\textwidth}
    \centering
        \begin{tikzpicture}[x=2cm, y=1cm, >=stealth]
            
            \foreach \m [count=\y] in {1,2,missing,3}
              \node [every neuron/.try, neuron \m/.try] (input-\m) at (0,4.5-\y*1.5) {};
            \foreach \m [count=\y] in {1,2,missing,3}
              \node [every neuron/.try, neuron \m/.try,fill=white] (hidden-\m) at (1,4.5-\y*1.5) {};
            \foreach \m [count=\y] in {1,2,missing,3}
              \node [every neuron/.try, neuron \m/.try,fill=white] (output-\m) at (2,4.5-\y*1.5) {};
            \node [] at (input-1) {$u_1$};
            \node [] at (input-2) {$u_2$};
            \node [] at (input-3) {$u_n$};
            \node [] at (output-1) {$v_1$};
            \node [] at (output-2) {$v_2$};
            \node [] at (output-3) {$v_n$};
            
            
            \draw [->] (input-1) -- (hidden-1);
            \draw [->] (input-2) -- (hidden-1);
            \draw [->] (input-2) -- (hidden-2);
            \draw [->] (input-missing) -- (hidden-2);
            \draw [->] (input-3) -- (hidden-3);
            \draw [->] (hidden-1) -- (output-1);
            \draw [->] (hidden-2) -- (output-2);
            \draw [->] (hidden-3) -- (output-3);
            
            \node [] at (1,-2) {};
            
        \end{tikzpicture}
        \caption{Hierarchical combination.}
        \label{fig::net_4}
    \end{subfigure}
    \caption{\textbf{Illustration of the different structures associated with constitutive equations.} The different schemes are illustrative for one-dimensional problems. Note that when convolutional layers are applied (local operators), values at the boundaries may not be conveniently described.}
    \label{fig::net}
\end{figure}

All symmetries and simplifications that may be assumed in the constitutive equation relating the two internal variable fields may be transcribed to the structure of the function $\bs{H}$:

\begin{itemize}
\item The linearity of the functional $\mathcal{H}$ is translated directly into the linearity of the function $\bs{H}$. Using the neural network language, this is equivalent to no internal layers between neurons associated with the tensors $\mathtt{u}$ and $\mathtt{v}$, associated with the fields considered. Different levels of complexity and non-linearity of $\mathcal{H}$ (and therefore $\bs{H}$) may be handled with appropriate topologies of the deep neural network relating $\bs{v}$ and $\bs{u}$.
\item The structure of the function $\bs{H}$ is further exploited using the deep neural network topology, involving different levels of sparsity (see Fig. \ref{fig::net}):
\begin{itemize}
    \item Local constitutive laws, such that that the operator $\mathcal{H}$ is local, that is, the value of $\bs{v}$ depends on the values of $\bs{u}(\bs{x})$ in a neighbourhood of $\bs{x}$. This means that $\mathcal{H}(\bs{u}) = \bs{H}(D^1[\bs{u}](\bs{x}),\ldots,D^m[\bs{u}](\bs{x}))$ where $D^k$ is a differential operator of order $k$. $m<\infty$ is called the order of the locality. 
    
    When $m=0$ we speak about order-zero local or pointwise constitutive laws. In that case $\bs{v}(\bs{x}) = \mathcal{H}(\bs{u}(\bs{x}))$, or, using an embedding notation, $\frac{\partial H^{\bullet,I}_{\bullet}}{\partial u^{\bullet,J}_{\bullet}}=\delta^I_J$. This entails \emph{block-diagonal} tensor structures, in the slots associated with the spatial discretization: $\mathtt{v}[:,i,j,k,l]=H(u[:,i,j,k,l])$. In the deep learning framework, these kinds of relationships are associated with partitioned networks, as it is illustrated in Fig. \ref{fig::net_1}.
    
    When $m>0$, the tensors are sparse but not necessarily \emph{block-diagonal}. In the deep learning framework, these operators are associated with convolutional filters, as shown in Fig. \ref{fig::net_2}.
    
    \item Non-local constitutive laws are models so that $\mathcal{H}$ is not a local operator, that is, the value of $\bs{v}$ depends on the values of $\bs{v}(\boldsymbol{x})$ on the whole spatial domain. In the language of differential operators, $\mathcal{H}(\bs{u}) = \bs{H}(D^1[\bs{u}](\boldsymbol x),\ldots,D^m[\bs{u}](\boldsymbol x),\ldots)$. There are many ways of defining non-local functionals (see for instance \cite{ros2015nonlocal} and included references for a motivation and examples in elliptic operators). In that case, the tensors are dense and so it is the topology of the neural network associated with the model, as illustrated in Fig. \ref{fig::net_3}.
\end{itemize}
Obviously, these different situations may be modulated in several hierarchical levels in the network. For instance, in linear elasticity, $\bs \sigma = \frac{1}{2}\bs{C}:\left(\nabla \otimes \bs{u} + \bs{u} \otimes \bs \nabla\right)$, so the material is local with respect to $\bs{u}$. Therefore, an accurate deep learning network for working with this model is obtained by combining the previous ideas, as shown in Fig. \ref{fig::net_4}.
\item The difference between homogeneous and heterogeneous constitutive relations may be exploited also. For local models (block-diagonal or sparse tensors), the different blocks or filters may be the same or may be dependent on the spatial considered point. For the former, the indexes referring to the spatial part (denoted using capital letters) of the tensor are spurious and therefore may be omitted. For instance, in the linear elastic problem, in general $\sigma^{ij,I} = C^{ijkl}(I,J)\upvarepsilon^{J}_{kl}$. The fact that elasticity assumes a relationship $\bs{\sigma}(\bs{x}) = \bs{H}(\bs \varepsilon(\boldsymbol x))$ (point-wise) implies that $C^{ijkl}(I,J)=C^{ijkl}\delta^I_J$  where $\bs{C}(I)$ is the common elasticity tensor. For homogeneous materials, $\bs{C}(I) = \bs{C}$. 
\item The tensor relations may be adapted for the exploitation of further symmetries of the constitutive equation. This includes:
\begin{itemize}
\item Relations derived from the principe of objectivity, that is, reference frame independence.
\item Constraints related to the physical or geometrical foundations of the model (e.g. major and minor symmetries of the elastic tensor, associated with thermodynamics, angular momentum conservation, and compatibility constraints). 
\item Additional constraints related to special symmetries of the constitutive model, that is, orthotropy, isotropy...
\end{itemize}

All these symmetries may be enforced by adding constraints to the PGNNIV (that is, in an implicit way) or by assuming a given topology for the deep neural network (explicit way). Indeed, if $\bs{v} = \bs{H}(\bs{u})$, the existence of a given symmetry is equivalent, in the Noether sense, to the action of a given group of transformations, so that $\bs{H}(\bs{u}) = \bs{H}(\bs{A}(\bs{u}))$ where $\bs{A} \in \mathcal{A}$, a group of transformations. Therefore, we can look for a finite set of transformations $\bs{A}_k$ in a way such that $\bs{H}(\bs{u}) = \bs{H}(\bs{A}(\bs{u})), A \in \mathcal{A} \Leftrightarrow \bs{A}_k(\bs{H}(\bs{u})) = \bs{H}(\bs{u})$ or to  \emph{a priory} set up a topology for the deep neural network so that it is invariant under the action of all $\bs{A}$.

\item Finally, the classical framework of parametric fitting is a very particular case, in which some of the internal layers are related to the others by means of a parametric explicit expression. In that case,  $\bs{v} = \bs{H}(\bs{u};\bs{\lambda})$ where the function $\bs{H}$ is explicitly imposed, and the functional relationship depends on the value of unknown parameters $\bs{\lambda}$, that are variable TensorFlow scalars obtained, in general, during the training process.
\end{itemize}

Sometimes it is useful to introduce other operators to enforce higher-order discretizations, or to adapt the problem to other numerical methods. This is the case of special filters for meshless approaches such as SPH \cite{gingold1977smoothed}, DEM \cite{nayroles1992generalizing} or NEM \cite{sukumar1998natural, chinesta2014} among a crowd. Also, it is easy to adapt this framework to integral formulations as in the FEM. Indeed, Finite Element integrals may be expressed in terms of the nodal values, being this relationship dependent on the shape function and the chosen numerical integrator, but otherwise fixed for a given degree of approximation. For instance, a moving averaging filter applied to the nodes recovers the framework of linear shape functions for a given element. It is also possible to increase the order of the differential operators. This relies on the fact that a higher-order differential operator can be expressed as the subsequent application of lower-order ones, enriching the differentiation scheme. For instance, if $\Delta^+_h$ is the forward difference operator ($\Delta^+_h f^i = \frac{1}{h}(f^{i+1}-f^{i})$), of order $h$, $\frac{1}{h}(\Delta_h^+ - \frac{1}{2}(\Delta_h^+)^2)$ is a forward difference operator of order $h^2$. Finally, another useful possibility is that of stabilization filters in time-dependent problems to ensure the fulfillment of well-known stability criteria \cite{fischer2001filter }or the filters that are designed for obtaining and high fidelity time integrations, such as Crank-Nicolson integration or Runge-Kutta integrators \cite{raissi2019physics}.

The last application of filters are probes. Probes are measurable or quantifiable values related to the different fields by a known function. The most common probe is the value of a tensor field evaluated at a point or a region. Other common probes are measurements defined in the Data-Driven context \cite{ayensa2019unsupervised}, such as surface stress forces over a plane or strains along a direction, at a certain point. These values may be expressed in terms of data-dependent or data-independent tensor quantities (contractions with other vector/tensors, trace, etc.). As a particular case, we may consider integral quantities such as the flow of a tensorial field over a surface (mass, momentum or energy flux), related to conserved quantities in field theories (Noether charges).

\clearpage

\section{Examples of validation} \label{sec::validation} 

\subsection{Problem description} \label{sec::problem_description}

The next example illustrates the use of PGNNIV in continuum physical problems after discretization. Let us suppose the following partial differential equation corresponding to a diffusion problem:

\begin{equation}
\label{eq::diffusion}
\boldsymbol \nabla \cdot \left( \boldsymbol{k} \boldsymbol \nabla u\right) = f
\end{equation} 
where $u$ is the solution field and $f$ is the source term. This problem is ubiquitous in physics and engineering. Indeed, Eq. (\ref{eq::diffusion}) is used for instance in stationary $u$ in heat transfer conduction problems with $u$ the temperature, also in steady-state, water seepage in soil mechanics, or in electrostatics, among others. Eq. (\ref{eq::diffusion}) is the combination of two different laws:

\begin{itemize}
\item \textbf{A fundamental functional principle} as it is energy conservation (heat transfer), mass conservation (diffusion) or Gauss law (electrostatics), that states as $\boldsymbol \nabla \cdot \boldsymbol{q} = f$, where $\boldsymbol{q}$ is the flow vector (heat flow, mass flow or electric displacement field) and $f$ is the source term (heat source, mass source or electric charge density).
\item \textbf{A constitutive functional equation} as it is the Fourier law (heat transfer), Fick's law (diffusion) or Dielectric behavior (electrostatics), relating the flux variable $\boldsymbol{q}$ that plays the role of internal state field (non-measurable if no additional assumption is made, e.g. uniform distribution of the transported magnitude through a certain area), with the essential field $u$. Commonly, this relationship is formulated in tensor form as $\boldsymbol{q} = -\boldsymbol{K} \boldsymbol{\nabla} \boldsymbol{u}$, where $\boldsymbol{K}$ is the (thermal) conductivity tensor, the diffusion tensor or the dielectric permittivity tensor, respectively, depending on the particular physical problem considered. For general nonlinear problems, the tensor $\boldsymbol{K}$ may be dependent (in a functional sense) on the field $u$  as well as on the point $\boldsymbol{x}$ as any other field. It is common however to particularize this equation for linear ($\boldsymbol{u}$-independent), homogeneous ($\boldsymbol{x}$-independent) and isotropic simplifications.
\end{itemize}

With these assumptions, Eq. (\ref{eq::diffusion}) may be splitted in:

\begin{subequations}
\label{eq::diffusion_PGNN}
\begin{alignat}{2}
\boldsymbol \nabla \cdot \boldsymbol{q} &= f &&\quad \text{(fundamental principle),} \label{eq::diffusion_PGNN_1} \\
\boldsymbol{q} &= -\boldsymbol{k} \boldsymbol \nabla u &&\quad \text{(constitutive equation),} \label{eq::diffusion_PGNN_2}
\end{alignat} 
\end{subequations}
together with appropriate boundary conditions.

Using the framework described above, Eq. (\ref{eq::diffusion_PGNN_1}) is the universal law of the problem, and Eq. (\ref{eq::diffusion_PGNN_2}) is the internal state equation. The only difference is that here, the fundamental principle and the state equation are expressed in functional form. But this subtlety is bypassed by using any common discretization technique (Finite Element Method, Finite Differences Method,...) so the values of $u$, $\boldsymbol{q}$ and $\boldsymbol{k}$ are replaced by the corresponding interpolating (nodal) values or by the approximation parameters, depending on the particular approach. Note that, if $L$ is the mesh-size and $n$ is the number of nodes, in the limit case when $n \rightarrow \infty$ and $L \rightarrow 0$, $\boldsymbol{u}$ and $u$ are indistinguishable, and so it is for the tensorial fields $\boldsymbol{q}$ and $\bs{k}$.

For instance, for one-dimensional problems, using forward first-order finite differences, a discretized version of Eq. (\ref{eq::diffusion_PGNN}) is:

\begin{subequations}
\label{eq::diffusion_PGNN_disc}
\begin{alignat}{2}
\frac{q_{i+1} - q_i}{L} &= f_{i+1}, &&\quad i=0,\ldots,n. \label{eq::diffusion_PGNN_disc_1} \\
q_i &= -k_i  \frac{u_{i+1} - u_i}{L}, &&\quad i=0,\ldots,n. \label{eq::diffusion_PGNN_disc_2}
\end{alignat}
\end{subequations}

where $L$ is an appropriate mesh size, $u_i$ are the field variables, $q_i$ are the internal state variables and the functional relationship $q = \mathcal{H}(u)$ is now in the form of an algebraic equation $\bs{q} = \bs{H}(\bs{u})$, where we have defined $\bs{q}= (q_0,\ldots,q_n)$, $\bs{u}= (u_0,\ldots,u_n)$ and $\bs{f}= (f_1,\ldots,f_{n})$. Of course, when solving Eq. (\ref{eq::diffusion_PGNN}), or its corresponding discrete version Eq. (\ref{eq::diffusion_PGNN_disc}), proper boundary conditions, parametrized in terms of a set of variables $\bs{g}$, must be supplied. If the problem is now formulated within the PGNNIV framework, and for many cases, these latter boundary values, together with $\bs{f}$, are the natural inputs of the problem, being $\bs{u}$ the output ones. We refer to this problem as the \emph{prediction problem}. However, in other cases, we are rather interested in characterizing a given material from its response to different stimuli. In that case $\bs{g}$, $\bs{f}$ and $\bs{u}$ are the input variables while $\bs{k}$ is the output one in what we denote as the \emph{characterization problem}.

We focus on this work in the prediction of the fields $u$, $q$ and $k$ given a value of the boundary conditions $\bs{g}$ and of the stimuli $\bs{f}$. This will be possible by using the approach stated in Section \ref{sec::methods} such that the nodal values of $\bs{u}$ will be learned from a sufficiently big and varied data set of input-output values, but
constrained by the two following equations:

\begin{subequations}
\label{eq::constraints}
\begin{alignat}{2}
\frac{q_{i+1} - q_i}{L} &= f_{i+1}, &&\quad i=0,\ldots,n. \label{eq::constraints_1} \\
q_i &= -k_i(u_1,\ldots,u_n) \frac{u_{i+1} - u_i}{L}, &&\quad i=0,\ldots,n. \label{eq::constraints_2} \\
u_0 &= g_1, \quad u_n = g_2 \label{eq::constraints_3}
\end{alignat}
\end{subequations}

Eqs. (\ref{eq::constraints}) are formulated so that the internal state variables verify the fundamental principle of flow conservation given by Eq. (\ref{eq::diffusion_PGNN_disc_1}) and the specific boundary conditions. At this level, the main problem relies on the form of the function $k_i = k_i(u_1,\ldots,u_n)$ that is a multiple input - multiple output relationship that will be learned using Deep Learning regression techniques. Further assumptions can be made about the functional form of this relationship, that may be translated to the structure of the deep subnetwork associated to the constitutive equation. For example:

\begin{itemize}
\item Assuming a local relationship of order $m$ between $q$ and $u$, that is, $k_i=F\left((\Delta^+)^m (u_i),(\Delta^+)^{m-1}(u_i),\ldots,(\Delta^+)^0 (u_i)\right)$ where $\Delta^+$ is the forward difference operator and $(\Delta^+)^0 (u_i) = u_i$. Furthermore, it is possible to extend the methodology for non-local operators \cite{ciaurri2018nonlocal}, with the inconvenience of the numerical and computational complexity. In particular, a nonlinear relationship may be reduced to a separable form involving the field $u$, the gradient of the field $u$, and higher-order derivatives, or even non-local operators. Thus, $q_i=\Pi_{j=1}^kF_j\left((\Delta^+)^{r_j}(u_i)^{s_j}\right)$ where now $F_j$ are the functions to be learned. Although this functional form seems arbitrary to some extent, it is ubiquitous in mathematical \cite{osserman2013survey}, physical \cite{frank2005nonlinear,ishizuka2008integral}, engineering \cite{barenblatt1989theory,caffarelli2010nonlinear} and financial \cite{barles1998option,ankudinova2008numerical} problems as it involves a huge range of problems. As a very particular but common case, a pointwise relationship is expressed as $k_i = F_0\left((\Delta^+)^0(u_i)^1\right)$.
\item Assuming a linear (possibly heterogeneous) relationship between the gradient of $u$ and the flow $q$. In that case, $q_i = k_i \Delta^+ \mathrm{u}_i$, that is, $k_i$ are constants. The homogeneous case is a particular one, provided that $k_i = k, i=1,\ldots,n$.
\end{itemize}

Finally the function $k_i = k_i(u_1,\ldots,u_n)$ may be parametrized using model parameters $\bs{\lambda}$ with physical meaning. This approach recovers the classical parametric fitting when a given constitutive model structure is assumed, and is either useful for prediction problems and for model selection or validation \cite{ayensajimenez2020identification}.

For illustrative purposes let us consider the problem:

\begin{equation}
\label{eq::p1}
\frac{d}{dx} \left(k\frac{du}{dx} \right) = 0
\end{equation} 
with boundary conditions

\begin{subequations}
\begin{align}
\label{eq::p1_b}
u(x=0) &= g_1  \\
u(x=1) &= g_2
\end{align}
\end{subequations}

To facilitate the discussion, we shall analyze separately the effect of including heterogeneity and nonlinearities, since each problem has its own particularities, even if both problems may be simultaneously studied in one stroke.

\subsection{Homogeneous and heterogeneous problem}

\subsubsection{Network construction}

In this case, we seek for solutions to the problem Eq. (\ref{eq::p1}) and boundary conditions (\ref{eq::p1_b}) when considering $k = k(x)$. Let us consider for such purposes two PGNNIVs that will be defined hereafter. To evaluate the network performance, we will consider the two cases $k(x)=1$ (homogeneous problem, P1) and $k(x) = x+1$ (heterogeneous problem, P2). The analytical solutions to these two problems are respectively $u(x) = (g_2-g_1)x+b_1$ and $u(x) = \frac{g_2-g_1}{\ln 2} \ln(x+1)+g_1$. The solution in terms of all the fields involved is summarized in Table \ref{table::solution_Homo_vs_Hetero}.

\begin{table}[htbp]
\centering
\begin{tabular}{lccc}
\textbf{Fields} & $u(x)$                                      & $q(x)$                  & $k(x)$ \\ \hline
\textbf{P1}     & $(g_2-g_1)x+g_1$                            & $g_2-g_1$               & $1$    \\ \hline
\textbf{P2}     & $u(x) = \frac{g_2-g_1}{\ln 2} \ln(x+1)+g_1$ & $\frac{g_2-g_1}{\ln 2}$ & $x+1$  \\ \hline
\end{tabular}
\caption{\textbf{Analytical solutions associated with the two considered problems.} The solution $u$, flow $q$ and diffusivity $k$ fields are shown.}
\label{table::solution_Homo_vs_Hetero}
\end{table}

$N$ profiles of $u(x)$ were synthetically generated for different values of the boundary conditions, independently and uniformly sampled on $[0;1]^2$, that is $g_1,g_2 \sim \mathcal{U}[0;1]$ and independents. Together with $g_1$ and $g_2$, $q_1 = q(x=0)$ and $q_2 = q(x=1)$ were considered as input variables to ensure that the PGNNIV is associated with a well-posed problem: we need at least one value of the flow, the solution is unique up to an additive constant value. The values of the field $u$ were particularized at $n=10$ points, $x_i = (i-1)/(n-1), i=1,\ldots,n$ so the output variables correspond therefore to the nodal values $u_i = u(x_i), i=1,\ldots,10$.

\paragraph{Construction of the Reduced Order Model NN.} Let us denote the tensor input as $\mathtt{x}$ (shape $[N,4]$), such that $\mathtt{x}[j,1]=g^j_1$, $\mathtt{x}[j,2]=g^j_2$, $\mathtt{x}[j,3]=q^j_1$ and $\mathtt{x}[j,4]=q^j_2$ while the tensor output is denoted as $\mathtt{u}$ (shape $[N,n]$) such that $\mathtt{u}[i,j]=u^j_i$. To predict the values $u_i$ we use standard neural network regression techniques. In this work, we consider multilayer perceptrons as the fundamental tool for building ROM models, although more sophisticated neural networks may be suitable for other problems. In particular, we used a 4-layer network with two hidden layers, each with 15 neurons, such that $\mathtt{y}=\mathtt{Y}[\mathtt{x}]$ with $\mathtt{Y}$ the nonlinear operator that identifies the input-output relation in the neural network. Up to this point, there is no novelty compared to a traditional ANN approach. With all these notations, the prediction error is

\begin{equation}
    \mathtt{e} = \mathtt{y} -\mathtt{u}
\end{equation} 

\paragraph{Construction of the continuum PGNNIV.} We establish now a tensor operator $\mathtt{D}^*$ associated with the first-order forward differential operator $\Delta^+$ and we define $\mathtt{dy}=\mathtt{D}^*[\mathtt{y}]$ (shape $[N,n-1]$). Now we set a variable tensor, $\mathtt{K}$, of shape $[n-1,n-1]$. The structure of the mathematical equation implies that the tensor $\mathtt{K}$ is diagonal. Here we try two possibilities, resulting in two different PGNNIV: a scalar-tensor (N1) and a general diagonal tensor (N2). The first one is associated with a generic homogeneous problem and the second one to a heterogeneous one. Now we define $\mathtt{q} = -\mathtt{dy}\cdot \mathtt{K}$ (shape $[N,n-1]$) and we concatenate with the two boundary flow values obtaining $\tilde{\mathtt{q}}$ (shape $[N,n+1]$). Finally, we define $\mathtt{f} = \mathtt{D}^*[\tilde{\mathtt{q}}]$, $\bar{\mathtt{u}} = (\mathtt{y}[\cdot,0] - g_1, \mathtt{y}[\cdot,n] - g_2)$ and $\bar{\mathtt{q}} = (\mathtt{q}[\cdot,0] - q_1, \mathtt{q}[\cdot,n-1] - q_2)$. Consequently, the penalties involved in the problem are

\begin{subequations}
\begin{align}
    \mathtt{\pi}_1 &= \mathtt{f} \\
    \mathtt{\pi}_2 &= \bar{\mathtt{u}} \\
    \mathtt{\pi}_3 &= \bar{\mathtt{q}}
\end{align}
\end{subequations}

\paragraph{Cost function, learning algorithm and metaparameters.} As explained in \cite{ayensajimenez2020identification}, a common PGNNIV may be interpreted as a standard NN where the output space is augmented by including new variables (associated with the added constraints) whose exact value is identically zero. Therefore, as in all ANN problems, we have to specify a cost function and a learning algorithm and its associated metaparameters, but also we have to specify the weights associated with the PILs related to the constraints. In the formal minimization problem to be solved, these weights are the penalty coefficients that become, therefore, new metaparameters associated with the NN. In this work, we consider the mean squared error (MSE) as the cost function, both for the error and penalty terms, and we select the ADAMS optimizer \cite{kingma2014adam}. Note that when referring to the MSE of a tensor, we understand the MSE of the sum of the squares of all its components. The differences in the tensor sizes and physical nature (i.e. units) is the reason for the capital role of the selection of the penalty weights. The resulting cost function (CF) of the optimization procedure is, therefore:

\begin{equation}
    \mathrm{CF} = c_0\mathrm{MSE}(\mathtt{e})+c_1\mathrm{MSE}(\mathtt{\pi}_1)+c_2\mathrm{MSE}(\mathtt{\pi}_2)+c_3\mathrm{MSE}(\mathtt{\pi}_3)
\end{equation}

The metaparameters of the PGNNIV are summarized in Table \ref{table::metaparam}.

\begin{table}[htbp]
\centering
\begin{tabular}{lcc}
\textbf{Parameter}    &                                     & \textbf{Value} \\ \hline
Learning rate         &                                     & $0.0003$       \\ \hline
Error coefficients    &                                     & $10^7$         \\ \hline
                      & Flow conservation                   & $10^2$         \\
Penalty coefficients  & Essential boundary conditions       & $10^3$         \\
                      & Natural boundary contitions         & $10^3$         \\ \hline
\end{tabular}
\caption{\textbf{PGNNIV metaparameters.}}
\label{table::metaparam}
\end{table}

\subsubsection{Network convergence}

\begin{figure}[htbp]
\centering
\includegraphics[width=0.8\linewidth]{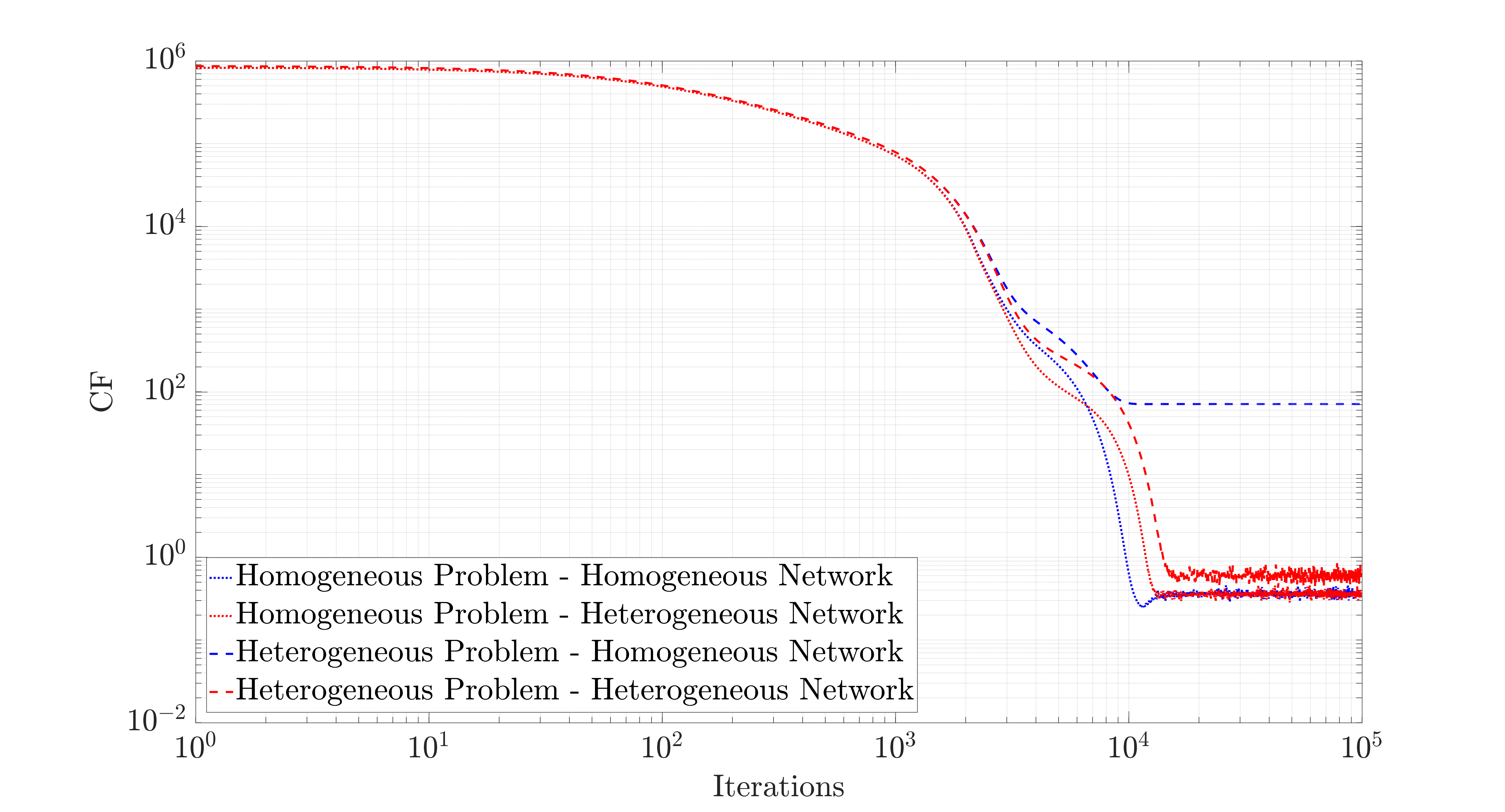}
\caption{\textbf{Learning curve for the different networks and datasets.} CF values are smoothed using a constant filter of bandwidth $w=1000$ to avoid sharp oscillations in the logarithmic scale.}
\label{fig::HH_convergence}
\end{figure}

To evaluate the performance of the continuum-based PGNNIV we generated $N=10^4$ samples of input-output values for both P1 and P2 problems. We used $80\%$ of the generated values as training data and $20\%$ as test data. At each iteration along the optimization process, the PGNNIV was fed with the whole training data. The process was stopped after $10^5$ iterations. The convergence of both neural networks is shown in Fig. \ref{fig::HH_convergence}, demonstrating that a low value of the cost function was obtained in both cases for the homogeneous problem. Nonetheless, for the homogeneous problem, the homogeneous network showed a faster convergence as it includes less learning parameters. However, only the heterogeneous network reached similar low values for the heterogeneous problem. 

\begin{figure}
	\centering
	\begin{subfigure}{.8\textwidth}
		\includegraphics[width=\linewidth]{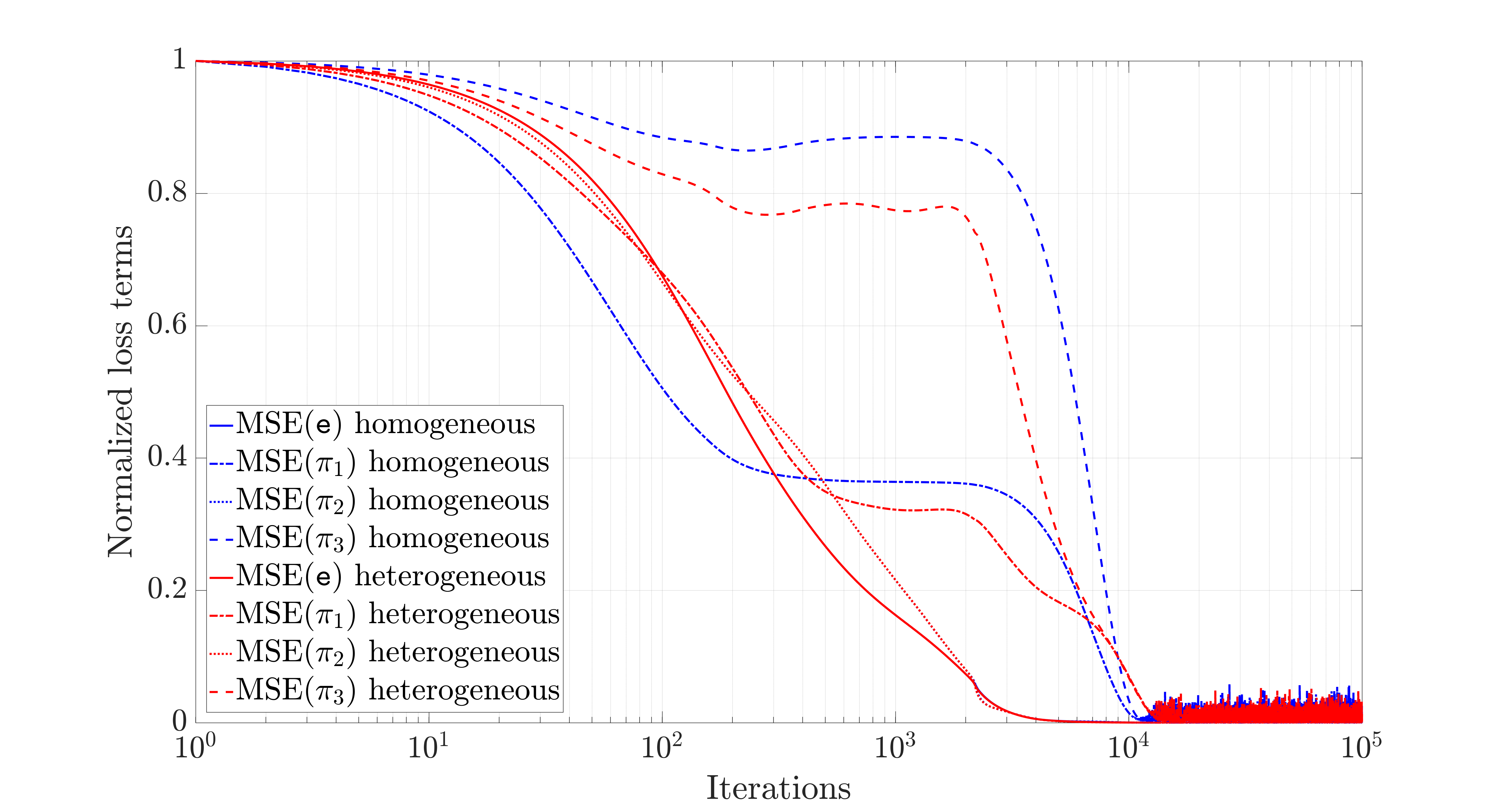}
		\caption{Homogeneous problem.}
		\label{fig::HH_Loss_1}
	\end{subfigure}\\
	\begin{subfigure}{.8\textwidth}
		\includegraphics[width=\linewidth]{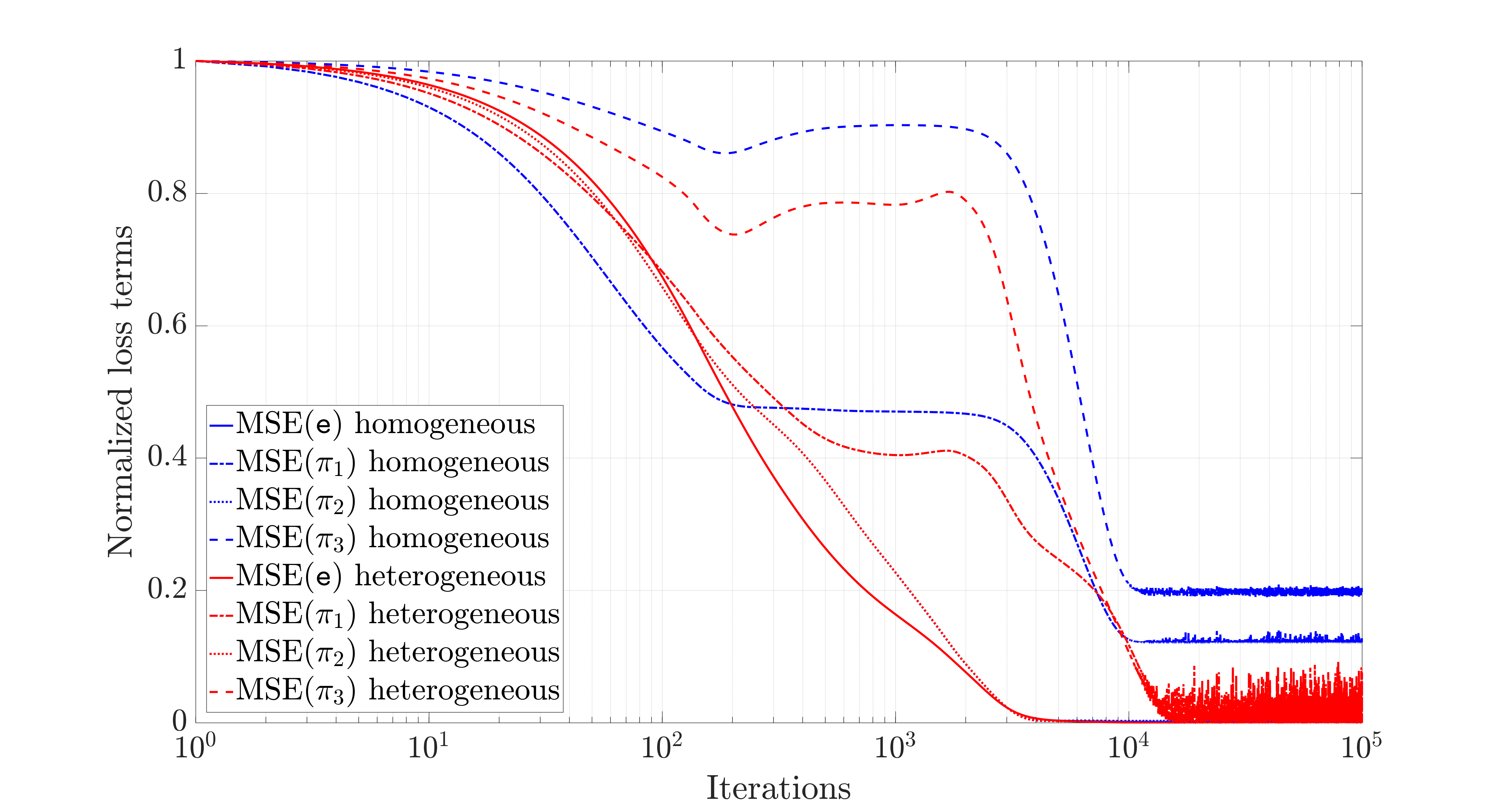}
		\caption{Heterogeneous problem.}
		\label{fig::HH_Loss_2}
	\end{subfigure}
	\caption{\textbf{Evolution of the loss term along the optimization process.} The different loss terms are normalized by its maximal value for a simpler and more consistent comparison. Curves associated with $\mathtt{e}$ and $\mathtt{\pi_2}$ overlap as they are both associated with displacement fields.}
	\label{fig::HH_loss}
\end{figure}

The inability of the homogeneous network to reach low error predictions for the heterogeneous problem is explained by the impossibility of obtaining arbitrarily small values for the penalties associated with flow conservation for such a limited model. This is well illustrated in Fig. \ref{fig::HH_loss}, wich shows the value of the different penalty terms during the optimization process. The penalty terms associated with the flow are unable to reach sufficiently low values because they are incompatible with the assumption of a homogeneous material.

\begin{figure}
    \centering
	\begin{subfigure}{.8\textwidth}
		\includegraphics[width=\linewidth]{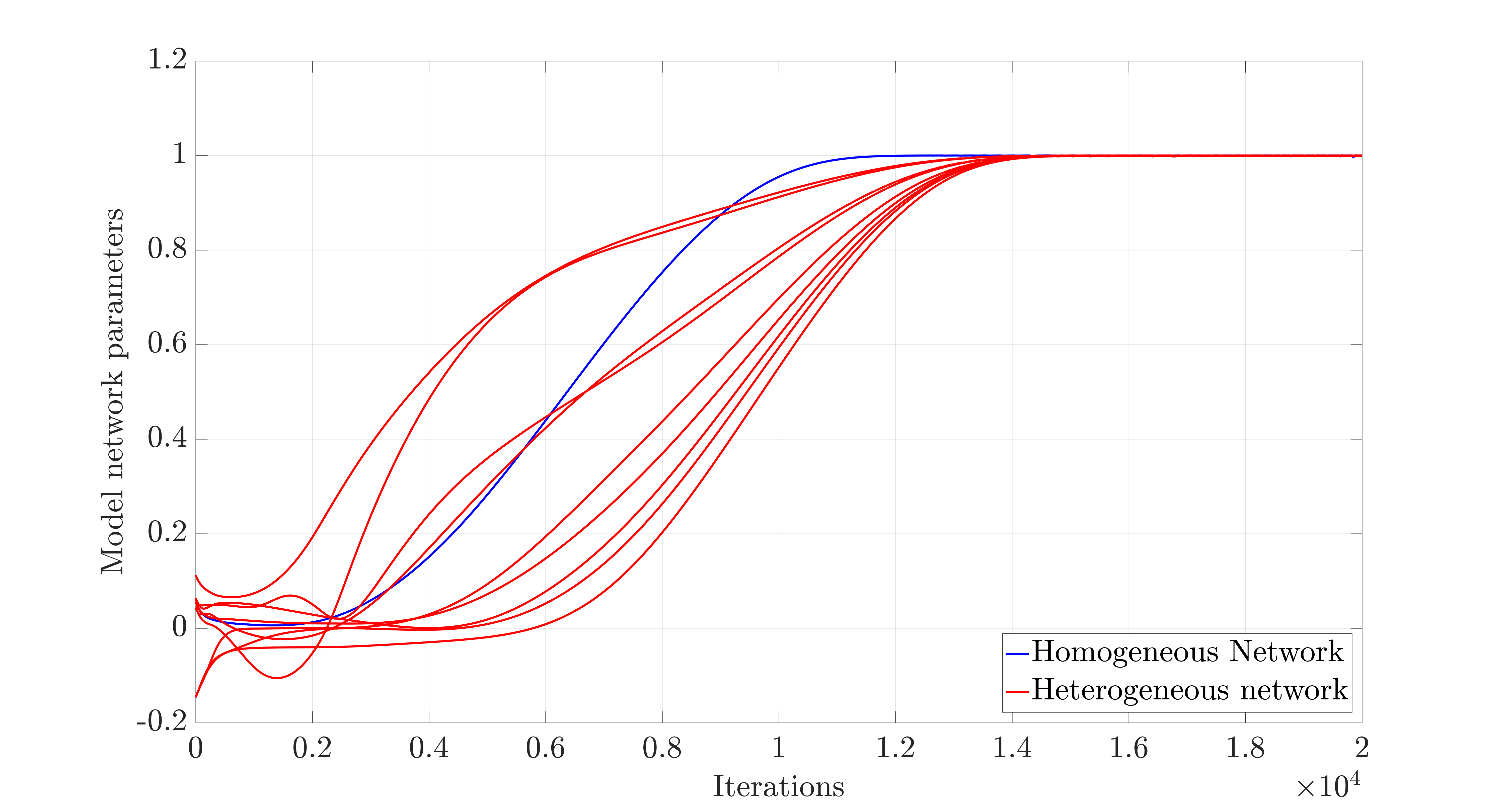}
		\caption{Homogeneous problem.}
		\label{fig::HH_Weights_1}
	\end{subfigure} \\
	\begin{subfigure}{.8\textwidth}
		\includegraphics[width=\linewidth]{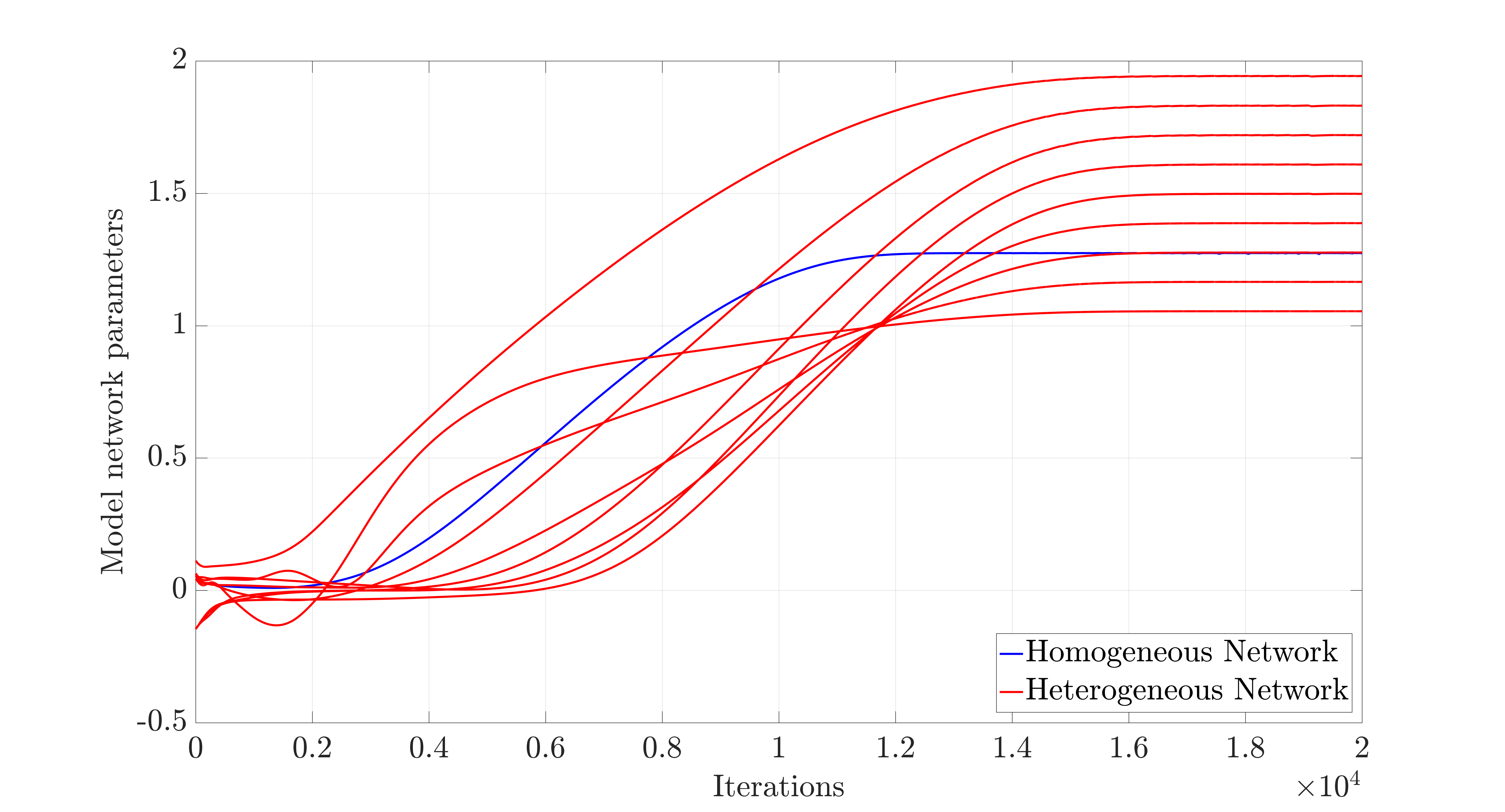}
		\caption{Heterogeneous problem.}
		\label{fig::HH_Weights_2}
	\end{subfigure} 
	\caption{\textbf{Network model parameter convergence.} The value of the parameters of the tensor $\mathtt{K}$ are plotted until $M=2\times 10^4$ iterations, where the model network has converged.}
	\label{fig::HH_weights}
\end{figure}

Finally, it is interesting to compare the convergence of the network in terms of the model network parameters. This is illustrated in Fig. \ref{fig::HH_weights}. As for the homogeneous problem, all model network parameters converge to the theoretical value ($k=1$), both for the homogeneous and heterogeneous network, albeit the homogeneous one converges faster. However, for the heterogeneous problem, each model parameter of the heterogeneous network model converges to a value satisfying the nodal constitutive relationship (that is, the nodal value $k_i$), while the homogeneous network does whatever possible to reduce the CF, that is, the only model parameter converges to an intermediate value of the diffusivity $k$, so it never achieves the same predictive power as the heterogeneous network.

\subsubsection{Predictive capacity}

\begin{figure}
	\centering
	\begin{subfigure}{.7\textwidth}
		\includegraphics[width=\linewidth]{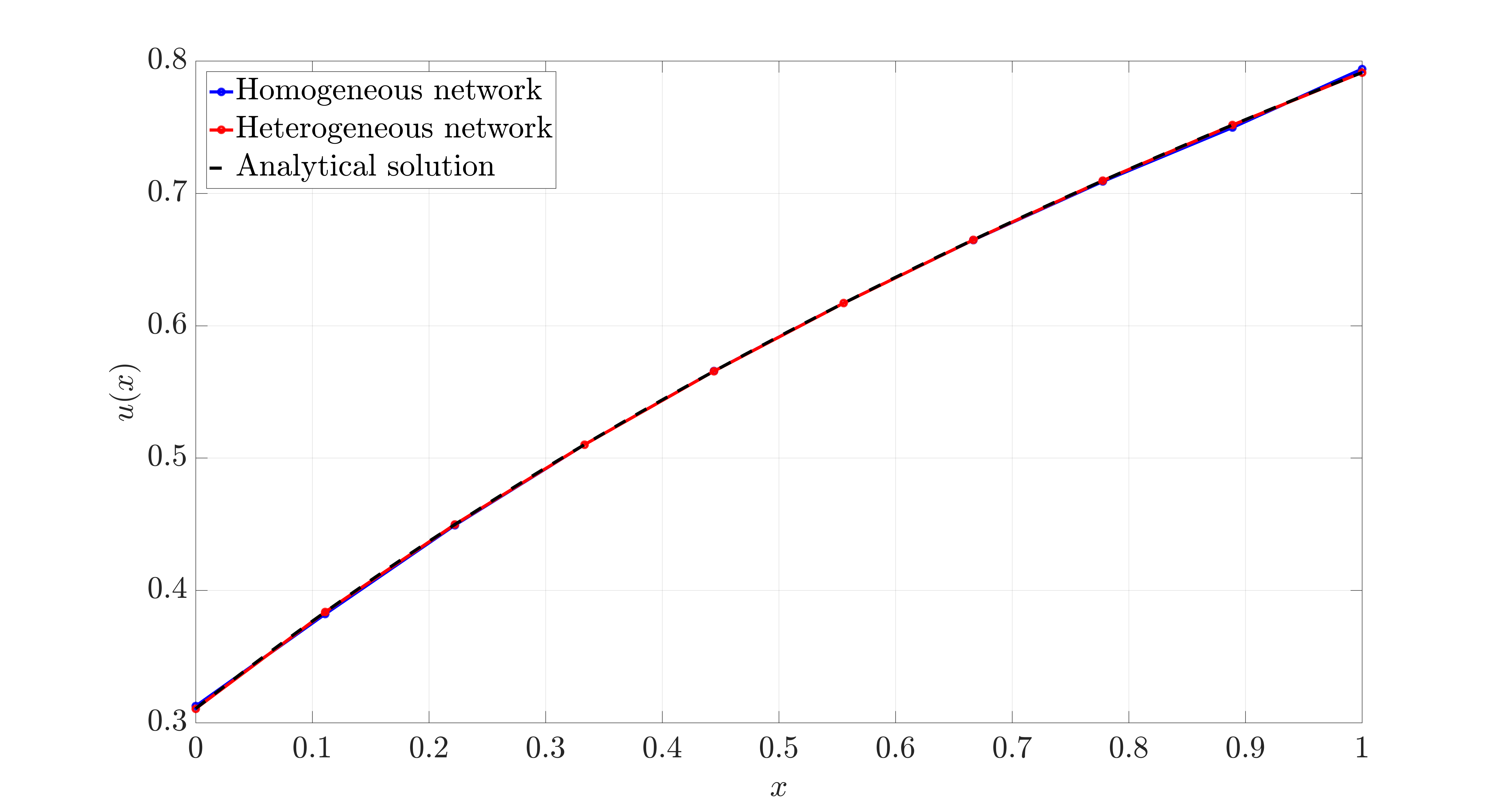}
		\caption{Field $u$.}
		\label{fig::HH_pred_u}
	\end{subfigure}\\
	\begin{subfigure}{.7\textwidth}
		\includegraphics[width=\linewidth]{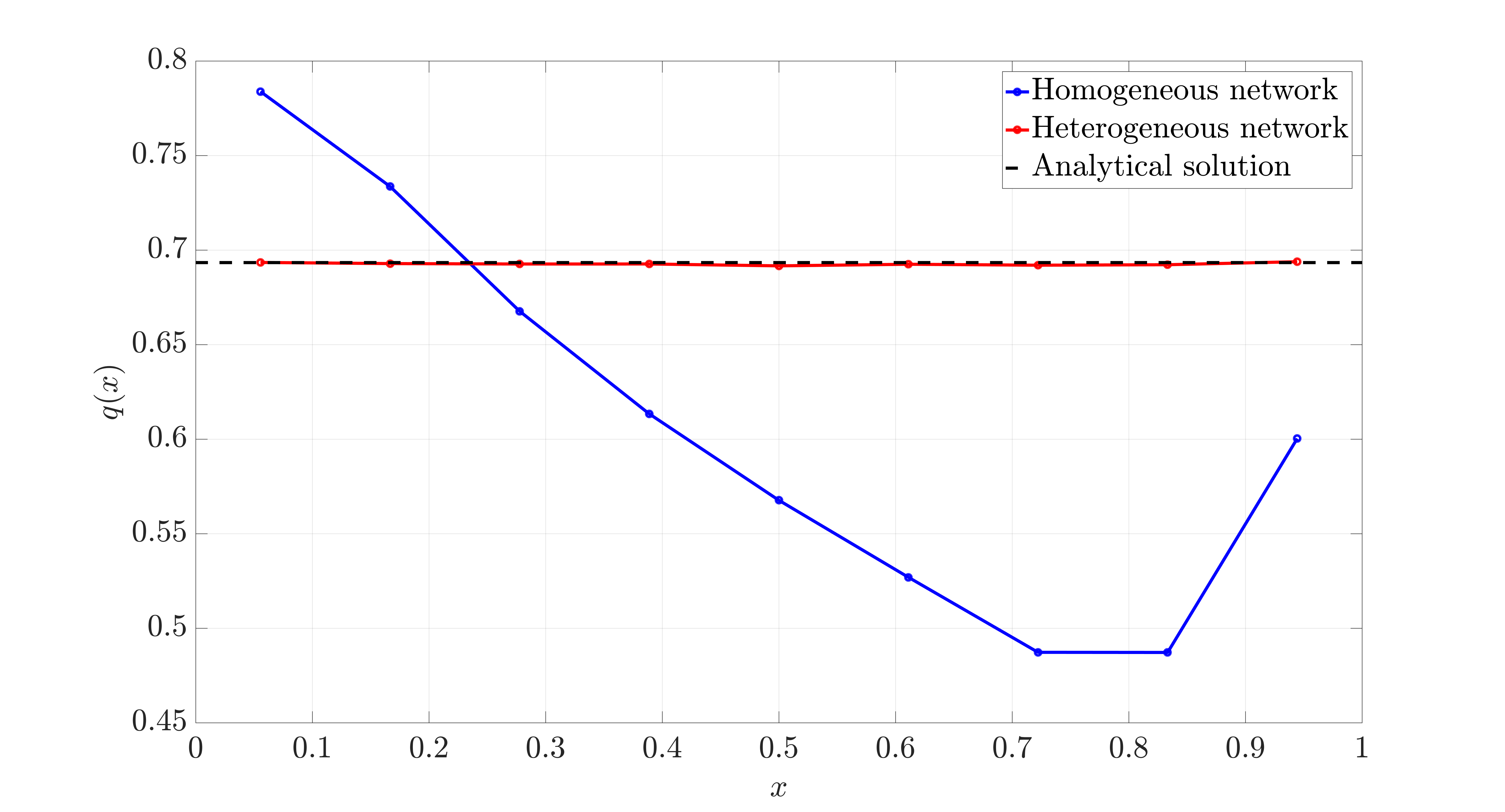}
		\caption{Field $q$.}
		\label{fig::HH_pred_q}
	\end{subfigure} \\
		\begin{subfigure}{.7\textwidth}
		\includegraphics[width=\linewidth]{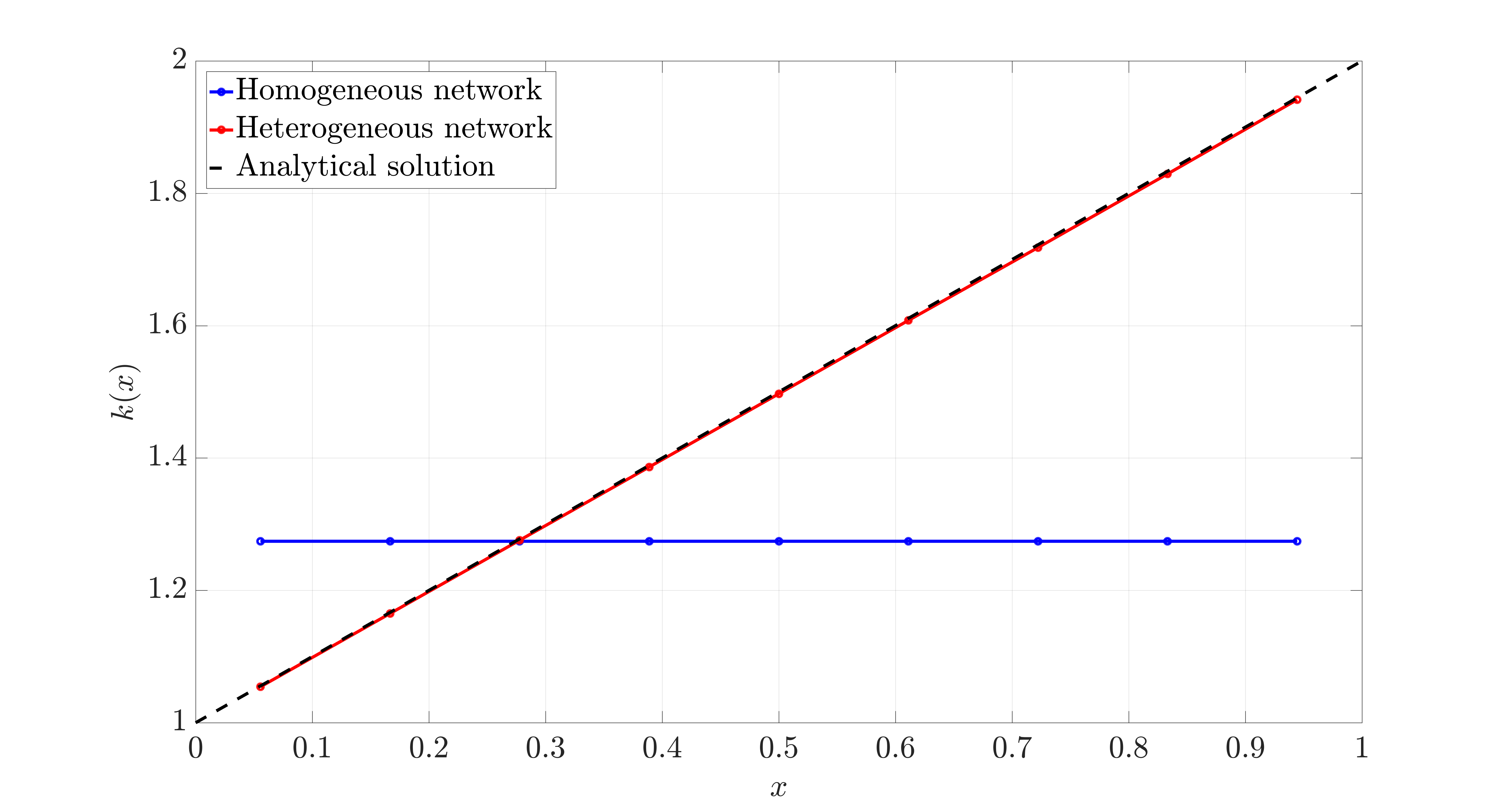}
		\caption{Field $k$.}
		\label{fig::HH_pred_k}
	\end{subfigure}
	\caption{\textbf{PGNNIV prediction of the output fields.} Both neural networks are able to predict the output field $u$ for the heterogeneous problem but the heterogeneous network outperforms the homogeneous one when predicting fields $q$ and $k$.}
	\label{fig::HH_illustration}
\end{figure}

Once the network has converged, we can predict the values of the field $u$ by simply interpolating the obtained values for the nodes, $u_i$, predicted by the network. Note that this prediction for the whole space where the input variables have been sampled (in our case, $b_1,b_2\in [0;1]$), has a minimal cost (the one of a single evaluation), as in any other neural network once trained,  since it does not require the inversion of any system of equations neither any iteration procedure. Moreover, the values of the fields $q$ and $k$ are obtained as a byproduct of the network without any post-process beyond the nodal interpolation. As a simple illustration, Fig. \ref{fig::HH_illustration} shows the neural network prediction of the essential and derivative fields for the heterogeneous problem and one particular set of boundary conditions ($g_1 = 0.31$, and $g_2 = 0.79$). 

Even if we show the field associated with one single value of the inputs, the performance is general for the whole coverage of the input values. Indeed, the statistics of the normalized $L^2$ errors corresponding to the prediction of the different fields are shown in Table \ref{table::HH_table_L2_1} and  \ref{table::HH_table_L2_2}. This error is computed by using the estimate:

\begin{equation}
    E^2_r[f] = \frac{\int_0^1(\tilde{f}(x)-f(x))^2\, dx}{\int_0^1 f(x)^2\, dx}
\end{equation}
with $\tilde{f}$ the predicted value of the field and $f$ the true value given by the analytical solution (see Table \ref{table::solution_Homo_vs_Hetero}).

\begin{table}[htbp]
\centering
\begin{tabular}{|l|ccccc|ccccc|}
\hline
                      &     &       & $E^2_r[u]$ &       &     &     &       & $E^2_r[q]$ &       &     \\ \hline
                      & min & $Q_1$ & $Q_2$      & $Q_3$ & max & min & $Q_1$ & $Q_2$      & $Q_3$ & max \\\hline
N1   &  $1 \cdot 10^{-4}$   &   $1 \cdot 10^{-4}$    &    $1 \cdot 10^{-4}$         &   $2 \cdot 10^{-4}$    &  $4 \cdot 10^{-1}$   &  $2 \cdot 10^{-4}$   &  $6 \cdot 10^{-4}$     &     $1 \cdot 10^{-3}$       &     $2 \cdot 10^{-3}$  &  $1$  \\
N2 &  $5 \cdot 10^{-5}$   &   $1 \cdot 10^{-4}$    &    $1 \cdot 10^{-4}$        &   $1 \cdot 10^{-4}$    &  $4 \cdot 10^{-1}$   &  $5 \cdot 10^{-4}$   &   $6 \cdot 10^{-4}$    &     $7 \cdot 10^{-4}$       &   $1 \cdot 10^{-3}$     &  $1$   \\ \hline
\end{tabular}
\caption{\textbf{Statistics of the error when predicting the output fields for the homogeneous problem.}}
\label{table::HH_table_L2_1}
\end{table}

\begin{table}[htbp]
\centering
\begin{tabular}{|l|ccccc|ccccc|}
\hline
                      &     &       & $E^2_r[u]$ &       &     &     &       & $E^2_r[q]$ &       &     \\ \hline
                      & min & $Q_1$ & $Q_2$      & $Q_3$ & max & min & $Q_1$ & $Q_2$      & $Q_3$ & max \\ \hline
N1   &  $7 \cdot 10^{-6}$   &   $7 \cdot 10^{-4}$    &    $1 \cdot 10^{-3}$         &   $1 \cdot 10^{-3}$    &  $4 \cdot 10^{-1}$   &  $2 \cdot 10^{-1}$   &  $2 \cdot 10^{-1}$     &     $2 \cdot 10^{-1}$       &     $2 \cdot 10^{-1}$  &  $1$  \\
N2 &  $1 \cdot 10^{-4}$   &   $3 \cdot 10^{-4}$    &    $6 \cdot 10^{-4}$        &   $1 \cdot 10^{-3}$    &  $4 \cdot 10^{-1}$   &  $1 \cdot 10^{-3}$   &   $2 \cdot 10^{-3}$    &     $2 \cdot 10^{-3}$       &   $3 \cdot 10^{-3}$     &  $1$    \\ \hline
\end{tabular}
\caption{\textbf{Statistics of the error when predicting the output fields for the heterogeneous problem.}}
\label{table::HH_table_L2_2}
\end{table}

\begin{table}[htbp]
\centering
\begin{tabular}{|lc|c|}
\hline
&   & $E^2_r[k]$  \\ \hline
Homogeneous problem           & N1 & $7 \cdot 10^{-6}$   \\
                              & N2 & $5 \cdot 10^{-4}$\\ \hline
Heterogeneous problem         & N1 & $2 \cdot 10^{-1}$   \\
                              & N2 &  $7 \cdot 10^{-3}$\\ \hline
\end{tabular}
\caption{\textbf{Statistics of the error when predicting the diffusivity field for both problems.}}
\label{table::HH_table_L2_3}
\end{table}

As the output field $k(x)$ does not depend on the value of the boundary conditions, all quantile indicators collapse to a single error value, as shown in Table \ref{table::HH_table_L2_3}.

Note that except for some very particular predictions, the error remains small if the PGNNIV is able to learn the constitutive relation (less than $1\%$ error for more than $75\%$ of the predictions). Only the homogeneous network fails when estimating the values of the field $q$, which is the one associated with the constitutive model. Consequently, the PGNNIV is not capable either of learning accurately the value of $k$ (error of the order $20\%$). 

\begin{figure}
    \centering
	\begin{subfigure}{.8\textwidth}
		\includegraphics[width=\linewidth]{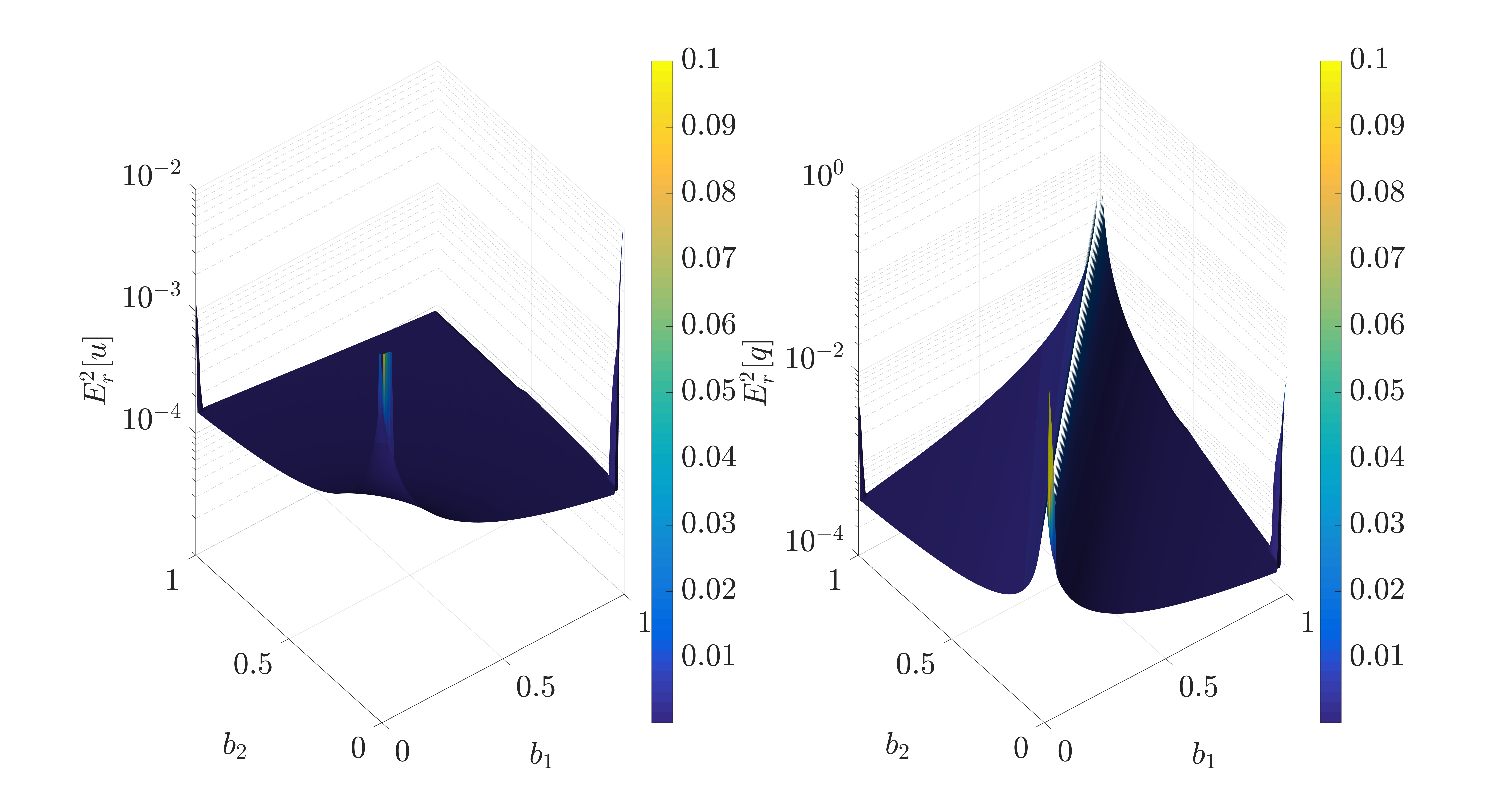}
		\caption{Homogeneous network.}
		\label{fig::HH_errors_b_1a}
	\end{subfigure} \\
	\begin{subfigure}{.8\textwidth}
		\includegraphics[width=\linewidth]{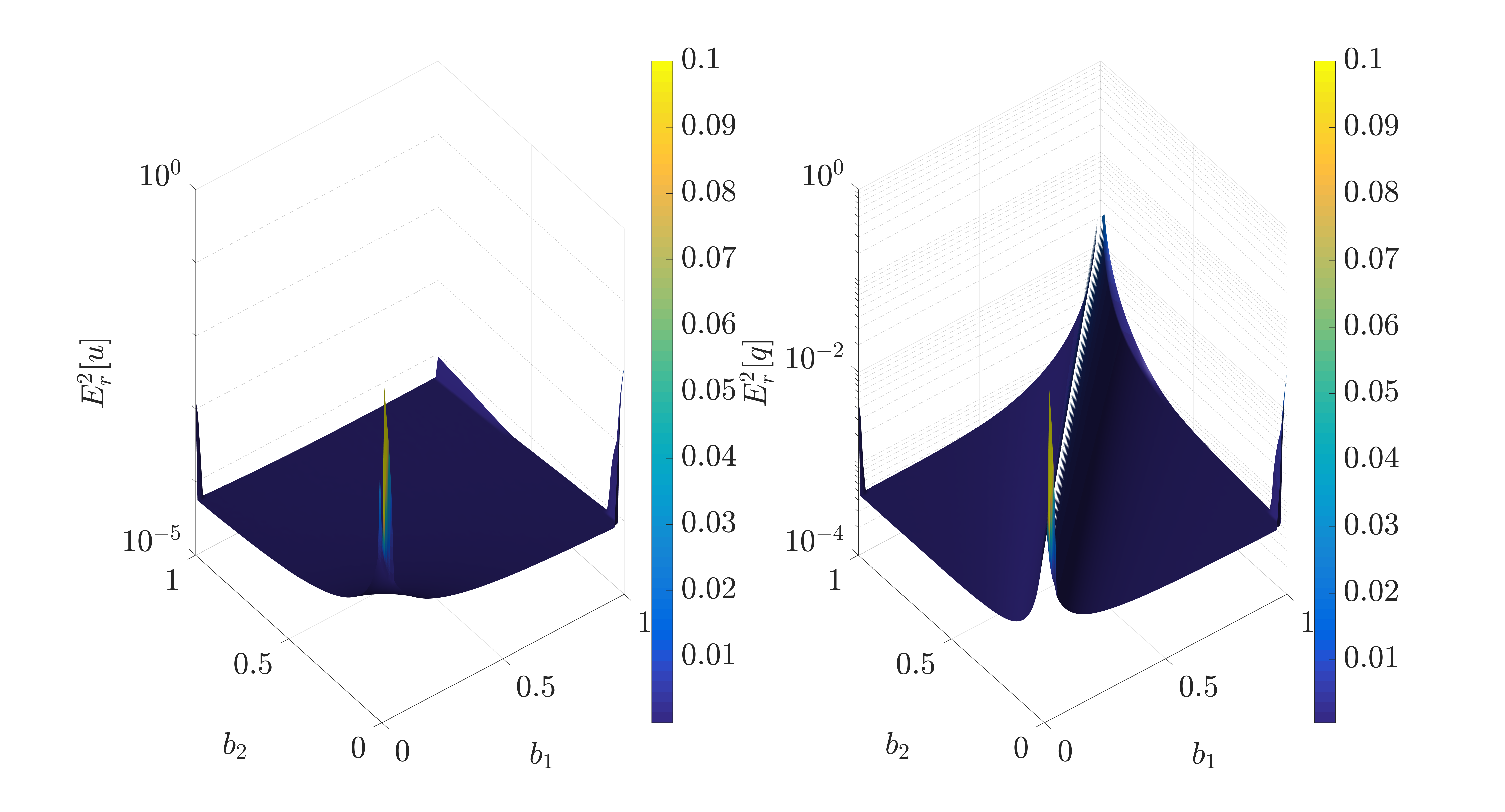}
		\caption{Heterogeneous network.}
		\label{fig::HH_errors_b_1b}
	\end{subfigure} 
	\caption{\textbf{Errors in the homogeneous problem for both neural networks.} Both neural networks are able to predict accurately the fields $u$ and $q$ (less than $1\%$ error) except for the case $g_1=g_2$ when $q(x)=0$.}
	\label{fig::HH_errors_b_1}
\end{figure}

\begin{figure}
    \centering
	\begin{subfigure}{.8\textwidth}
		\includegraphics[width=\linewidth]{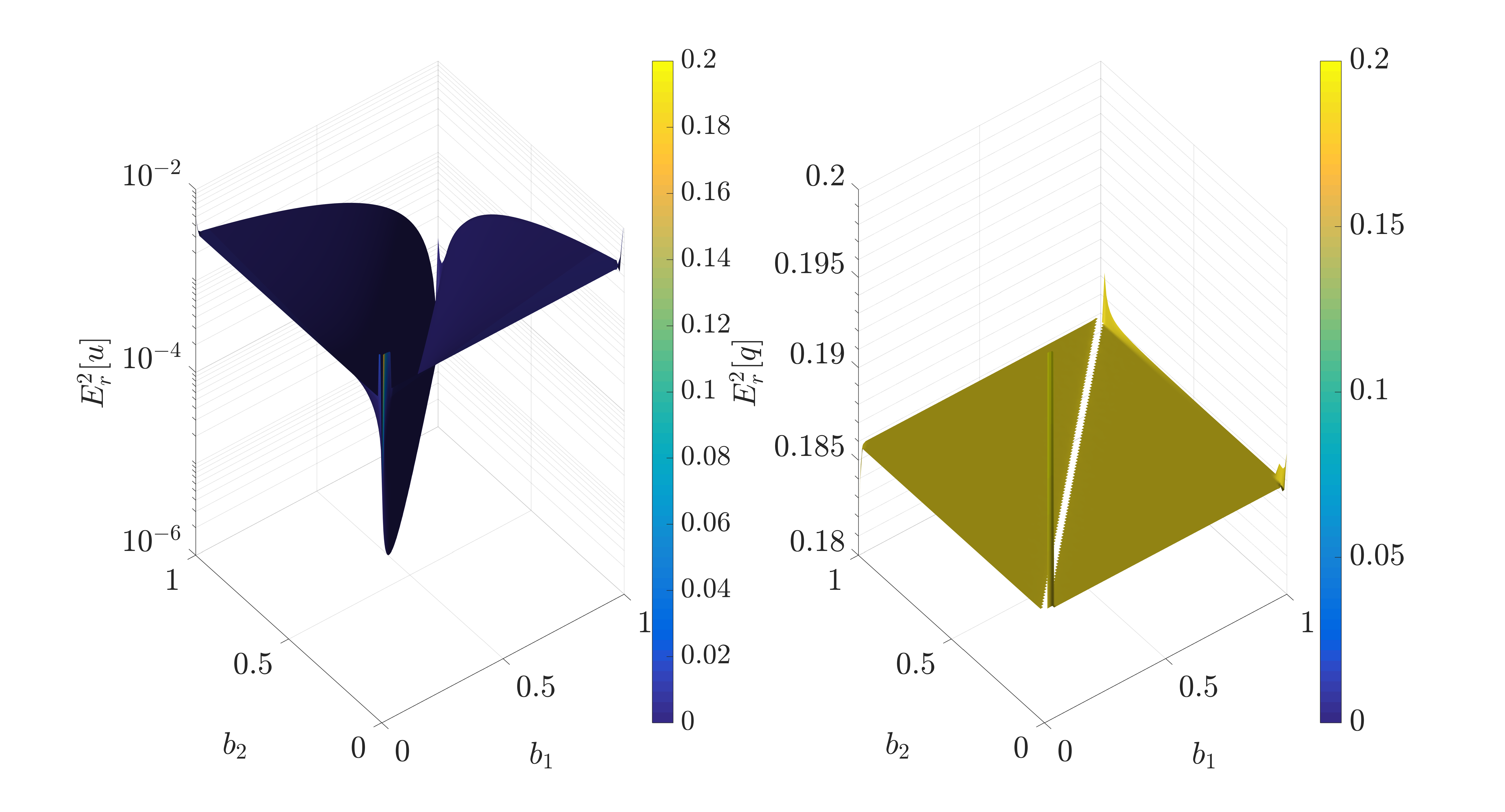}
		\caption{Homogeneous network.}
		\label{fig::HH_errors_b_2a}
	\end{subfigure} \\
	\begin{subfigure}{.8\textwidth}
		\includegraphics[width=\linewidth]{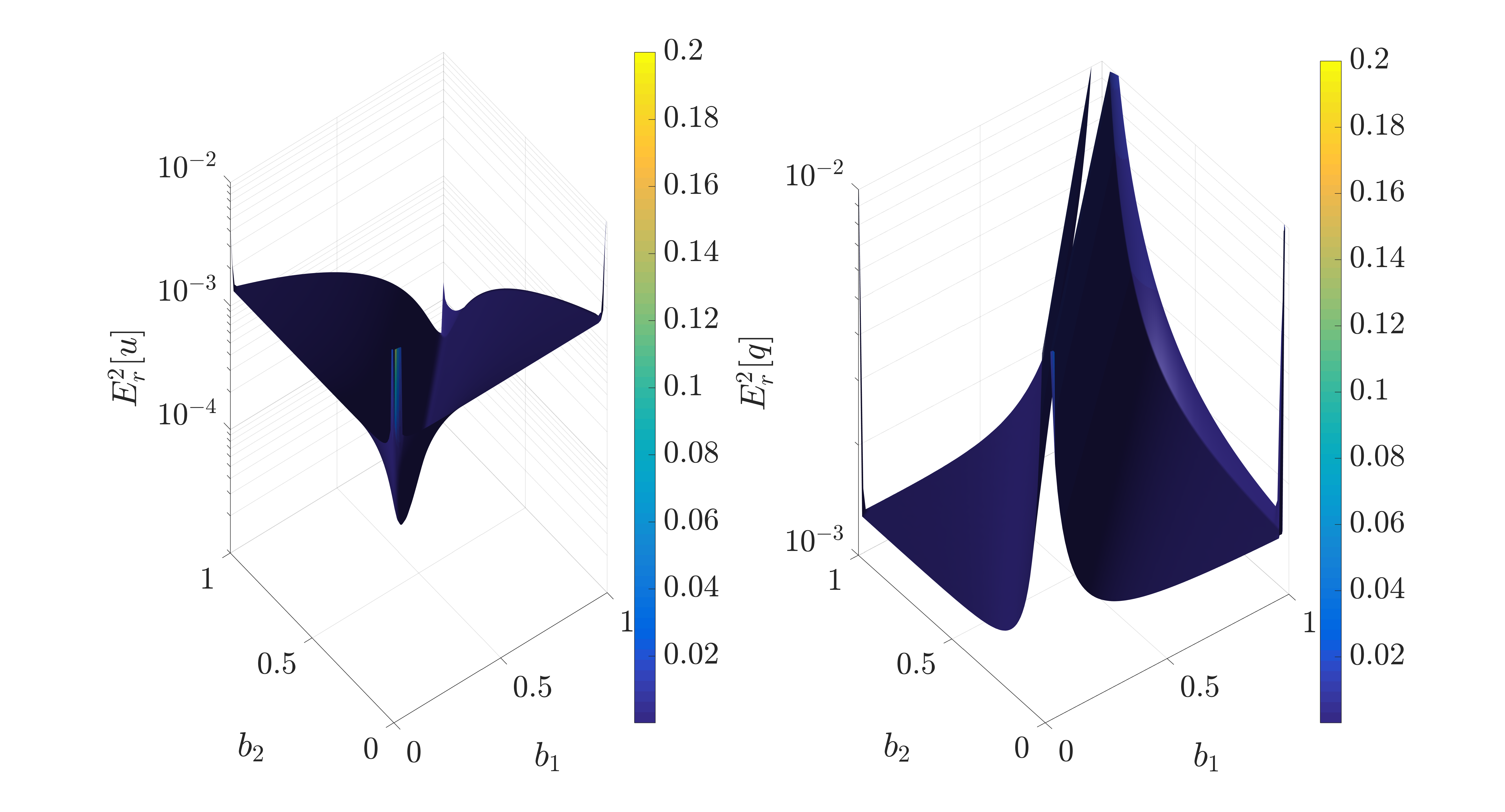}
		\caption{Heterogeneous network.}
		\label{fig::HH_Weights_2b}
	\end{subfigure} 
	\caption{\textbf{Errors in the heterogeneous problem for both neural networks.} $u$ is accurately predicted by the two neural networks, while only the heterogeneous neural network is able to predict accurately the field $q$ (less than $1\%$ error, except for the case $g_1=g_2$ when $q(x)=0$).}
	\label{fig::HH_errors_b_2}
\end{figure}

In Figs. \ref{fig::HH_errors_b_1} and \ref{fig::HH_errors_b_2} the error is depicted as a function of the boundary conditions for both neural networks and problems. As it may be seen, the error remains always small, independently of the values of $g_1$ and $g_2$, when estimating the field $u$ for both problems and neural networks. The heterogeneous PGNNIV is additionally able to accurately estimate the field $q$, except for values close to the line $g_1=g_2$, when of course $q(x)=0$ and therefore $E^2_r[q] \rightarrow \infty$. Apart from this singular case, the error is generally higher when getting closer to the boundaries of the dataset coverage ($g_1=0,1$, and/or $g_2=0,1$), which is expected due to the self-learning nature of the method presented.

\clearpage

\subsection{Linear and nonlinear problem}

\subsubsection{Network construction}

With the aim of getting now a nonlinear version of the problem (\ref{eq::p1}) and boundary conditions (\ref{eq::p1_b}), we state $k = k(u)$. Let us consider now three PGNNIVs that are described in what follows. To evaluate the network performance, we shall consider three cases $k(u)=1$ (constant diffusivity, P1) and $k(u) = u$ (linear diffusivity, P2) and $k(u)=\exp(u)$ (exponential diffusivity, P3). The analytical solutions to these three problems are, respectively, $u(x) = (g_2-g_1)x+g_1$, $u(x) = \sqrt{(g^2_2 - g^2_1)x+b_1^2}$ and $u(x) = \ln\left((\exp(g_2)- \exp(g_1))x+\exp(g_1)\right)$ as well as the corresponding fields are summarized again in Table \ref{table::solution_NN}.

\begin{table}[htbp]
\centering
\begin{tabular}{lccc}
\textbf{Fields} & $u(x)$                                      & $q(x)$                  & $k(x)$ \\ \hline
\textbf{P1}     & $(g_2-g_1)x+g_1$                            & $g_2-g_1$               & $1$    \\ \hline
\textbf{P3}     & $u(x) = \sqrt{(g^2_2 - g^2_1)x+g_1^2}$      & $\frac{g^2_2-g^2_1}{2}$  & $\sqrt{(g^2_2 - g^2_1)x+g_1^2}$  \\ \hline
\textbf{P2}     & $u(x) = \ln\left((\exp(g_2)- \exp(g_1))x+\exp(g_1)\right)$ & $\exp(g_2)- \exp(g_1)$ & $(\exp(g_2)- \exp(g_1))x+ \exp(g_2)$  \\ \hline
\end{tabular}
\caption{\textbf{Analytical solutions associated with the three considered problems.} The solution $u$, flow $q$ and diffusivity field $k$ are shown.}
\label{table::solution_NN}
\end{table}

The input and output values that feed the neural network are generated analogously to the previous example.

\paragraph{Construction of the Reduced Order Model NN.} One of the advantages of this methodology is that the ROM neural network only depends on the nature of the input and output variables. Therefore, the ROM-NN for this nonlinear problem is exactly the same as in the previous example and so it is for the prediction error. The nature of the hidden state equation only affects the physical constraints associated with the PILs layers.

\paragraph{Construction of the continuum PGNNIV.} It is at this level where a supplementary effort has to be made. From the tensor $\texttt{y}$, we obtain the derivative field again using the tensorial operator $\mathtt{D}^*$, $\mathtt{dy}=\mathtt{D}^*[\mathtt{y}]$ (shape $[N,n-1]$). However, we define additionally a new tensor $\mathtt{k}$ representing the diffusivity associated with each element. Note that this field has a shape of $[N,n-1]$ as it is defined on the elements rather than on the nodes. When linking the diffusivity with the value of the field, this has to be done at the element level, so we define an element field $\mathtt{um}=\mathtt{M}^*[\mathtt{u}]$ (shape $[N,n-1]$) obtained by averaging the nodal values of the field associated with the considered element. Note that, as stated in the Methodology section, this operator is formulated in terms of a tensor operator, so a convolutional one-dimensional filter of size 2 with constant kernel $1/2$ was used. The next step is to define a neural network model relating the tensors $\mathtt{um}$ and $\mathtt{k}$. Note that the point-wise character of this (unknown) relationship, that is, the fact that $\mathtt{k}[N,j] = f(\mathtt{um}[N,j])$, is easily formulated in the deep learning TensorFlow framework by defining a convolutional neural network that expands in higher dimensional spaces the content of each neuron associated with each element. 

For illustrative purposes we try two possibilities, resulting in two different PGNNIV: a 2-layer CNN (no hidden layers) and a 3-layer CNN. For the first, no activation functions were used so the CNN is able to reproduce only linear relationships. For the second one, we used sigmoid activation functions in the hidden layer. In all layers, we consider bias terms before and after applying the activation function. Another possibility, which was explored alternatively, was to prescribe a parametric relation between the two tensors $\mathtt{um}$ and $\mathtt{k}$. Here we illustrate this possibility by prescribing $k(u) = \alpha + \beta u^\gamma$ where $\alpha, \beta$ and $\gamma$ are model parameters.

As a summary the tensorial flow at the model network is, for the three proposed approaches:

\begin{align}
    &\underbrace{\mathtt{um}}_{[N,n-1]} \underbrace{\rightarrow}_{[1,n-1,n-1]} \underbrace{\mathtt{k}}_{[N,n-1]} \nonumber \\
    &\underbrace{\mathtt{um}}_{[N,n-1]} \underbrace{\rightarrow}_{[1,m,n-1,n-1]} \underbrace{\mathtt{h}}_{[N,m,n-1]} \underbrace{\rightarrow}_{[m,1,n-1,n-1]} \underbrace{\mathtt{k}}_{[N,n-1]} \nonumber \\
    &\underbrace{\mathtt{um}}_{[N,n-1]} \underbrace{\rightarrow}_{[1,n-1,n-1]} \underbrace{\mathtt{k}}_{[N,n-1]} \nonumber
\end{align}
where we indicate under each tensor or operation, the shape of the tensor or filter. The different penalty terms related to the PGNNIV are the same used in the previous example.

\paragraph{Cost function, learning algorithm and metaparameters.} All algorithms and metaparameters used in these examples are the same as in the preceding one, with the exception that there is an extra metaparameter related to the network topology, which is the size of the convolutional filter in the second network, that is, $m$. For the following results obtained, we set a value of $m=5$.

\subsubsection{Network convergence}

\begin{figure}
    \centering
	\begin{subfigure}{.7\textwidth}
		\includegraphics[width=\linewidth]{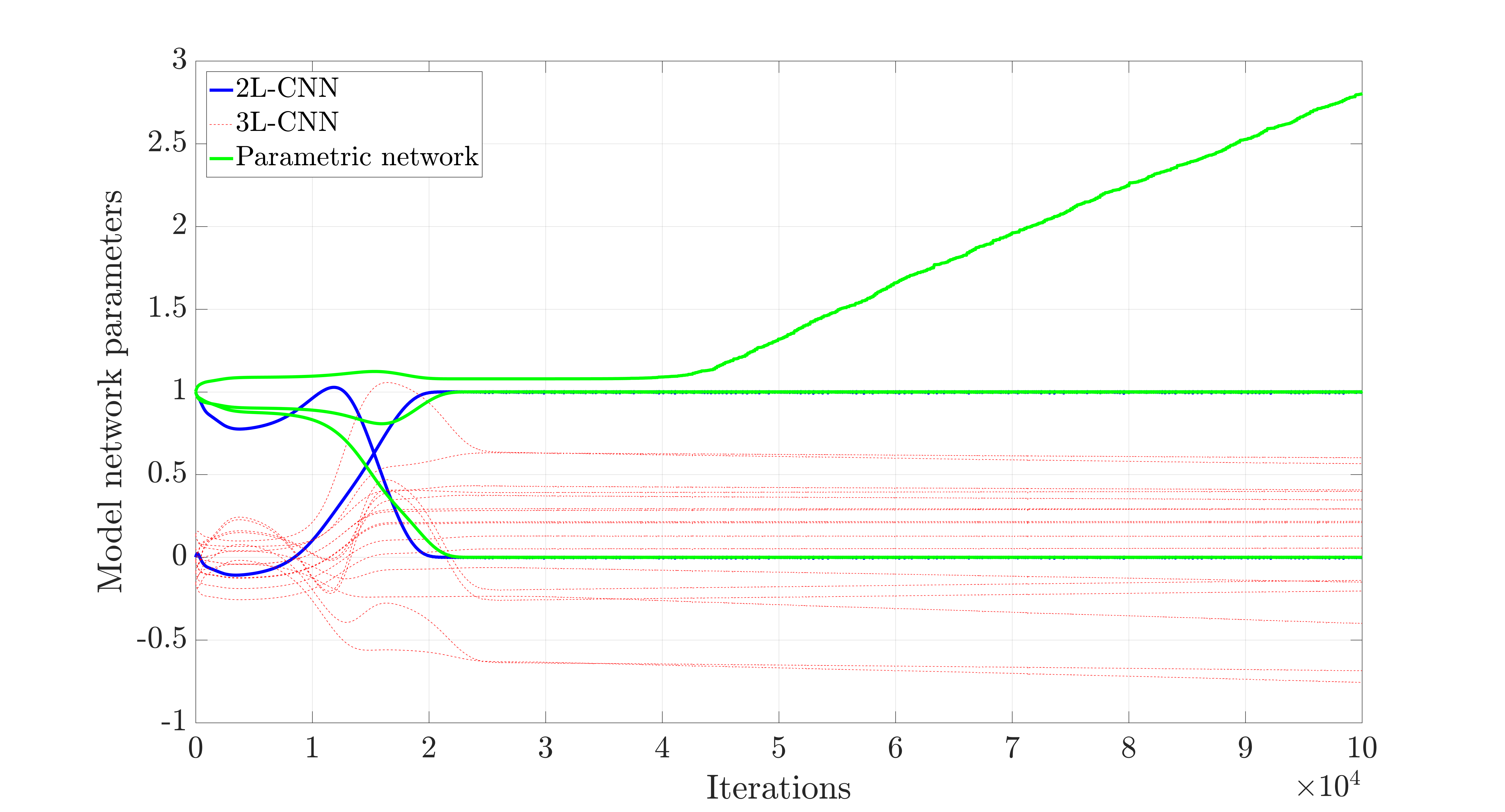}
		\caption{Constant diffusivity.}
		\label{fig::NN_Weights_1}
	\end{subfigure} \\
	\begin{subfigure}{.7\textwidth}
		\includegraphics[width=\linewidth]{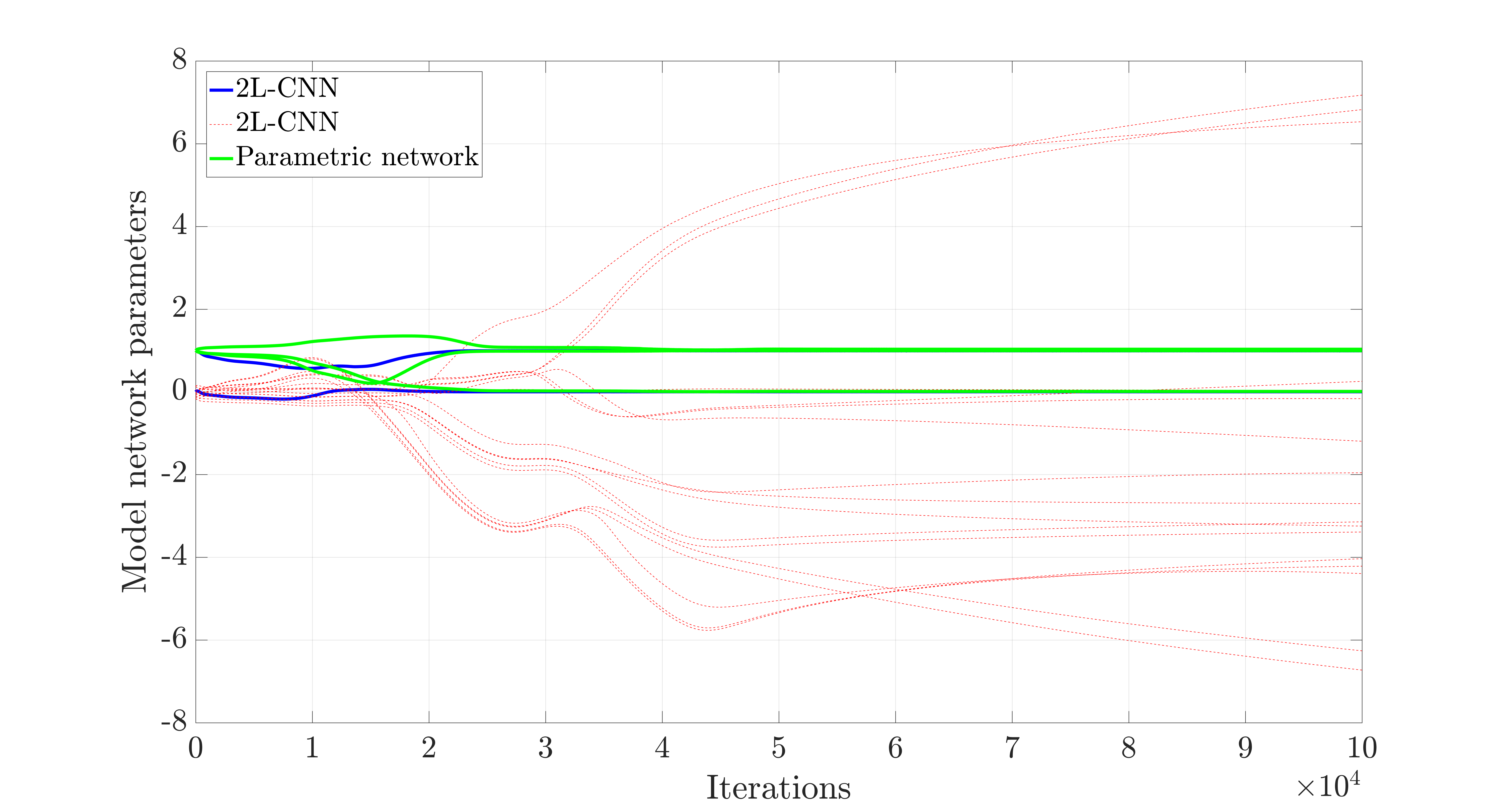}
		\caption{Linear diffusivity.}
		\label{fig::NN_Weights_2}
	\end{subfigure}  \\
	\begin{subfigure}{.7\textwidth}
		\includegraphics[width=\linewidth]{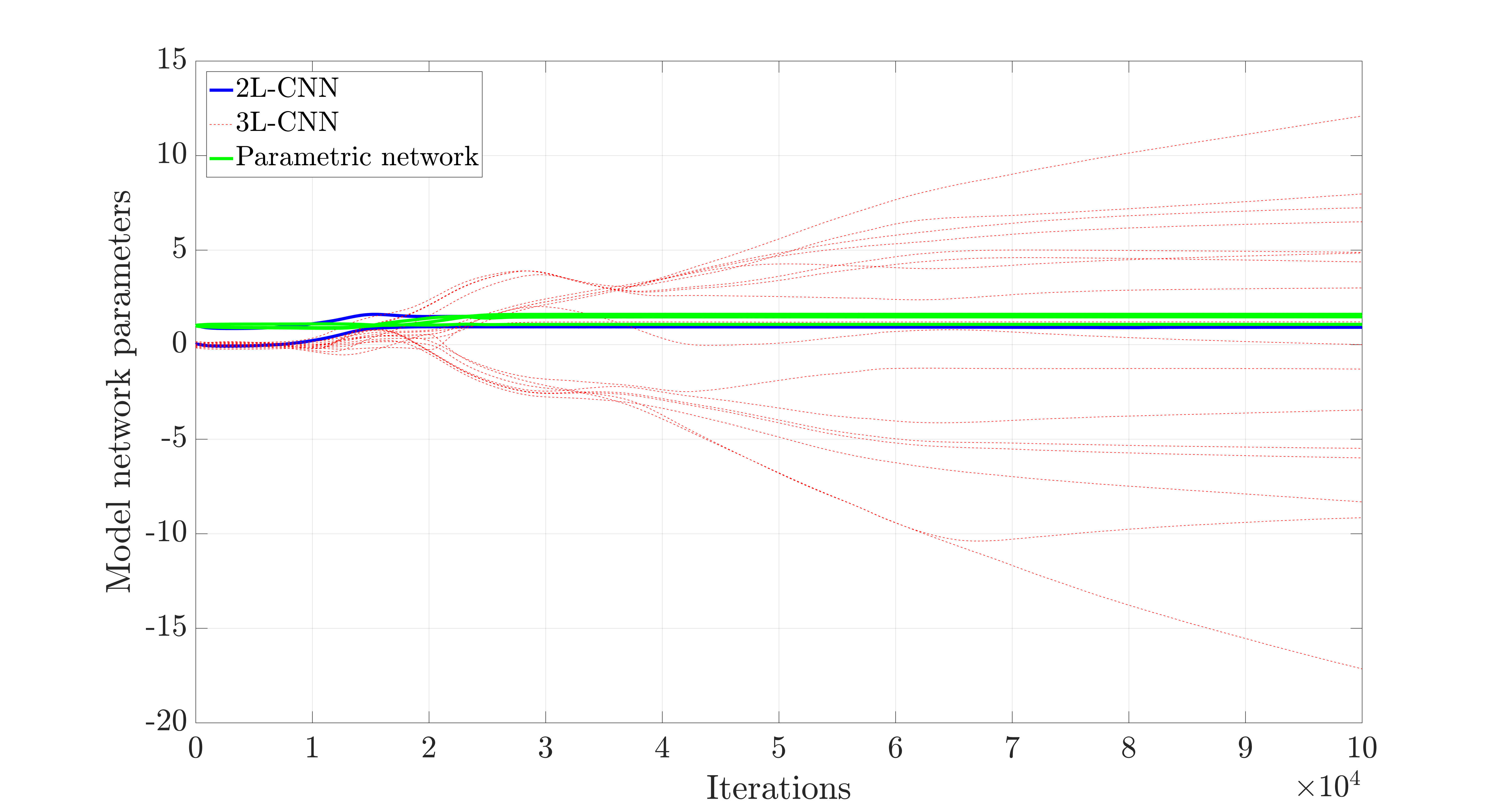}
		\caption{Exponential diffusivity.}
		\label{fig::NN_Weights_3}
	\end{subfigure} 
	\caption{\textbf{Network model parameter convergence.} The values of all the network parameters are displayed, but only those represented in continuous line have a direct interpretation in terms of parameters in the relation $k = k(u)$ .}
	\label{fig::NN_weights}
\end{figure}

The convergence in terms of the CF and the different penalties presents the same trend discussed before when analyzing Fig. \ref{fig::HH_convergence}. However, it is interesting here to show the evolution of the model parameters during the optimization problem, which is informative about the nature of the model $k = k(u)$ (Fig. \ref{fig::NN_weights}). 

First, it is easy to observe that the parameters of both the 2L-CNN model and the parametric CNN are easily interpreted in physical terms. Indeed, for the problem $k(u) = 1$, the weight of the 2L-CNN converges to $0$ and the bias converges to $1$ because $k(u) = 1\cdot u + 0$.  This also happens in the parametric network, where $\alpha \rightarrow 1$ and $\beta \rightarrow 0$ although there is a third spurious parameter, $\gamma$ that remains undetermined and therefore its convergence is not guaranteed. The parameters associated with the 3L-CNN, althought not so easily interpretable, do converge, which is indicative of the fact that the CNN network is able to find an optimal solution, despite the solution got accuracy or not. 

For the linear diffusivity case, the interpretation is similar: the weight of the 2L-CNN converges to $1$ and the bias converges to $0$ while for the parametric network, $\alpha \rightarrow 0.06$, $\beta \rightarrow 0.97$ and $\gamma \rightarrow 1.2$. Note that even if the underlying model is learned well enough, the numerical error intrinsic to the network induces another error in the parameter estimation, which darkens the linear $k-u$ relationship. This may be dramatic when extrapolating, being this a well-known drawback when using complex parametric models without paying attention to overfitting. The 3L-CNN did not totally converged after $M=10^5$ iterations even if, as we will see later, yields good enough results. This is possibly due to an excess of network parameters (excess of neurons) and/or to an insufficiently good model learning approach that could be improved with longer runs or using different optimization algorithms (from the mathematical point of view, the problem is not bounded or the search algorithm has not reached a local minimum).

Finally, in the exponential diffusivity problem, the 2L-CNN network and the parametric network reach convergence close to $M=3\times 10^4$ iterations while for the parametric network we get $\alpha = 1.06$, $\beta = 1.60$ and $\gamma=1.47$. This  may be interpreted as the optimal least squares solution for the parametric problem, when using the CF considered. This solution may be also obtained by using another optimization approach different from backpropagation (for instance, the Levenberg-Marquardt algorithm\cite{levenberg1944method} commonly used in parametric fitting). This latter family of algorithms is usually resource-intensive and complex to use in large scale problems. The behavior of the 3L-CNN performance is, in that case, similar to the one for the linear diffusivity problem.

\subsubsection{Predictive capacity}

\begin{figure}[htbp]
	\centering
	\begin{subfigure}{.7\textwidth}
		\includegraphics[width=\linewidth]{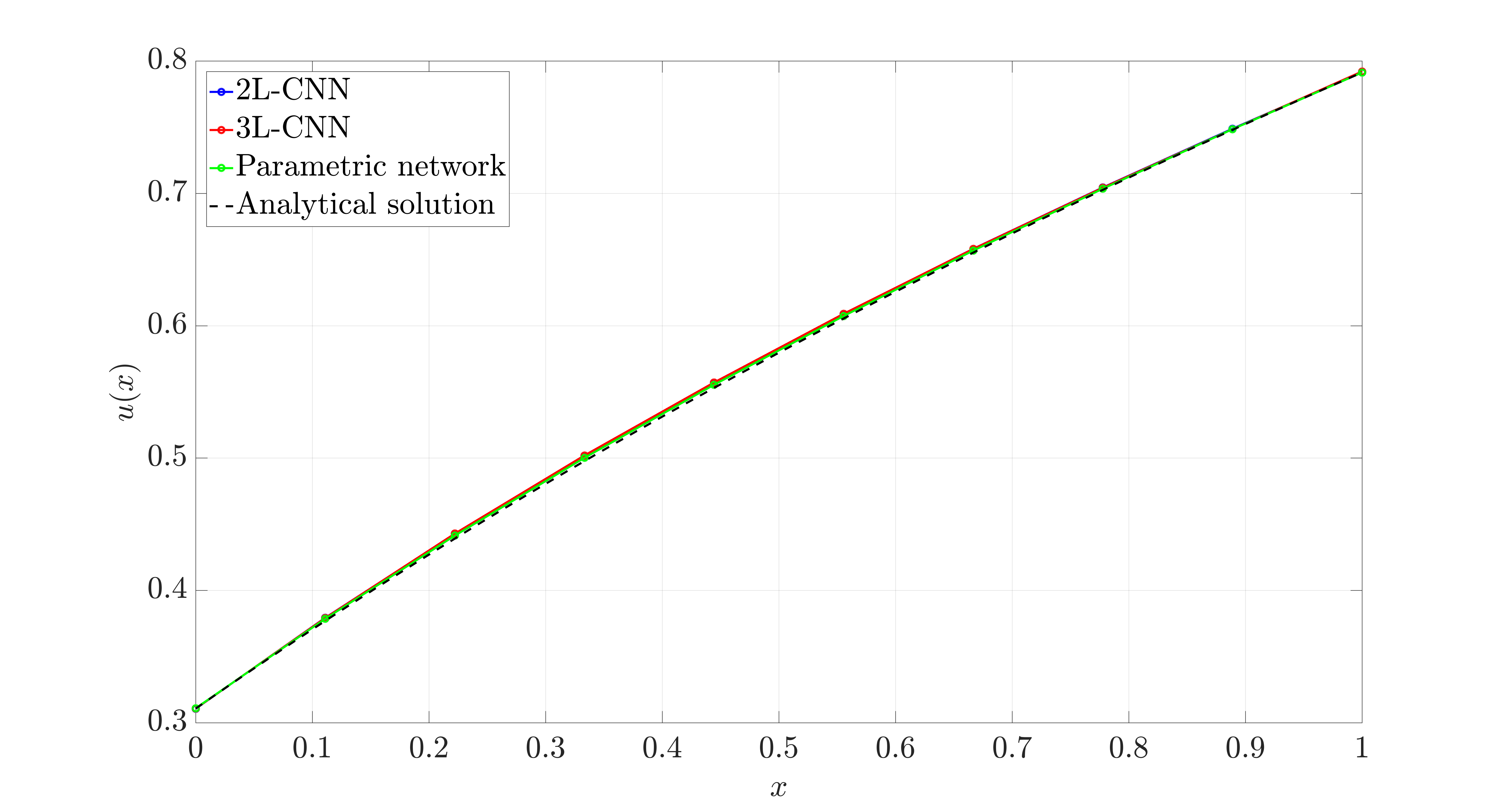}
		\caption{Field $u$.}
		\label{fig::NN_pred_u}
	\end{subfigure}\\
	\begin{subfigure}{.7\textwidth}
		\includegraphics[width=\linewidth]{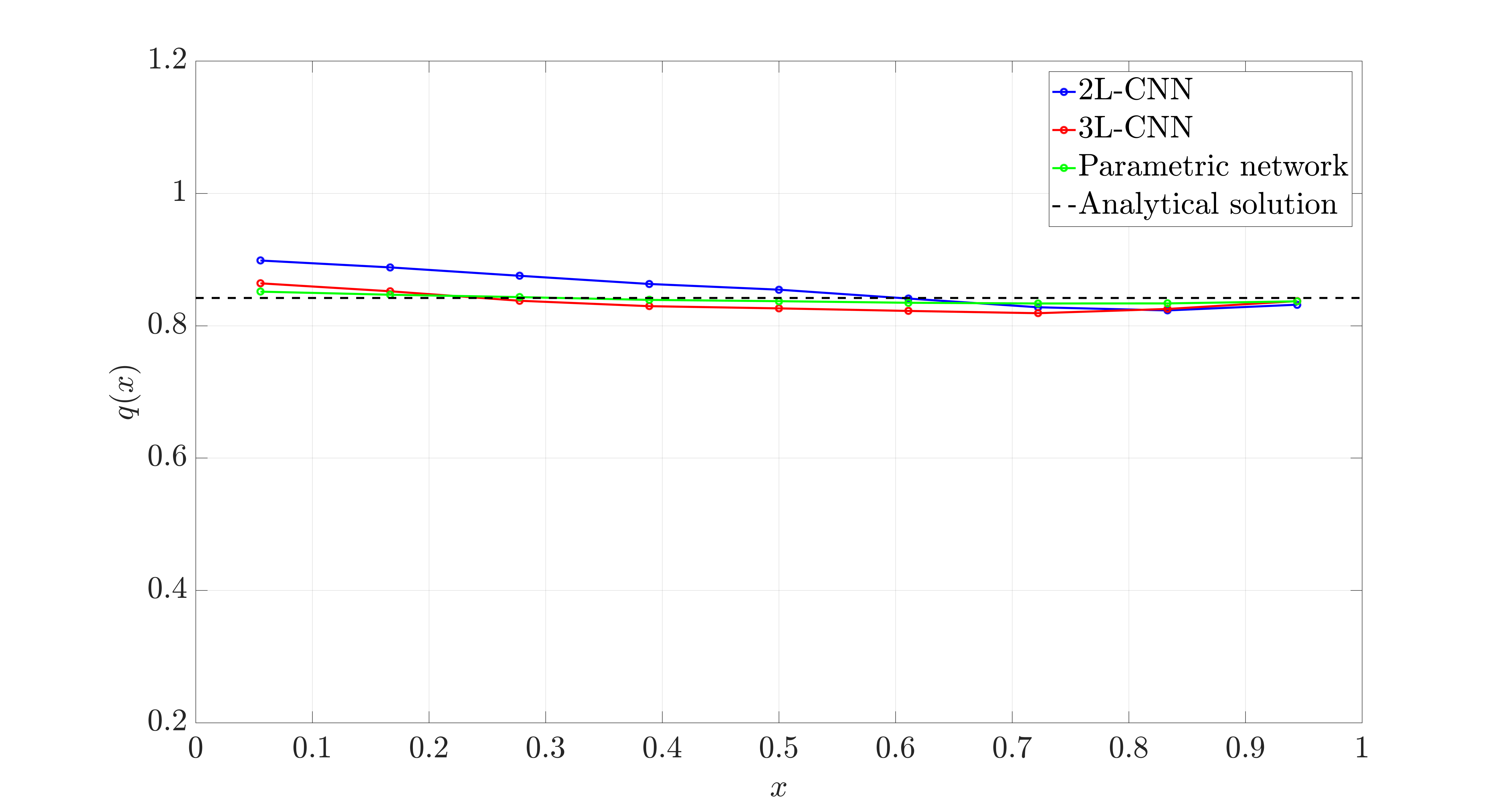}
		\caption{Field $q$.}
		\label{fig::NN_pred_q}
	\end{subfigure} \\
		\begin{subfigure}{.7\textwidth}
		\includegraphics[width=\linewidth]{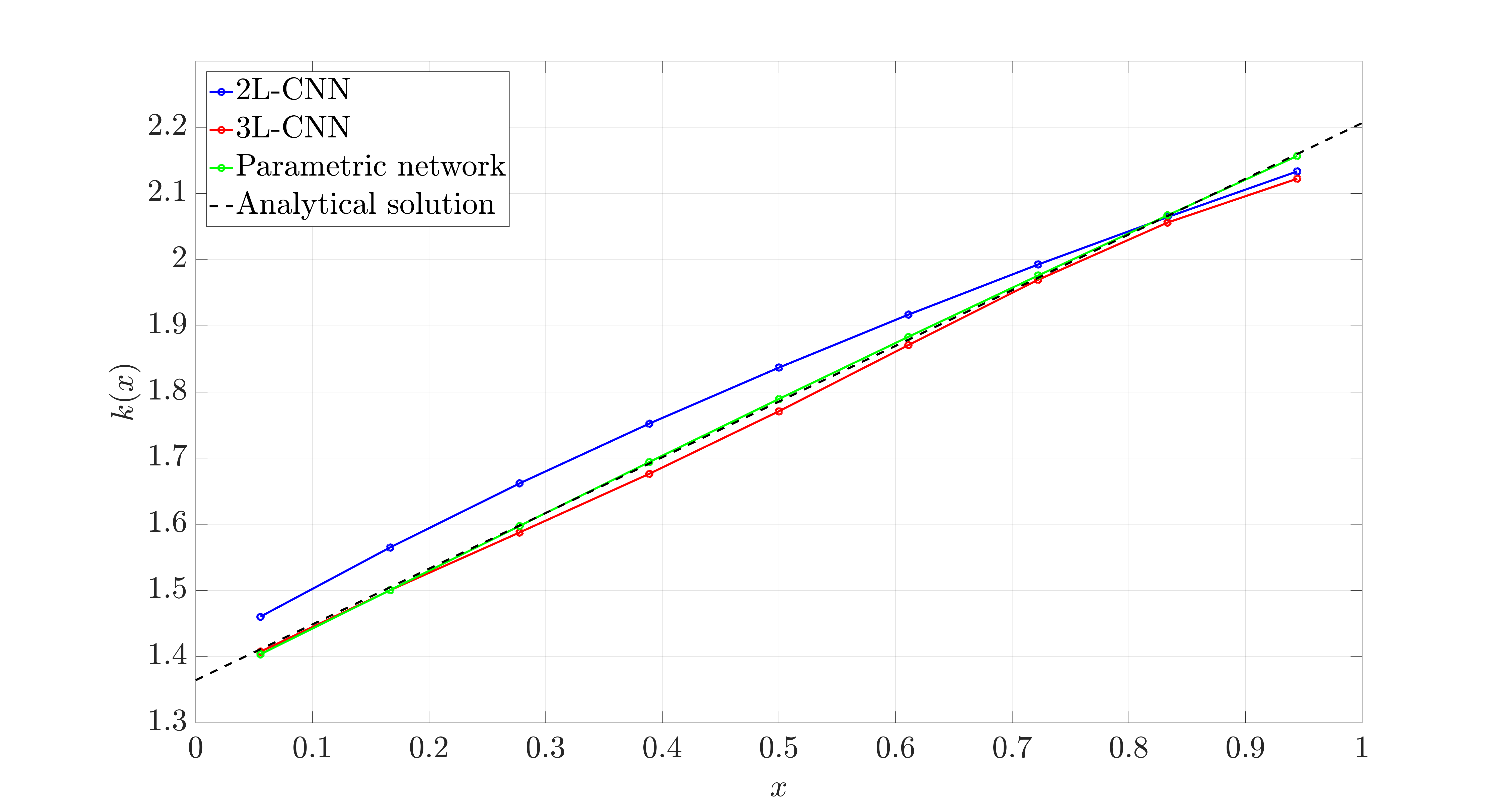}
		\caption{Field $k$.}
		\label{fig::NN_pred_k}
	\end{subfigure}
	\caption{\textbf{PGNNIV prediction of the output fields.} The three neural networks are able to predict the output field $u$ for the exponential diffusivity problem but the predictive capacity varies from one network to another for $q$ and $k$.}
	\label{fig::NN_illustration}
\end{figure}

As in the previous example (homogeneous vs heterogeneous), the performance of the presented methodology in predicting the value of the different fields is now evaluated. There is, however, a difference with respect to that example that is inherent to nonlinear problems: the value of the field $k(x)$ depends now on the boundary conditions, as $k = k(u(x))$ and $u$ depends on the boundary conditions. Fig. \ref{fig::NN_illustration} shows the neural network prediction of all the fields involved in the problem in the hardest case analyzed, that is, exponential diffusivity, for one particular set of boundary conditions ($g_1 = 0.31$ and $g_2 = 0.79$). 

More important than a particular prediction for one set of boundary conditions are the statistics of the errors for the boundary conditions varying in the whole learning space. These statistics are shown in Tables \ref{table::NN_table_L2_1}, \ref{table::NN_table_L2_2} and \ref{table::NN_table_L2_3} for the fields $u$ and $q$ and the three problems, respectively, while the one of the diffusion field $k$ is presented in Table \ref{table::NN_table_L2_4}.

\begin{table}[htbp]
\centering
\begin{tabular}{|l|ccccc|ccccc|}
\hline
                   &     &        & $E^2_r[u]$&        &      &      &       & $E^2_r[q]$  &        &    \\ \hline
                   & min & $Q_1$  & $Q_2$     & $Q_3$  & max  & min  & $Q_1$ & $Q_2$       & $Q_3$  & max \\ \hline
2L-CNN             & $4\cdot 10^{-7}$    & $1\cdot 10^{-6}$        &    $2\cdot 10^{-6}$       & $4\cdot 10^{-6}$       &   $1\cdot 10^{-1}$    & $3\cdot 10^{-5}$     & $3\cdot 10^{-5}$       & $5\cdot 10^{-5}$  & $7\cdot 10^{-5}$        & $3\cdot 10^{-1}$    \\
3L-CNN             & $9\cdot 10^{-7}$    & $5\cdot 10^{-6}$        &    $9\cdot 10^{-6}$       & $2\cdot 10^{-5}$       &   $2\cdot 10^{-1}$    & $4\cdot 10^{-5}$     & $8\cdot 10^{-5}$       & $1\cdot 10^{-4}$  & $2\cdot 10^{-4}$        & $4\cdot 10^{-1}$    \\
Parametric & $3\cdot 10^{-5}$    & $3\cdot 10^{-5}$        &    $4\cdot 10^{-5}$       & $4\cdot 10^{-5}$       &   $1\cdot 10^{-1}$    & $1\cdot 10^{-4}$     & $2\cdot 10^{-4}$       & $4\cdot 10^{-4}$  & $8\cdot 10^{-4}$        & $4\cdot 10^{-1}$    \\ \hline
\end{tabular}
\caption{\textbf{Statistics of the error when predicting the output fields for the problem with constant diffusivity}}
\label{table::NN_table_L2_1}
\end{table}

\begin{table}[htbp]
\centering
\begin{tabular}{|l|ccccc|ccccc|}
\hline
                   &     &        & $E^2_r[u]$&        &      &      &       & $E^2_r[q]$  &        &    \\ \hline
                   & min & $Q_1$  & $Q_2$     & $Q_3$  & max  & min  & $Q_1$ & $Q_2$       & $Q_3$  & max \\ \hline
2L-CNN             & $3\cdot 10^{-4}$    & $3\cdot 10^{-3}$        &    $5\cdot 10^{-3}$       & $8\cdot 10^{-3}$       &   $3\cdot 10^{-1}$    & $5\cdot 10^{-3}$     & $2\cdot 10^{-2}$       & $3\cdot 10^{-2}$  & $4\cdot 10^{-2}$        & $2$    \\
3L-CNN             & $2\cdot 10^{-4}$    & $3\cdot 10^{-3}$        &    $5\cdot 10^{-3}$       & $8\cdot 10^{-3}$       &   $3\cdot 10^{-1}$    & $4\cdot 10^{-3}$     & $2\cdot 10^{-3}$       & $3\cdot 10^{-3}$  & $5\cdot 10^{-2}$        & $2\cdot 10^{1}$    \\
Parametric & $3\cdot 10^{-4}$    & $2\cdot 10^{-3}$        &    $4\cdot 10^{-3}$       & $6\cdot 10^{-3}$       &   $9\cdot 10^{-2}$    & $2\cdot 10^{-3}$     & $1\cdot 10^{-2}$       & $2\cdot 10^{-2}$  & $3\cdot 10^{-2}$        & $2$    \\ \hline
\end{tabular}
\caption{\textbf{Statistics of the error when predicting the output fields for the problem with linear diffusivity}}
\label{table::NN_table_L2_2}
\end{table}

\begin{table}[htbp]
\centering
\begin{tabular}{|l|ccccc|ccccc|}
\hline
                   &     &        & $E^2_r[u]$&        &      &      &       & $E^2_r[q]$  &        &    \\ \hline
                   & min & $Q_1$  & $Q_2$     & $Q_3$  & max  & min  & $Q_1$ & $Q_2$       & $Q_3$  & max \\ \hline
2L-CNN             & $1\cdot 10^{-4}$    & $2\cdot 10^{-3}$        &    $3\cdot 10^{-3}$       & $5\cdot 10^{-3}$       &   $1\cdot 10^{-1}$    & $3\cdot 10^{-3}$     & $2\cdot 10^{-2}$       & $3\cdot 10^{-2}$  & $4\cdot 10^{-2}$        & $7\cdot 10^{-1}$    \\
3L-CNN             & $1\cdot 10^{-4}$    & $2\cdot 10^{-3}$        &    $3\cdot 10^{-3}$       & $5\cdot 10^{-3}$       &   $2\cdot 10^{-1}$    & $4\cdot 10^{-3}$     & $2\cdot 10^{-2}$       & $3\cdot 10^{-2}$  & $5\cdot 10^{-2}$        & $7\cdot 10^{-1}$    \\
Parametric & $1\cdot 10^{-4}$    & $2\cdot 10^{-3}$        &    $4\cdot 10^{-3}$       & $5\cdot 10^{-3}$       &   $2\cdot 10^{-1}$    & $2\cdot 10^{-3}$     & $9\cdot 10^{-3}$       & $2\cdot 10^{-2}$  & $3\cdot 10^{-2}$        & $ 8\times 10^{-1}$    \\ \hline
\end{tabular}
\caption{\textbf{Statistics of the error when predicting the output fields for the problem with exponential diffusivity}}
\label{table::NN_table_L2_3}
\end{table}

\begin{table}[htbp]
\centering
\begin{tabular}{|lc|ccccc|}
\hline
                              &        &     &        &  $E^2_r[k]$ &            &        \\
                              &        & min & $Q_1$  & $Q_2$       & $Q_3$      & max \\ \hline
Constant diffusivity          & 2L-CNN & $3 \cdot 10^{-4}$ & $3 \cdot 10^{-4}$ & $3 \cdot 10^{-4}$ & $4 \cdot 10^{-4}$ & $4\cdot 10^{-4}$   \\
                              & 3L-CNN & $7 \cdot 10^{-7}$ & $4 \cdot 10^{-5}$ & $5 \cdot 10^{-5}$ & $5 \cdot 10^{-5}$ & $2\cdot 10^{-3}$   \\
                              & Parametric & $2 \cdot 10^{-7}$ & $1 \cdot 10^{-5}$ & $2 \cdot 10^{-5}$ & $2 \cdot 10^{-5}$ & $2 \cdot 10^{-5}$   \\ \hline
Linear diffusivity            & 2L-CNN & $4 \cdot 10^{-3}$ & $7 \cdot 10^{-3}$ & $1 \cdot 10^{-2}$ & $2 \cdot 10^{-2}$ & $\infty$   \\
                              & 3L-CNN & $1 \cdot 10^{-3}$ & $9 \cdot 10^{-3}$ & $1 \cdot 10^{-2}$ & $2 \cdot 10^{-2}$ & $\infty$   \\
                              & Parametric & $1 \cdot 10^{-3}$ & $6 \cdot 10^{-3}$ & $1 \cdot 10^{-2}$ & $2 \cdot 10^{-2}$ & $\infty$   \\ \hline
Exponential diffusivity         & 2L-CNN & $6 \cdot 10^{-4}$ & $2 \cdot 10^{-2}$ & $3 \cdot 10^{-2}$ & $3 \cdot 10^{-2}$ & $8 \cdot 10^{-2}$   \\
                              & 3L-CNN & $6 \cdot 10^{-4}$ & $1 \cdot 10^{-2}$ & $2 \cdot 10^{-2}$ & $5 \cdot 10^{-2}$ & $2 \times 10^{-1}$   \\
                              & Parametric & $6 \cdot 10^{-4}$ & $4 \cdot 10^{-3}$ & $7 \cdot 10^{-3}$ & $1 \cdot 10^{-2}$ & $7 \cdot 10^{-2}$ \\ \hline
\end{tabular}
\caption{\textbf{Statistics of the error when predicting the diffusivity field for the three problems.}}
\label{table::NN_table_L2_4}
\end{table}

\begin{figure}[htbp]
	\centering
	\begin{subfigure}{.82\textwidth}
		\includegraphics[width=\linewidth]{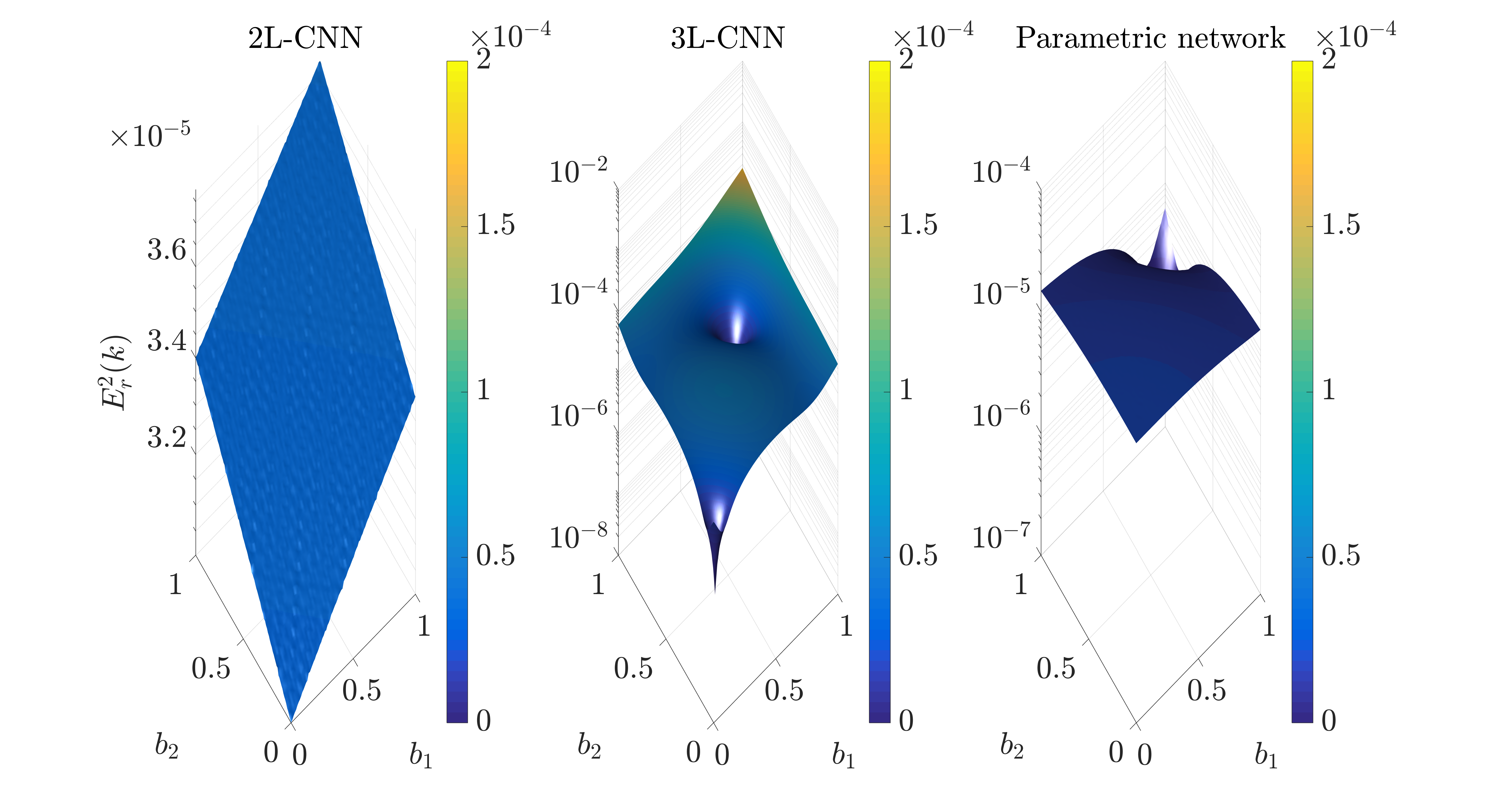}
		\caption{Constant diffusivity.}
		\label{fig::NN_k_D1}
	\end{subfigure}\\
	\begin{subfigure}{.82\textwidth}
		\includegraphics[width=\linewidth]{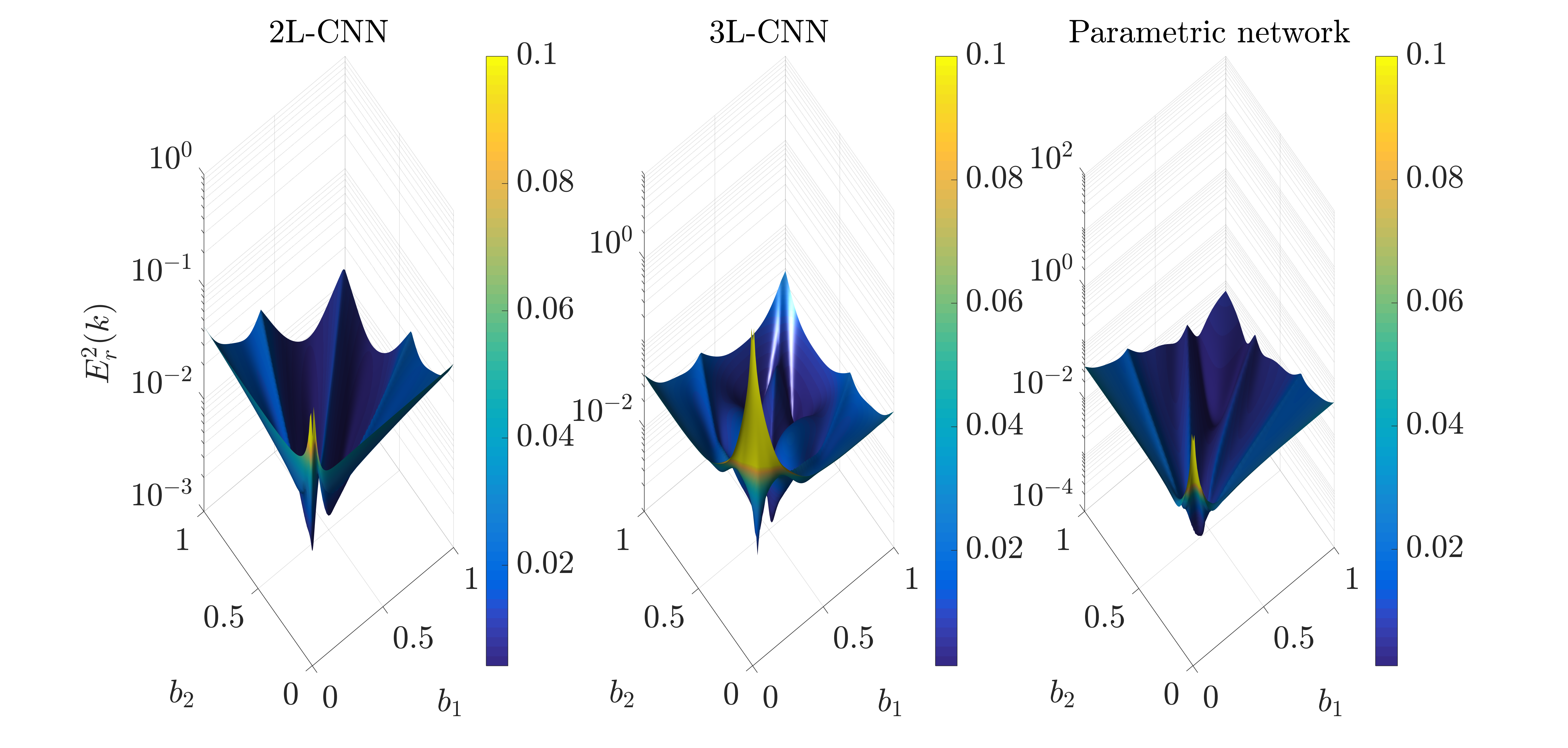}
		\caption{Linear diffusivity.}
		\label{fig::NN_k_D2}
	\end{subfigure} \\
		\begin{subfigure}{.82\textwidth}
		\includegraphics[width=\linewidth]{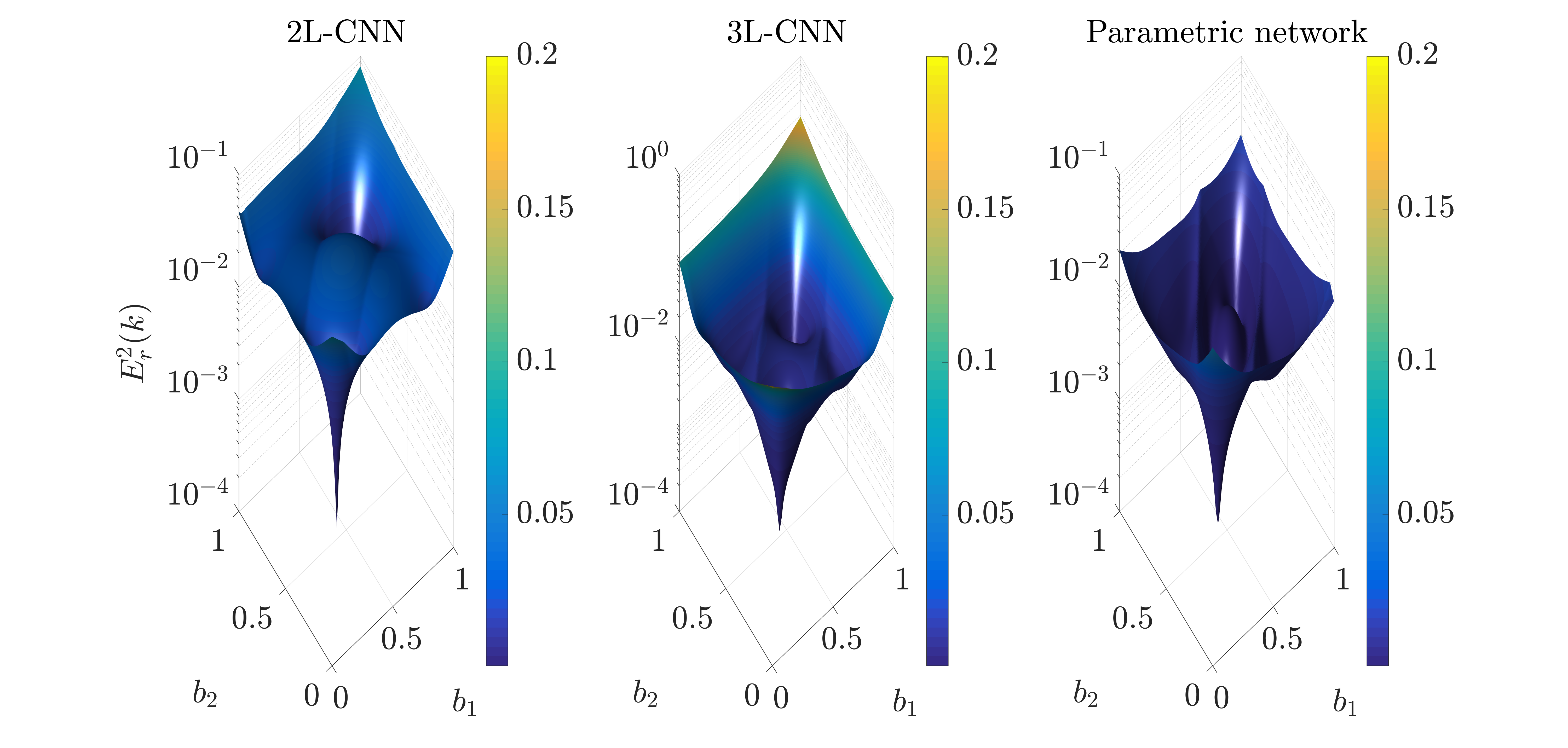}
		\caption{Exponential diffusivity.}
		\label{fig::NN_k_D3}
	\end{subfigure}
	\caption{\textbf{PGNNIV error in predicting the field $\bs{k}$.} All PGNNIV are able to estimate the value of the field $k$ for the constant and linear diffusivity problems. Although the 2L-CNN network yields small enough errors for the exponential case, the 3L-CNN and the parametric networks provide more accurate results.}
	\label{fig::NN_k}
\end{figure}

In addition to the error associated to the fields $u$ and $q$, it is possible to evaluate the error of the field $k$ for the different boundary conditions. Fig. \ref{fig::NN_k} shows the relative $L_2$ error for the three tested networks and the three datasets. 

From the figures and tables presented, we can draw several important observations. For the first problem, the three PGNNIV have a good accuracy, though the 2L-PGNNIV, whose learning power is specific for linear models, provides the best results in terms of errors. Besides, as the problem is linear, the error has a linear shape when visualized in terms of the boundary conditions. The 3L-PGNNIV and the parametric PGNNIV are nonlinear models and so it is the error. For the second problem, the three PGNNIV estimate accurately the value of the field $k$ except for some values close to the boundaries of the sampling space. The three networks are useful therefore as model learners. Finally, for the third problem, we observe that, although the error is low in general for the three models, the 3L-CNN and the parametric network achieves smaller values. That is because the first network has less predictive power than the second for a general class of functions, while, for that case, a parametric model of the form $k(u) = \alpha + \beta u^\gamma$ is able to describe the underlying physics accurately enough. Note that this, however, is particular to the problem in hands, that is, $k(u)=\exp(u)$. For all cases, the highest errors appear always close to the boundaries of the learning domain. If we compare only the second and third PGNNIV, the third one reaches better predictions close to the boundaries. This is a consequence of the multi-parametric nature of neural networks when used in regression problems: certain overfitting is unavoidable to some extent that is glimpsed especially close to the boundary values.

\subsubsection{Unraveling capacity}

\begin{figure}
	\centering
	\begin{subfigure}{.7\textwidth}
		\includegraphics[width=\linewidth]{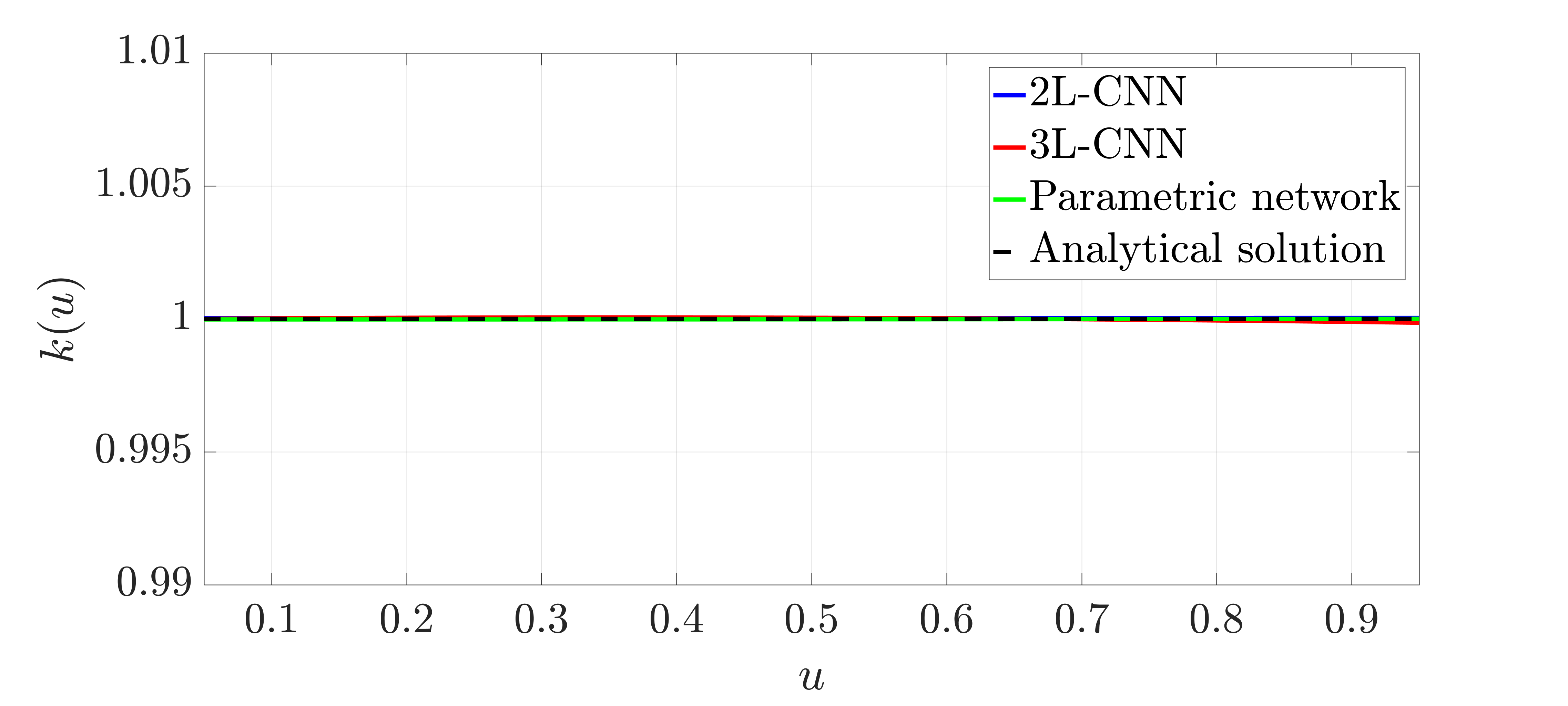}
		\caption{Constant diffusivity.}
		\label{fig::NN_model_1}
	\end{subfigure}\\
	\begin{subfigure}{.7\textwidth}
		\includegraphics[width=\linewidth]{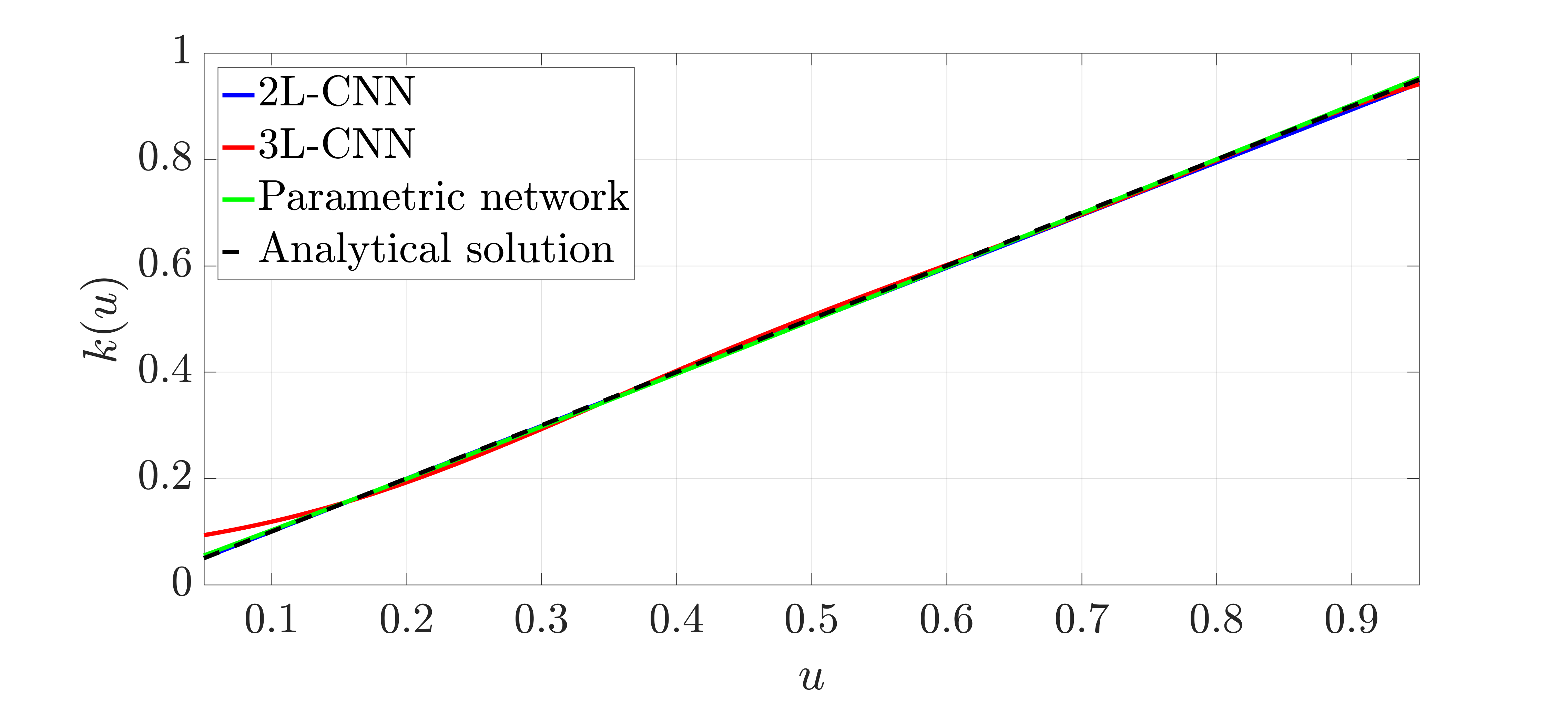}
		\caption{Linear diffusivity.}
		\label{fig::NN_model_2}
	\end{subfigure} \\
		\begin{subfigure}{.7\textwidth}
		\includegraphics[width=\linewidth]{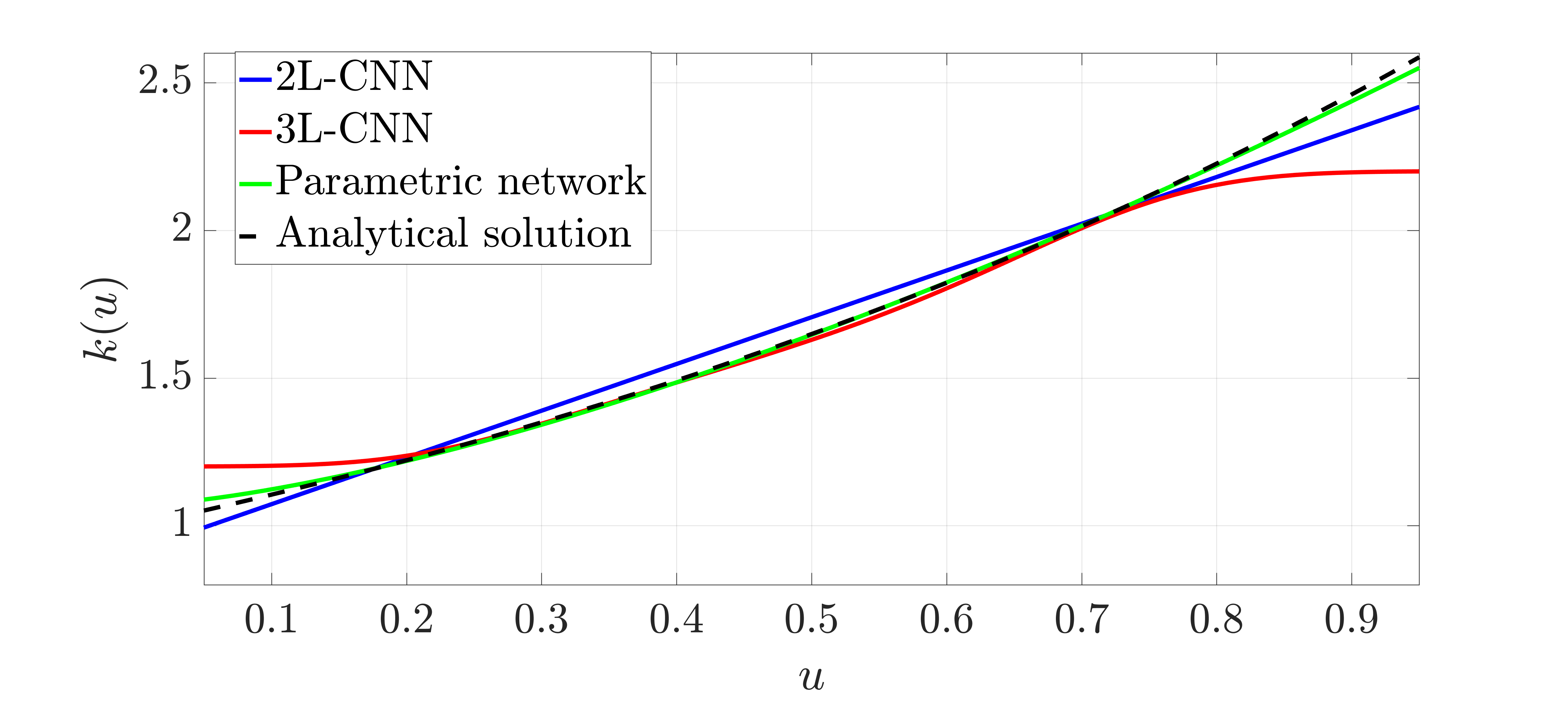}
		\caption{Exponential diffusivity.}
		\label{fig::NN_model_3}
	\end{subfigure}
	\caption{\textbf{Unraveling capacity of the PGNNIV.} The three presented PGNNIV try to explain the constitutive model $k = k(u)$. Dataset coverage leads to sampled values of $\mathtt{um}$ in $[0.13,0.87]$ which explains the differences obtained beyond those values.}
	\label{fig::NN_model}
\end{figure}

As explained in Section \ref{sec::problem_description}, PGNNIVs have both predictive and unraveling capacity. This has been explored in the precedent example by reproducing the field $k = k(x)$. This field is a direct output of the problem, but when expressed in terms of the variable $x$ may be seen as an explanation (identification) of the heterogeneous constitutive model. For nonlinear problems, however, the problem becomes richer. The interest here is to learn the model $k = k(u)$. One way is to sample the two output fields $k(x)$ and $u(x)$ for each $x$ and all boundary conditions. This leads however to a point cloud due to the noise and discretization errors. But there is one more elegant alternative that consists on exporting the network related to the model. Its convolutional nature makes it independent of the considered point $x$ (that is, element independent). Fig. \ref{fig::NN_model} shows the model predictions for the three datasets and the three tested PGNNIV. It is important to emphasize once again that the parametric network has good learning capacity since $1.06+1.60u^{1.47}$ is a good enough approximation of $\exp(u)$,  although this characteristic is specific to the problem in hands.

\clearpage

\section{Numerical experiments} \label{sec::numerical_experiments}

Next, we evaluate the performance of the methodology presented according to different parameters, inherent to the presented methodology: the dataset size, the error in the training dataset and the size of the hidden layers. The following discussion corresponds to the more complex exponential diffusivity problem using the 3L-CNN network.

\subsection{Dataset size}

\begin{figure}[htbp]
\centering
\includegraphics[width=0.8\linewidth]{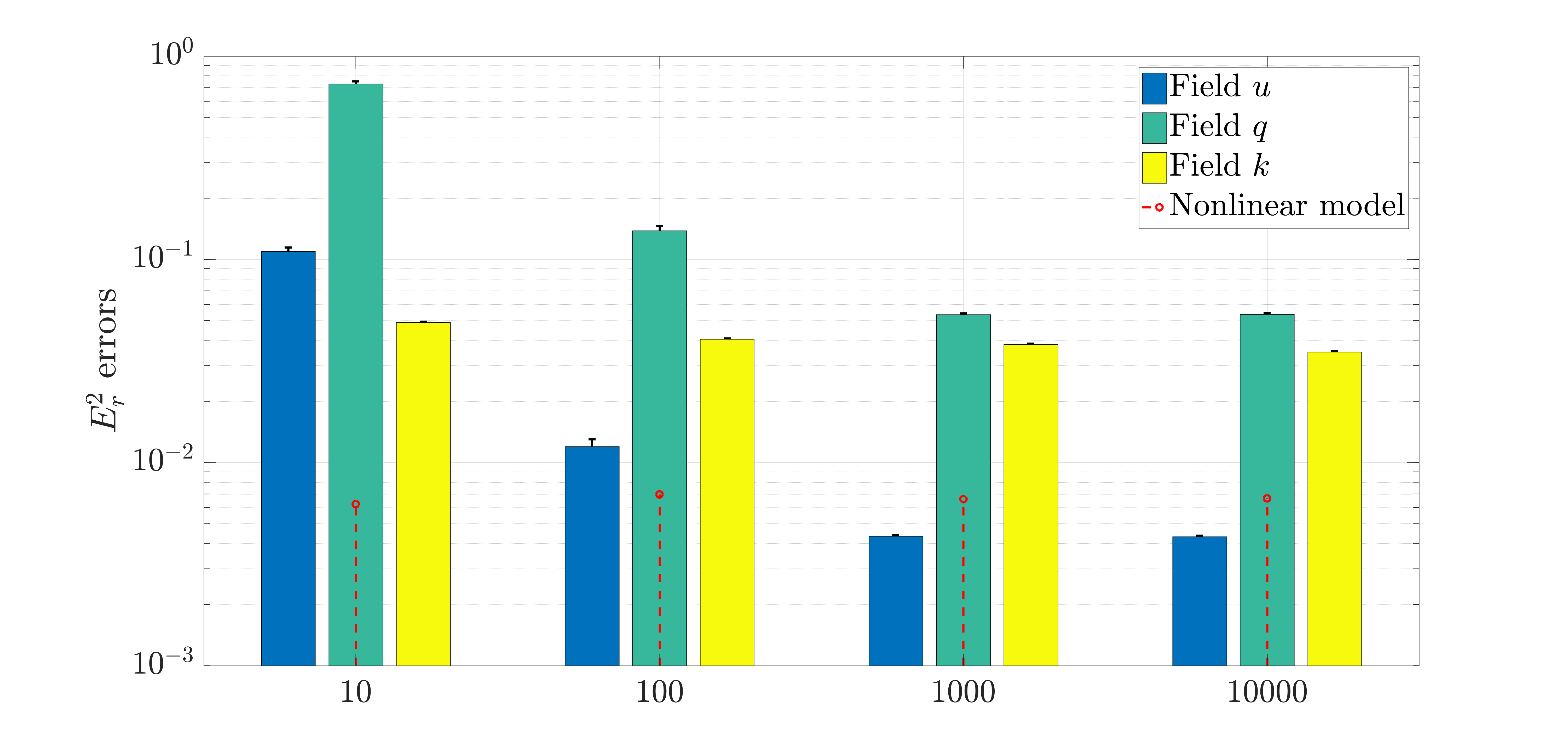}
\caption{\textbf{Impact of the dataset size on the predictive and unraveling capacity of the network.} The relative $E_r^2$ is shown for different dataset sizes. As the prediction of the fields depends on the boundary conditions, the error is shown with its associated standard error bar.}
\label{fig::NE_Dataset_error}
\end{figure}

Fig. \ref{fig::NE_Dataset_error} shows the $E^2_r$ errors when varying the dataset size, both considering the predictive and unraveling capacity of the network. The main conclusion is that the dataset size has an important impact on accuracy and precision, but not much on the model learning capacity. This result is expected since for each sample of the dataset, the model learning is performed at the nodal level, so the learning capacity is amplified as a consequence of the discretization. This is even more evident when analyzing the spatial error defined for a spatial field $f$ as:

\begin{equation}
    \varepsilon_r[f](x) =\left|\frac{\tilde{f}(x)-f(x)}{f(x)}\right|
\end{equation}

\begin{figure}[htbp]
\centering
\includegraphics[width=\linewidth]{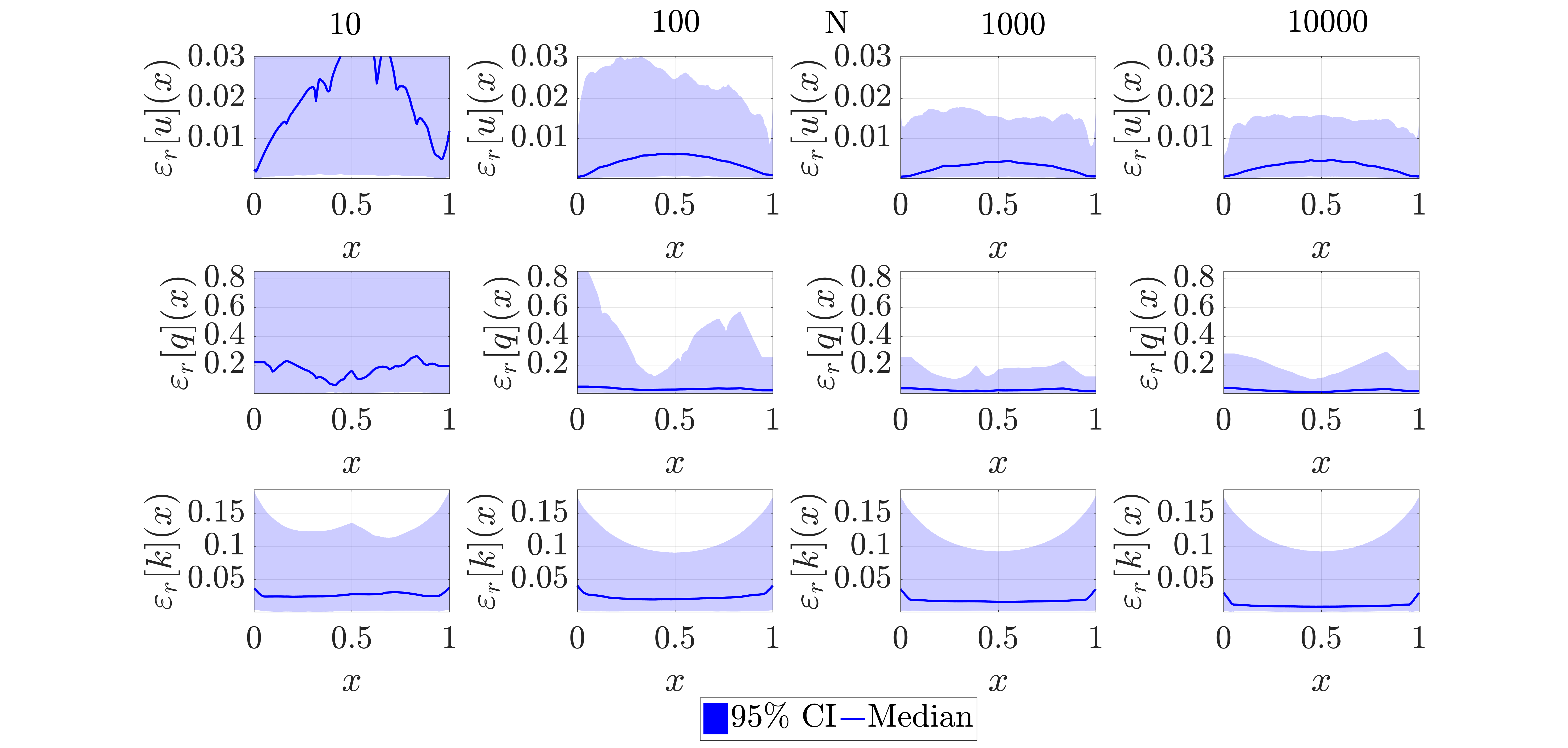}
\caption{\textbf{Impact of the dataset size on the spatial field prediction.} As the different fields involveds depend on boundary conditions, the median of the relative error is shown together with a $95\%$ confident interval.}
\label{fig::NE_Dataset_space}
\end{figure}

This is illustrated in Fig. \ref{fig::NE_Dataset_space}: even if for all fields, a larger dataset implies better estimations, the field $k$ is the less data demanding to be learned.

\subsection{Noise impact}

\begin{figure}[htbp]
\centering
\includegraphics[width=0.8\linewidth]{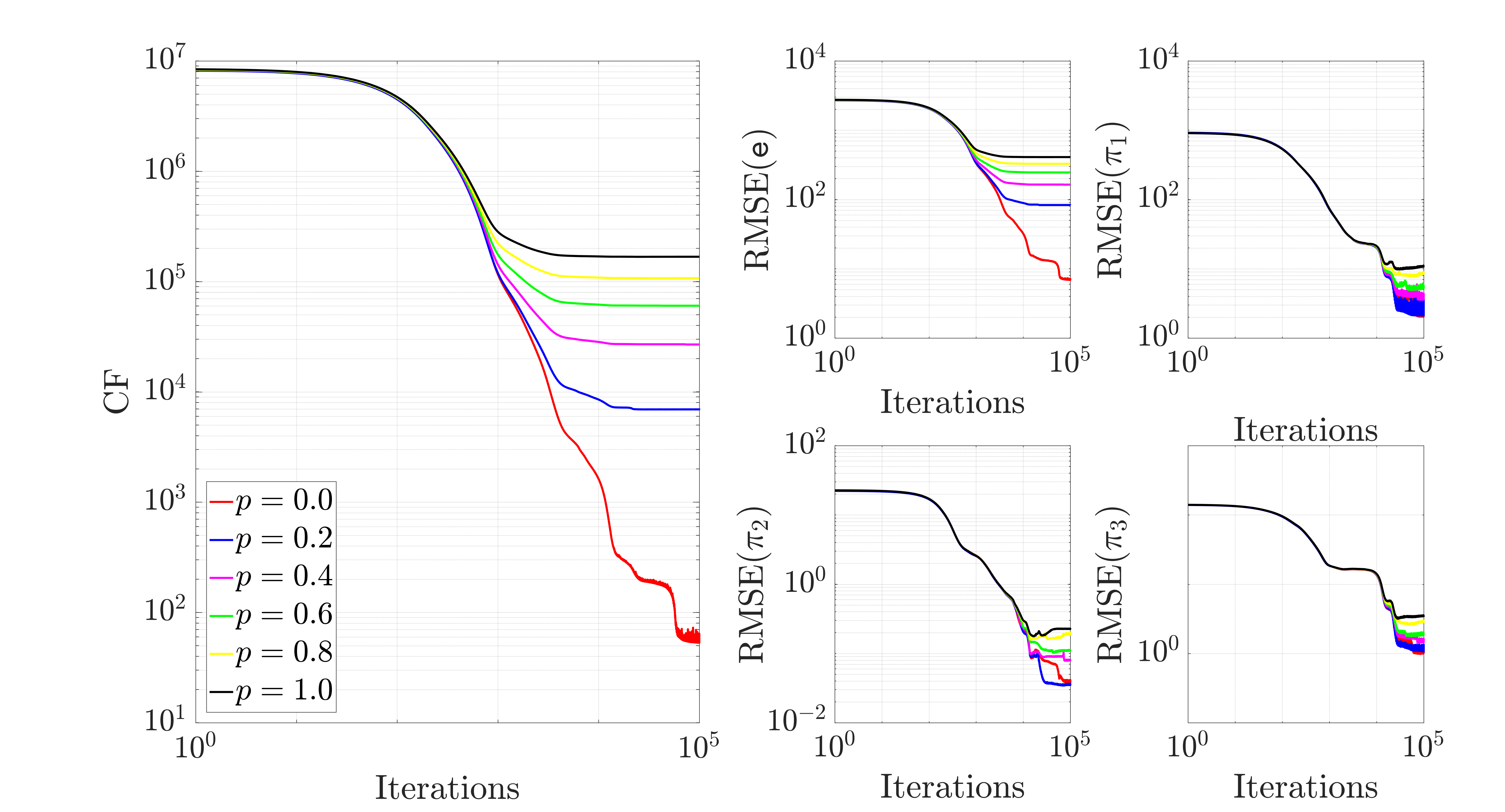}
\caption{\textbf{Impact of the noise level on the network convergence.} The CF plotted at the left panel is the squared sum of the terms shown at the right panel.}
\label{fig::NE_Noise_conv}
\end{figure}

To analyze the impact of the data noise on the results we added a white noise to the training data. The noise level is evaluated by introducing a white noise, proportional to the standard deviation of spatial profile. That is, $u^n_i \sim \mathcal{N}(u_i,s)$ where $u_i$ is the noise-free nodal values, $u^n_i$ is their noisy counterparts, and $s = p \sigma$, being $\sigma$ the standard deviation of $\{u_i\}_{i=1,\ldots,n}$. First, we analyze the noise impact on the network convergence. Fig. \ref{fig::NE_Noise_conv} shows the CF evolution as well as the different penalty terms during the optimization process. As the noise affects directly the output field $u$, the differences in the convergence of the CF function are associated mainly to the prediction error term associated with $u$, already in the 1000-th iteration. However, once the output field noise has been filtered by the PGNNIV, there is another error source associated with the numerical discretization, which plays an important role in later stages of convergence, around $M=2 \times 10^4$.

\begin{figure}[htbp]
\centering
\includegraphics[width=0.8\linewidth]{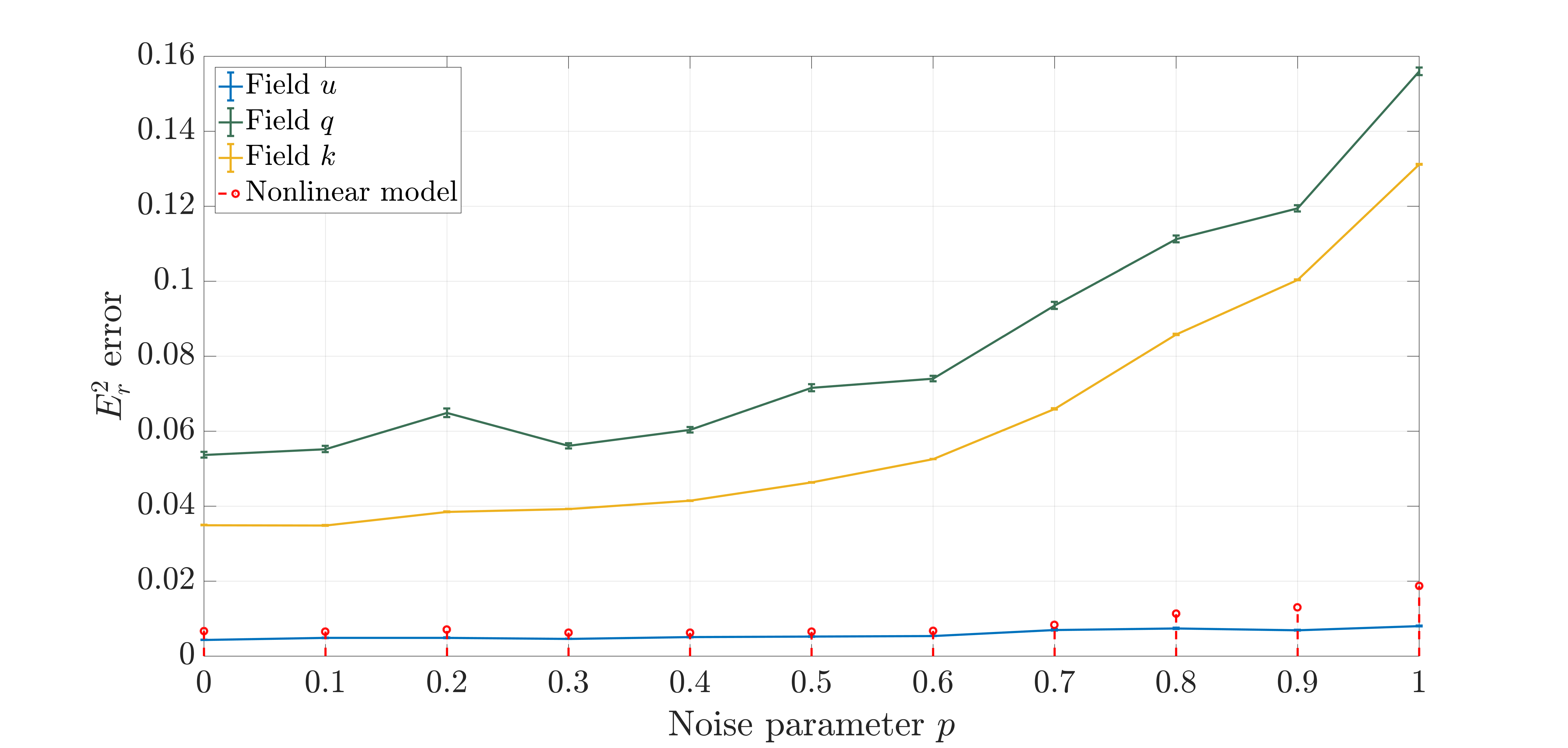}
\caption{\textbf{Impact of the noise level on the predictive and unraveling capacity of the network.} The relative $E_r^2$ is shown for different dataset sizes. As the prediction of the fields depends on the boundary conditions, the error is shown with its associated standard error bar.}
\label{fig::NE_Noise_error}
\end{figure}

With respect to the impact of the noise in the accuracy and precision of the network, we analyze now the impact of the noise in the $E^2_r$ errors, both for predictive and unraveling capacity. The results are shown in Fig. \ref{fig::NE_Noise_error}. The interpretation is similar to the one for the dataset impact. However, there is one subtlety: the fields $k$ and $q$ are associated with the derivatives of the field $u$ so even if the filtering capacity of the network is remarkable for the prediction of all fields, as it is common when using PGNNIV methodology \cite{ayensajimenez2020identification}, the error in the prediction is higher for the fields involving a derivative of the discretized function, as the error is amplified by numerical discretization. Fig. \ref{fig::NE_Dataset_space} shows the spatial errors, illustrating this fact more clearly.

\begin{figure}[htbp]
\centering
\includegraphics[width=\linewidth]{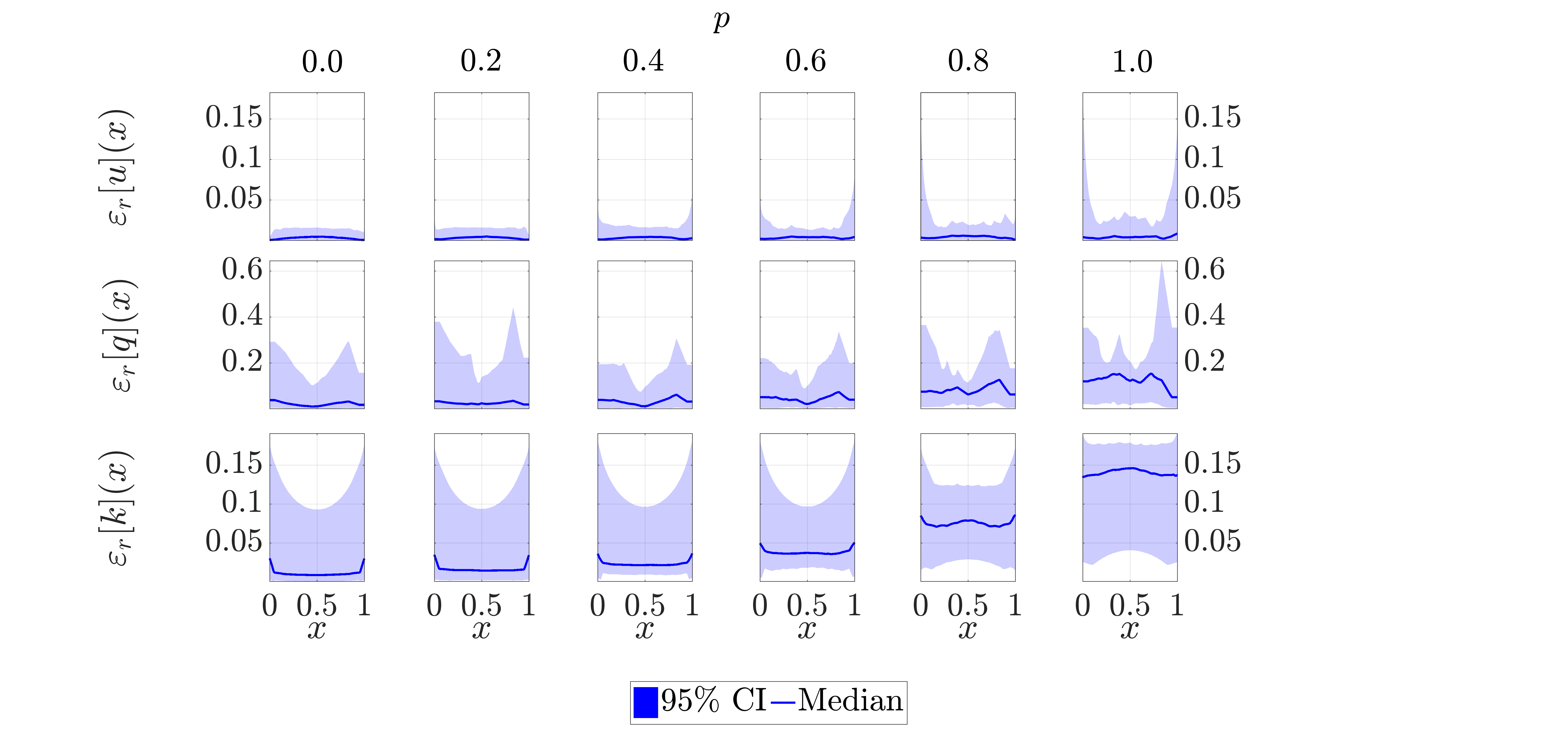}
\caption{\textbf{Impact of the noise level on the spatial field prediction.} As the different fields involved depend on boundary conditions, the median of the relative error together with a $95\%$ confidence interval is shown.}
\label{fig::NE_Noise_space}
\end{figure}

\subsection{Learning space size}

\begin{figure}[htbp]
\centering
\includegraphics[width=0.8\linewidth]{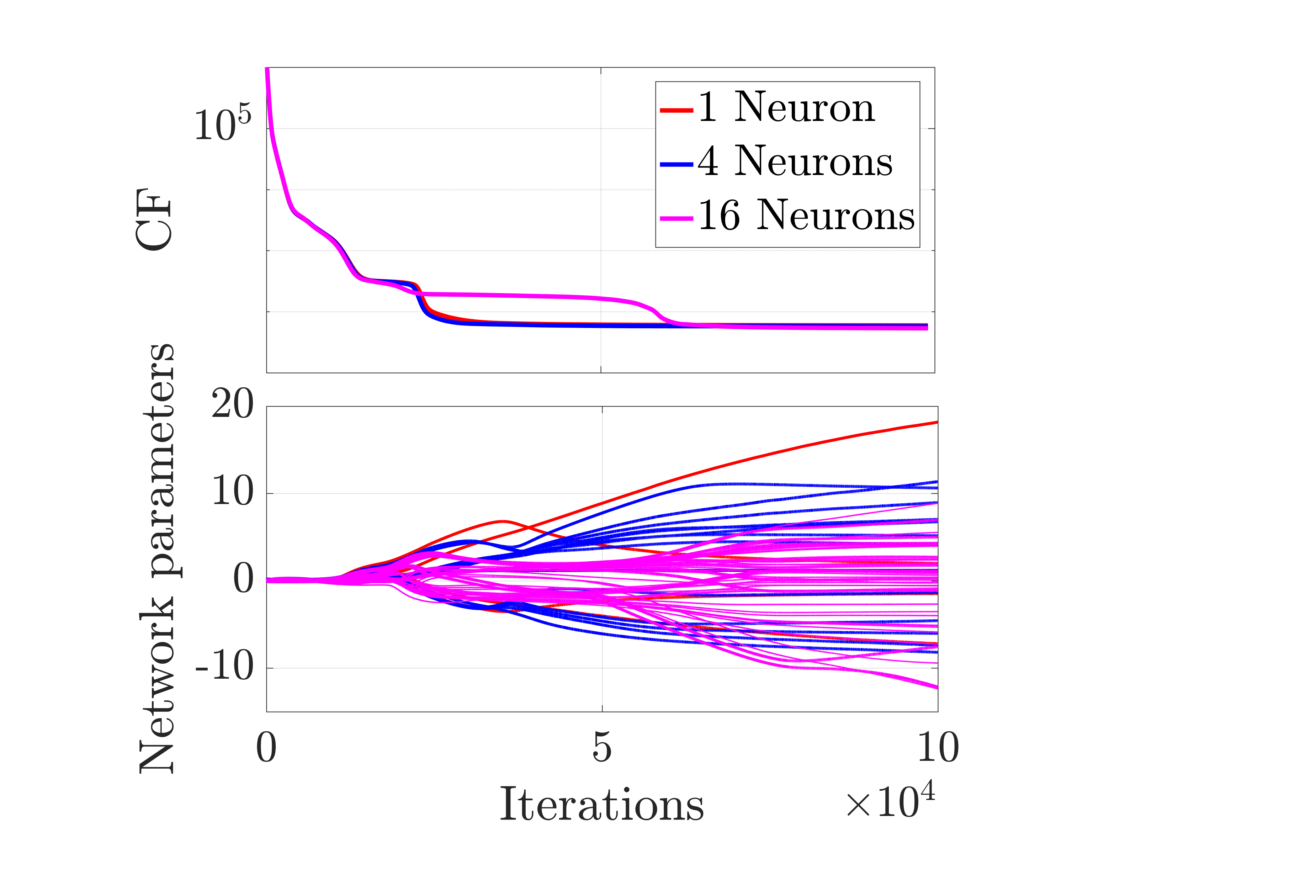}
\caption{\textbf{Impact of the learning space size on the network convergence.} The upper pannel shows the value of the CF and the lower pannel the value of the weights at each iteration. CF values are smoothed using a constant filter of bandwidth $w=1000$ for avoiding sharp oscillations in the visualization in the logarithmic scale. Note that a sharp change in the value of the parameters is associated with a significative change in the value of CF.}
\label{fig::NE_Neurons_conv}
\end{figure}

Finally, we evaluate the effect of varying the learning space associated with the model. The first comment corresponds to the convergence of the network, which is shown in Fig. \ref{fig::NE_Neurons_conv}: the larger the learning space (i.e. the higher is the number of neurons in the network), the higher the converge cost.

\begin{figure}
	\centering
	\begin{subfigure}{.6\textwidth}
		\includegraphics[width=\linewidth]{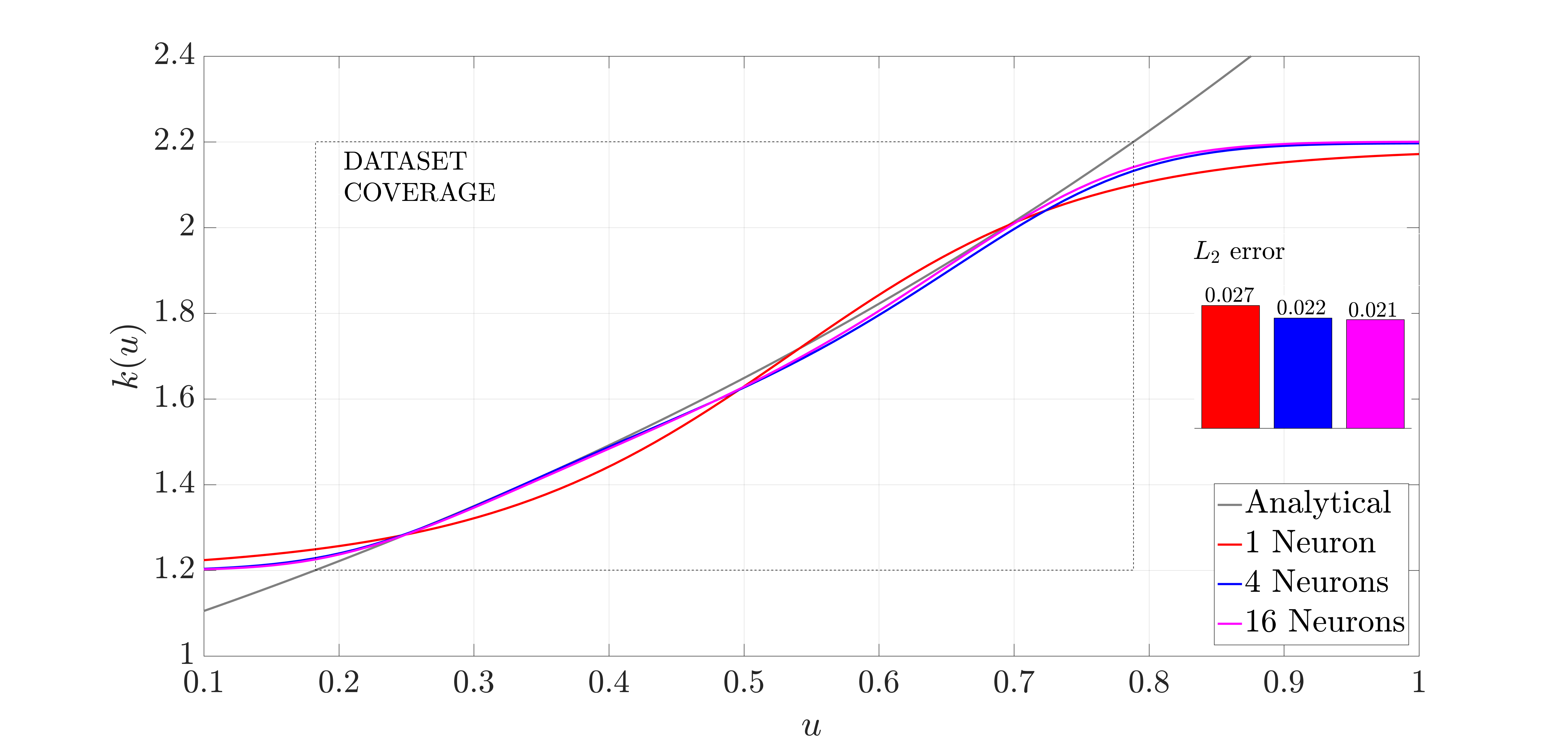}
		\caption{Predictive capacity.}
		\label{fig::NE_Neurons_error_1}
	\end{subfigure}\\
	\begin{subfigure}{.6\textwidth}
		\includegraphics[width=\linewidth]{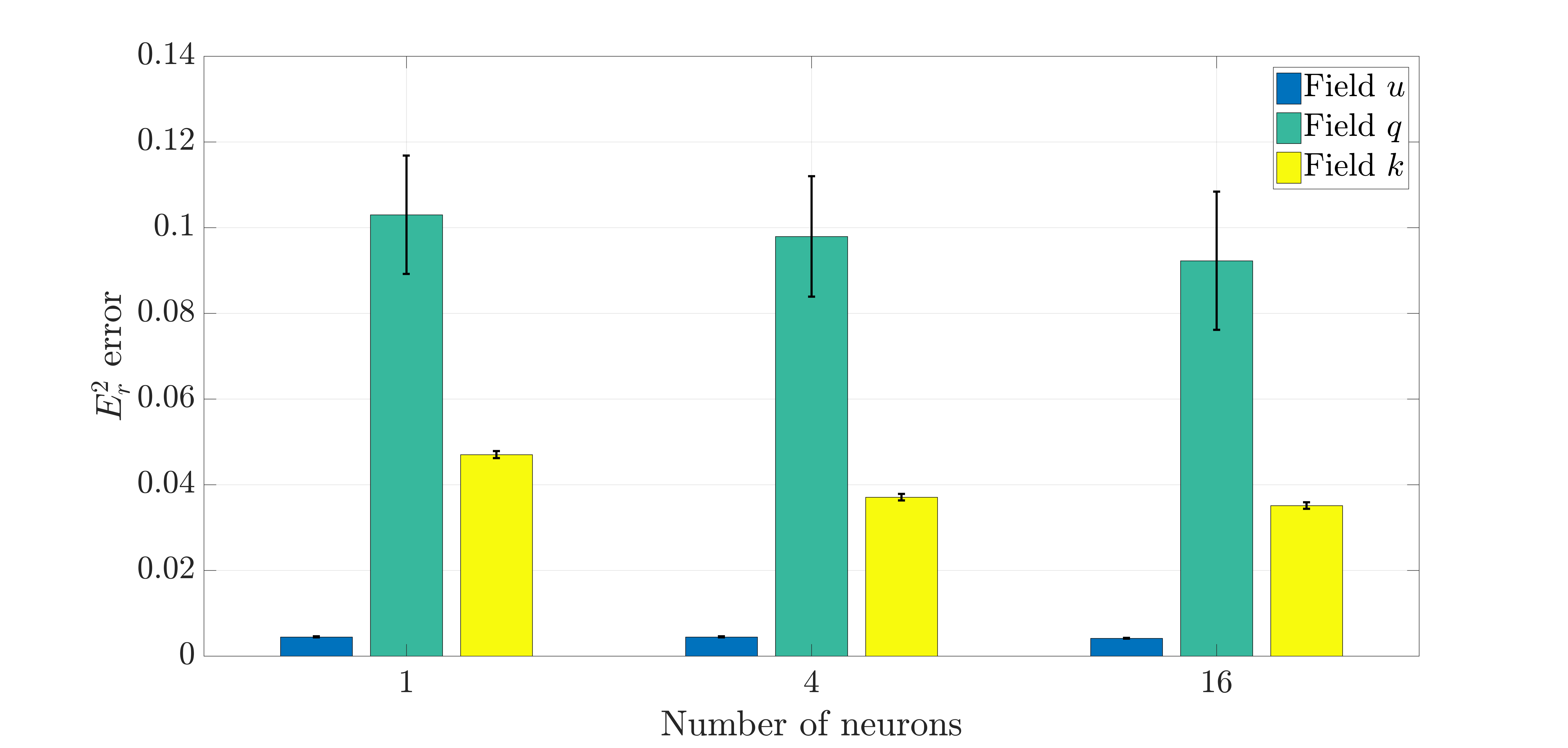}
		\caption{Unraveling capacity.}
		\label{fig::NE_Neurons_error_2}
	\end{subfigure} \\
	\caption{\textbf{Predictive and unraveling capacity of the PGNNIV.} The unraveling capacity is evaluated using the $L^2$ error instead of the relative $L^2$ error since it is not reported in the same plot.}
	\label{fig::NE_Neurons_error}
\end{figure}

\begin{figure}
	\centering
	\begin{subfigure}{.6\textwidth}
		\includegraphics[width=\linewidth]{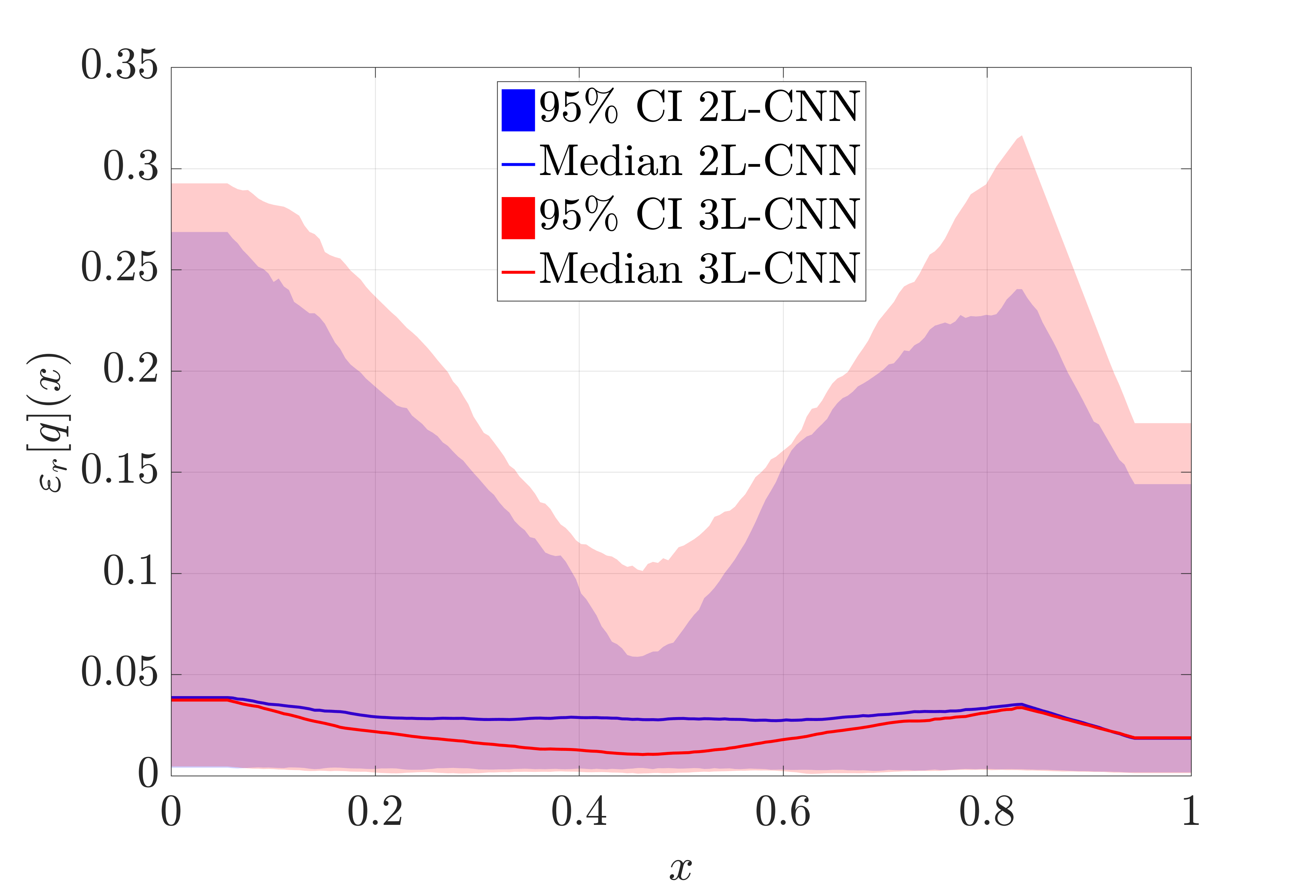}
		\caption{Field $q$.}
		\label{fig::NE_Neurons_space_2}
	\end{subfigure}\\
	\begin{subfigure}{.6\textwidth}
		\includegraphics[width=\linewidth]{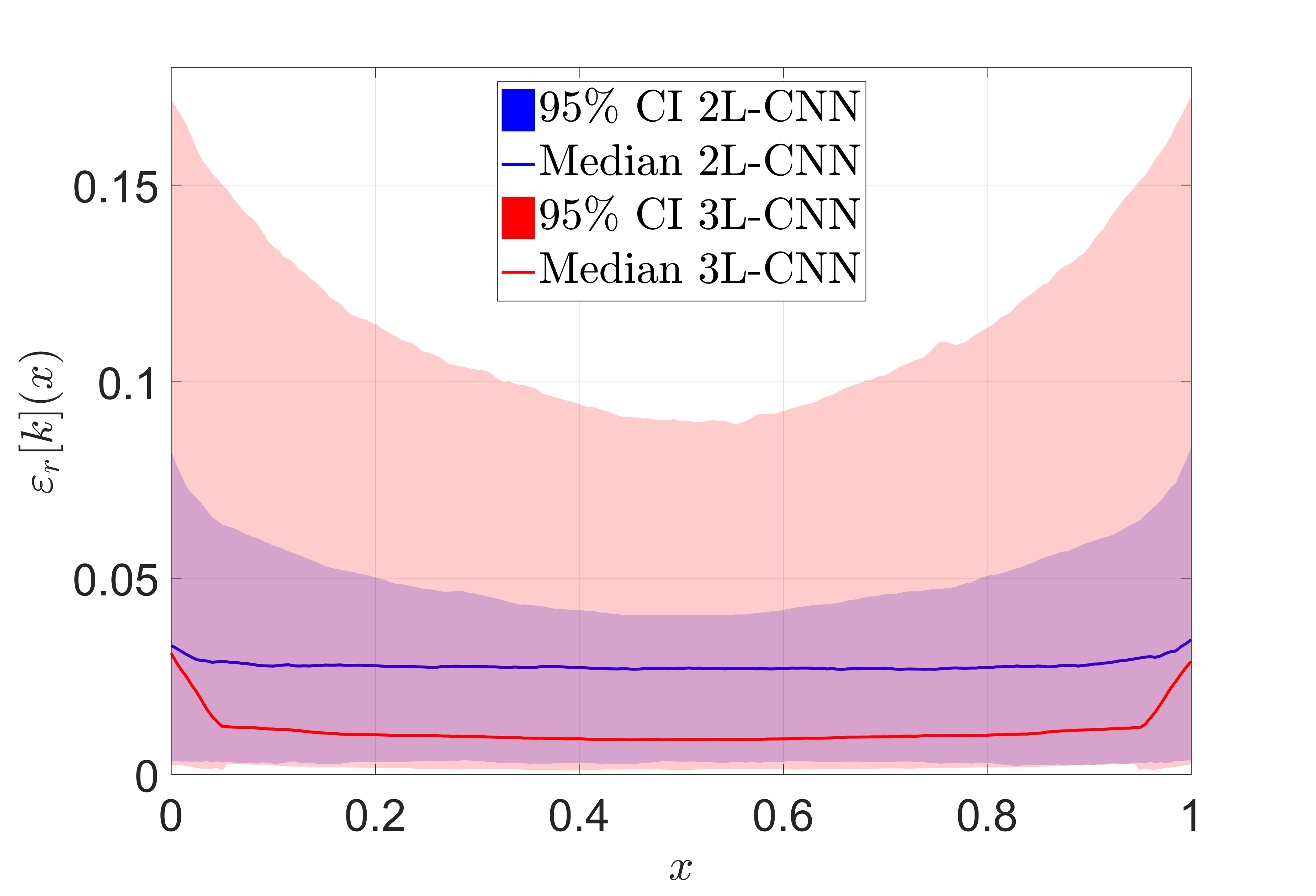}
		\caption{Field $k$.}
		\label{fig::NE_Neurons_space_3}
	\end{subfigure} \\
	\caption{\textbf{Spatial distribution of the error for $q$ and $k$ fields and 2L-CNN and 3L-CNN PGNNIV.} As the different fields involved depend on boundary conditions, the median of the relative error is shown together with a $95\%$ confidence interval.}
	\label{fig::NE_Neurons_space}
\end{figure}

The predictive and unraveling capacity of the network is shown in Fig. \ref{fig::NE_Neurons_error}. As it is common in the neural network framework, the learning power of the model increases with the model parameters (number of neurons) until it reaches a stagnation point, beyond which the accuracy does not improve. It is important however to remark that an augmentation of the learning space is always related with an average accuracy improvement but not necessarily with a precision improvement for all specific problems. Indeed, if the learning space is large, the model risks being overfitted, resulting in poor predictions for some special cases. This may be seen when comparing the error bar for the different predictions. To illustrate this fact, Fig. \ref{fig::NE_Neurons_space} plots the spatial distributions of the errors of fields $q$ and $k$ (the ones involved in the model learning). The results are in agreement with our conclusion: lower values of median estimations and larger confident intervals.

In summary, an increase of the number of neurons is a good starting point for reaching good enough results, but it is a rude strategy when the goal is to fit the model precisely and avoiding overfitting. For these purposes, better and more suited strategies are required \cite{hou2017convnets}.

\clearpage

\section{Discussion}

The presented framework introduces a new and singular way of combining of physical knowledge and the power of the most recent data science techniques to solve problems in continuum physics. Although the presented illustrative problems are simple and academic, they highlight all the ingredients and the main features of the methodology. We summarize the following achievements:
\begin{itemize}
    \item The capacity of dealing with arbitrary complex models, equations and structures. We have considered nonlinearities and different degrees of material knowledge such as spatial symmetry. Also, locality has been implicitly considered when establishing a local relation between the flow and the essential field.
    \item Flexibility to add some or the whole available physical knowledge to the network. Several degrees of knowledge have been tested, involving symmetries, the mathematical character of the constitutive operators and their explicit parametric dependence.
    \item The two-stages character of the methodology. In the first step, the predictive and unraveling capacities of the methods are clearly revealed. This process is computationally expensive. However, nowadays, there are a lot of resources to perform this task, such as scalability and parallelization tools, cloud and distributed computing and adapted hardware technologies such as graphical processor units (GPU) and, tensorial processor units (TPUs) and Field Programmable Gate Arrays (FPGAs). The second step is a pure evaluation of the output for the desired input. Primary and derived fields and parameters are obtained in evaluation cost. Only, the post process is related to the interpolation of the nodal values to get the different fields in the whole domain.. As an example, Fig. \ref{fig::NE_Neurons_space} was generated using a $100 \times 100$ grid, so it includes the resolution of $10^4$ nonlinear PDEs. It was however generated in less than one second using a personal laptop as it involves only $10^4$ network evaluations for each model.
\end{itemize}

Nonetheless, there is an important limitation that arises from the present study. It is crucial to correcly choose  the best strategy when selecting what is known and what is not in the problem to solve. If something is known, the best strategy is to include it in the PGNNIV, explicitly if possible, or implicitly if not. In the limit case, parametric models are the best ones and less expensive to train. So they are the ones that provide the best results, when correctly assumed. Indeed, centuries dedicated to establish models cannot be wrong. This may drive to think that in that limit case nothing new has been presented, but this is simplistic since the model learning and the predictive capacity are acquired in one stroke and once for all and can be continuously improved by new sets of data input (Dynamic Data Driven Applications Systems, DDDAS). Then, any prediction may be performed later via offline calculation in one single evaluation. This is extremely advantageous for optimization, inverse problems or stochastic computations based on Montecarlo strategies, among others. In particular, we may establish the whole constitutive model, which is equivalent to build a response surface using NN procedures or to derive a Reduced Order Model using backpropagation, with the particularity that the computational cost is reduced due to the constraints with respect to usual NN procedures \cite{ayensajimenez2020identification}.

\begin{figure}[htbp!]
\centering
\includegraphics[width=1.0\linewidth]{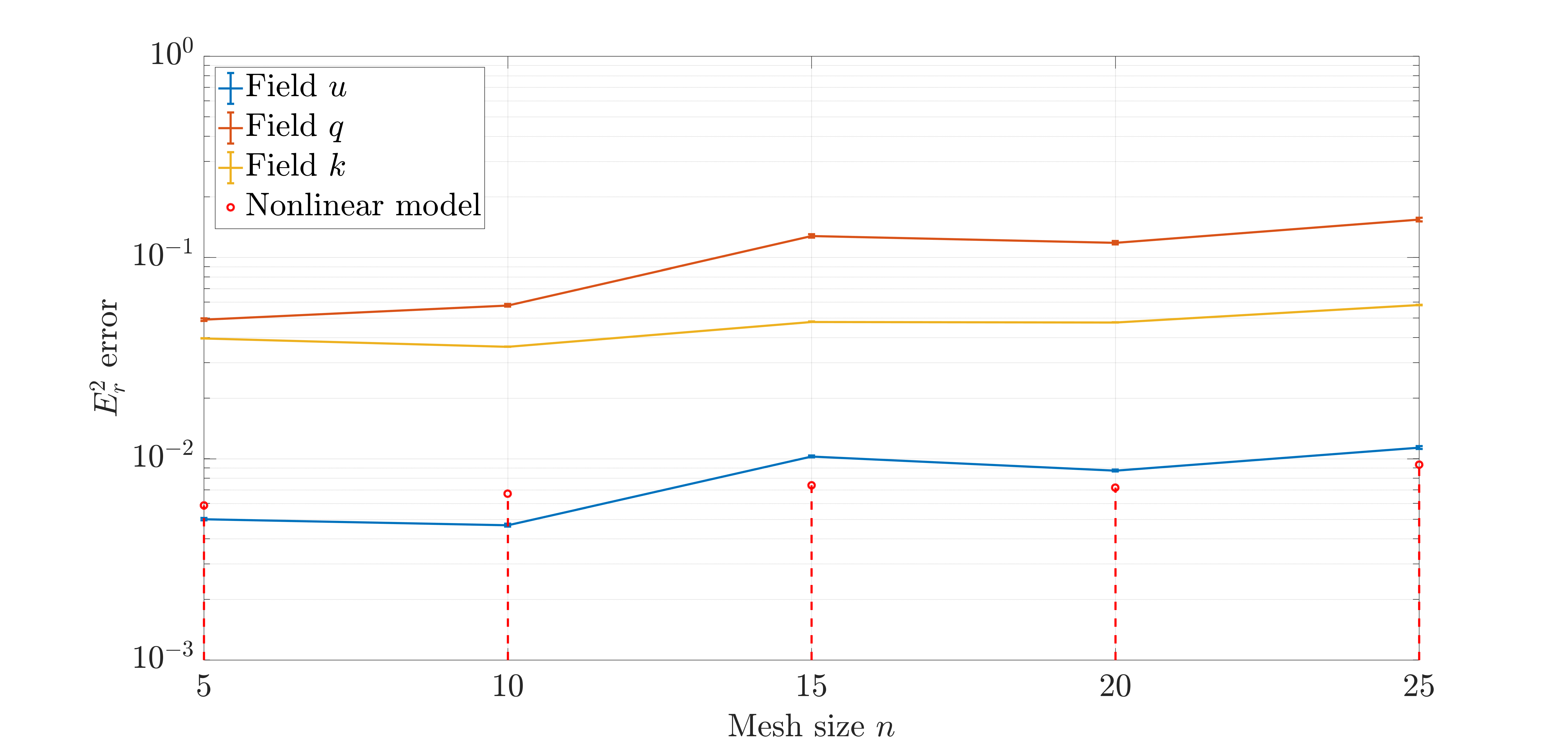}
\caption{\textbf{Effect of the discretization in the predictive and unraveling results.} Note that a mesh refinement is not necessarily accompanied by an improvement in the results.}
\label{fig::Discretizationv}
\end{figure}

Finally, two last important remarks should be done about the limitations of the presented methodology:

\begin{itemize}
    \item The first one is the obvious fact that the methodology is based on the availability of enough data. Data quantity but mainly data quality is required: large amounts of data are not enough, but they have also to be well distributed and uncorrelated for the model to be accurately learned. This is more complicated to be thought, since there is not a simple way to guarantee that internal variables have a pertinent coverage (see for instance Fig. \ref{fig::NE_Neurons_error}) when the constitutive relation is not \emph{a priori} known.
    \item A rough increase of the discretization (that is, to augment the number of nodes or to reduce the mesh size) does not necessarily improve the prediction or unraveling capacity. This is a fundamental difference between this methodology and the usual simulation approach (that is merely predictive). Convergence with the mesh is a very critical aspect, as shown in Fig. \ref{fig::Discretizationv}. Here we point out the need of new research results in this line, related to the mathematical structure of a broad range of problems. In a certain sense, we still suffer from the lack of mathematical results, playing the role that, for instance, convergence theorems play in Finite Element analysis.
\end{itemize}

As future work we point in two directions:

\begin{itemize}
\item The methodology is new and lacks of sound theoretical results that justify its application for general classes of problems. Indeed, the performance of the trained PGNNIV may be evaluated by analyzing the value of the cost function for the test data, or, even better, the value of each of the loss terms, as each one is associated with one physical aspect of the problem in hands (e.g. mass, momentum or energy conservation, etc.). Although the predictive capacity is directly assessable by evaluating the corresponding loss term of the cost function, it is not the case for the unraveling capacity: if input and output data are not rich enough or the physical added content is insufficient, the network may have good predictive capacity while misleading the constitutive state model. Consider, for example, the case of an elastic clamped beam subjected to axial load. If the displacement is measured only at the loaded node, many possible stiffness ($K(x) = E(x)A(x)/L$) distributions along the beam are possible resulting in the same displacement associated with a given load. The problem is solved: (i) by monitoring the whole displacement field or (ii) by explicitly imposing stiffness homogeneity. This fundamental problem, which we call the slack problem, requires further research, as now its treatment is based on the modeler's previous knowledge and intuition about the problem.
\item As the methodology is based on a TensorFlow reframing of mathematical problems, it is especially well suited for easy and structured mesh discretizations. An octree-based formulation is one of the common strategies to face complex geometries, but this would, in principle, greatly increase the input and output size of the problem. These problems also aggravates when going to higher dimensions in what is called the curse of dimensionality \cite{poggio2017and}. It is therefore important to think of strategies to bypass this difficulty.
\end{itemize}

\section{Conclusions}
 In the present work, we introduce a general framework for the analysis of problems in continuum physics from a data science perspective that incorporates the fundamental physical laws in the computations. We have taken advantage of the characteristics of a very recent idea, the so-called Physically-Guided Neural Networks with Internal Variables, for a very general formulation of a broad class of problems whose physical content is expressed by means of a system of partial differential equations. The key point is a cunning splitting of the equations of the problem into two sets, what is known and what is not, and therefore you want it to be learned. The result is a method that offers both predictive and unraveling capacity: it is possible to predict in a single evaluation the state of the system for any prescribed value of the input variable, and to learn about the mathematical structure of the constitutive state equation of the problem in hands. Indeed, good results have been demonstrated for some paradigmatic cases in science and engineering related to constitutive modeling: local character, heterogeneity and nonlinearity.
 
 Several numerical experiments have been carried out to illustrate some of the characteristics of the method, in particular those related to the dataset size and noise, and the size of the model networks. The results show the trends and features that are common to machine learning techniques, in particular, the importance of the dataset, the filtering capacity and the problem of overfitting. Nevertheless, the numerical experiments and examples shown in this work should be reinforced by further research exploring the mathematical structure of the PGNNIV problem, depending on the particular selection of the topology for $\mathsf{Y}$ (ROM) and $\mathsf{H}$ (constitutive state equation) subnetworks, as well as the prescription of the known functions $\bs{I}$, $\bs{O}$ and $\bs{R}$ in Eq. (\ref{eq::optimization}). In any case, PGNNIV are very promising for both predicting and unraveling problems formulated in terms of partial differential equations, ubiquitous in sciences and engineering.

\section*{Acknowledgements}
The authors gratefully acknowledge the financial support from the Spanish Ministry of Economy and Competitiveness (MINECO) and FEDER, UE through the project PGC2018-097257-B-C31, the Spanish Ministry of Science and Innovation through the project PID2019-106099RB-C44/AEI/10.13039/501100011033, the Government of Aragon (DGA) through the grant T24\_17R and the Centro de Investigacion Biomedica en Red en Bioingenieria, Biomateriales y Nanomedicina (CIBER-BBN). CIBER-BBN is Financed by the Instituto de Salud Carlos III with assistance from the European Regional Development Fund.








\bibliographystyle{unsrt}  
\bibliography{references}  

\begin{thebibliography}{10}

\bibitem{atzori2010}
L.~Atzori, A.~Iera, and G.~Morabito.
\newblock The internet of things: A survey.
\newblock {\em Computer Networks}, 54(15):2787--2805, 2010.

\bibitem{manyika2011}
J.~Manyika, M.~Chui, B.~Brown, J.~Bughin, R.~Dobbs, C.~Roxburgh, and
  A.~Hung-Byers.
\newblock {\em Big data: The next frontier for innovation, competition, and
  productivity}.
\newblock McKinsey Global Institute Reports, 2011.

\bibitem{maggiora1992computational}
Gerald~M Maggiora, David~W Elrod, and Robert~G Trenary.
\newblock Computational neural networks as model-free mapping devices.
\newblock {\em Journal of chemical information and computer sciences},
  32(6):732--741, 1992.

\bibitem{stulp2015many}
Freek Stulp and Olivier Sigaud.
\newblock Many regression algorithms, one unified model: A review.
\newblock {\em Neural Networks}, 69:60--79, 2015.

\bibitem{hoffmann2019benchmarking}
Frank Hoffmann, Torsten Bertram, Ralf Mikut, Markus Reischl, and Oliver Nelles.
\newblock Benchmarking in classification and regression.
\newblock {\em Wiley Interdisciplinary Reviews: Data Mining and Knowledge
  Discovery}, 9(5):e1318, 2019.

\bibitem{Hill2006}
Shawndra Hill, Foster Provost, and Chris Volinsky.
\newblock Network-based marketing: Identifying likely adopters via consumer
  networks.
\newblock {\em Statistical Science}, 21(2):256--276, 2006.

\bibitem{Aneshensel2013}
Carol~S. Aneshensel.
\newblock {\em Theory-based data analysis for the social sciences}.
\newblock SAGE Publications, Inc., 2013.

\bibitem{Raghupathi2014}
Wullianallur Raghupathi and Raghupathi Viju.
\newblock Network-based marketing: Identifying likely adopters via consumer
  networks.
\newblock {\em Health information science and systems}, 2:3--7, 2014.

\bibitem{Lopez-Moreno2014}
Ignacio Lopez-Moreno, Javier Gonzalez-Dominguez, Oldrich Plchot, David
  Martinez, Joaquin Gonzalez-Rodriguez, and Pedro Moreno.
\newblock Automatic language identification using deep neural networks.
\newblock {\em Proceeding ICASSP, 2014, Proceeding of the IEEE International
  Conference on Acoustic, Speech and Signal Processing}, pages 5337--5341,
  2014.

\bibitem{Krizhevsky2012}
Alex Krizhevsky, Ilya Sutskever, and Geoffrey~E. Hinton.
\newblock Imagenet classification with deep convolutional neural networks.
\newblock pages 1097--1105, 2012.

\bibitem{purwins2019deep}
Hendrik Purwins, Bo~Li, Tuomas Virtanen, Jan Schl{\"u}ter, Shuo-Yiin Chang, and
  Tara Sainath.
\newblock Deep learning for audio signal processing.
\newblock {\em IEEE Journal of Selected Topics in Signal Processing},
  13(2):206--219, 2019.

\bibitem{minto2018}
W.~Minto.
\newblock {\em Logic, inductive and deductive}.
\newblock Alpha Editions, 2018.

\bibitem{LeCun2019}
Y.C. LeCun.
\newblock Deep learning hardware: Past, present, and future.
\newblock pages 12--19, 2019.

\bibitem{gulli2017deep}
Antonio Gulli and Sujit Pal.
\newblock {\em Deep learning with Keras}.
\newblock Packt Publishing Ltd, 2017.

\bibitem{geron2019hands}
Aur{\'e}lien G{\'e}ron.
\newblock {\em Hands-on machine learning with Scikit-Learn, Keras, and
  TensorFlow: Concepts, tools, and techniques to build intelligent systems}.
\newblock O'Reilly Media, 2019.

\bibitem{bergstra2011theano}
James Bergstra, Fr{\'e}d{\'e}ric Bastien, Olivier Breuleux, Pascal Lamblin,
  Razvan Pascanu, Olivier Delalleau, Guillaume Desjardins, David Warde-Farley,
  Ian Goodfellow, Arnaud Bergeron, et~al.
\newblock Theano: Deep learning on gpus with python.
\newblock In {\em NIPS 2011, BigLearning Workshop, Granada, Spain}, volume~3,
  pages 1--48. Citeseer, 2011.

\bibitem{bastien2012theano}
Fr{\'e}d{\'e}ric Bastien, Pascal Lamblin, Razvan Pascanu, James Bergstra, Ian
  Goodfellow, Arnaud Bergeron, Nicolas Bouchard, David Warde-Farley, and Yoshua
  Bengio.
\newblock Theano: new features and speed improvements.
\newblock {\em arXiv preprint arXiv:1211.5590}, 2012.

\bibitem{paszke2017automatic}
Adam Paszke, Sam Gross, Soumith Chintala, Gregory Chanan, Edward Yang, Zachary
  DeVito, Zeming Lin, Alban Desmaison, Luca Antiga, and Adam Lerer.
\newblock Automatic differentiation in pytorch.
\newblock 2017.

\bibitem{paszke2019pytorch}
Adam Paszke, Sam Gross, Francisco Massa, Adam Lerer, James Bradbury, Gregory
  Chanan, Trevor Killeen, Zeming Lin, Natalia Gimelshein, Luca Antiga, et~al.
\newblock Pytorch: An imperative style, high-performance deep learning library.
\newblock In {\em Advances in neural information processing systems}, pages
  8026--8037, 2019.

\bibitem{berry2011}
D.M. Berry.
\newblock The computational turn: Thinking about the digital humanities.
\newblock {\em Culture Machine}, 12, 2011.

\bibitem{gould1981}
P.~Gould.
\newblock Letting the data speak for themselves.
\newblock {\em Annals of the Association of American Geographers},
  71(2):166--176, 1981.

\bibitem{kitchin2014big}
Rob Kitchin.
\newblock Big data, new epistemologies and paradigm shifts.
\newblock {\em Big data \& society}, 1(1):2053951714528481, 2014.

\bibitem{kitchin2014}
R.~Kitchin.
\newblock Big data and human geography: Opportunities, challenges and risks.
\newblock {\em Dialogues in Human Geography}, 3:262--267, 12 2013.

\bibitem{Xue2019}
Ying Xue.
\newblock An overview of overfitting and its solutions.
\newblock {\em J. Phys. Conf. Ser}, 1168:022022, 6 2019.

\bibitem{xu2019explainable}
Feiyu Xu, Hans Uszkoreit, Yangzhou Du, Wei Fan, Dongyan Zhao, and Jun Zhu.
\newblock Explainable ai: A brief survey on history, research areas, approaches
  and challenges.
\newblock In {\em CCF International Conference on Natural Language Processing
  and Chinese Computing}, pages 563--574. Springer, 2019.

\bibitem{ayensa2019unsupervised}
Jacobo Ayensa-Jim{\'e}nez, Mohamed~H Doweidar, Jose~A Sanz-Herrera, and Manuel
  Doblar{\'e}.
\newblock An unsupervised data completion method for physically-based
  data-driven models.
\newblock {\em Computer Methods in Applied Mechanics and Engineering},
  344:120--143, 2019.

\bibitem{karpatne2017theory}
Anuj Karpatne, Gowtham Atluri, James~H Faghmous, Michael Steinbach, Arindam
  Banerjee, Auroop Ganguly, Shashi Shekhar, Nagiza Samatova, and Vipin Kumar.
\newblock Theory-guided data science: A new paradigm for scientific discovery
  from data.
\newblock {\em IEEE Transactions on knowledge and data engineering},
  29(10):2318--2331, 2017.

\bibitem{raissi2017physics}
Maziar Raissi, Paris Perdikaris, and George~Em Karniadakis.
\newblock Physics informed deep learning (part i): Data-driven solutions of
  nonlinear partial differential equations.
\newblock {\em arXiv preprint arXiv:1711.10561}, 2017.

\bibitem{raissi2019physics}
Maziar Raissi, Paris Perdikaris, and George~E Karniadakis.
\newblock Physics-informed neural networks: A deep learning framework for
  solving forward and inverse problems involving nonlinear partial differential
  equations.
\newblock {\em Journal of Computational Physics}, 378:686--707, 2019.

\bibitem{li2019combination}
Xiang Li, Ziming Yan, and Zhanli Liu.
\newblock Combination and application of machine learning and computational
  mechanics.
\newblock {\em Chinese Science Bulletin}, 64(7):635--648, 2019.

\bibitem{darema2004dynamic}
Frederica Darema.
\newblock Dynamic data driven applications systems: A new paradigm for
  application simulations and measurements.
\newblock In {\em International Conference on Computational Science}, pages
  662--669. Springer, 2004.

\bibitem{peherstorfer2015dynamic}
Benjamin Peherstorfer and Karen Willcox.
\newblock Dynamic data-driven reduced-order models.
\newblock {\em Computer Methods in Applied Mechanics and Engineering},
  291:21--41, 2015.

\bibitem{kirchdoerfer2016data}
Trenton Kirchdoerfer and Michael Ortiz.
\newblock Data-driven computational mechanics.
\newblock {\em Computer Methods in Applied Mechanics and Engineering},
  304:81--101, 2016.

\bibitem{kutz2016dynamic}
J~Nathan Kutz, Steven~L Brunton, Bingni~W Brunton, and Joshua~L Proctor.
\newblock {\em Dynamic mode decomposition: data-driven modeling of complex
  systems}.
\newblock SIAM, 2016.

\bibitem{ayensa2018new}
Jacobo Ayensa-Jim{\'e}nez, Mohamed~H Doweidar, Jose~A Sanz-Herrera, and Manuel
  Doblar{\'e}.
\newblock A new reliability-based data-driven approach for noisy experimental
  data with physical constraints.
\newblock {\em Computer Methods in Applied Mechanics and Engineering},
  328:752--774, 2018.

\bibitem{karpatne2017physics}
Anuj Karpatne, William Watkins, Jordan Read, and Vipin Kumar.
\newblock Physics-guided neural networks (pgnn): An application in lake
  temperature modeling.
\newblock {\em arXiv preprint arXiv:1710.11431}, 2017.

\bibitem{lu2019deepxde}
Lu~Lu, Xuhui Meng, Zhiping Mao, and George~E Karniadakis.
\newblock Deepxde: A deep learning library for solving differential equations.
\newblock {\em arXiv preprint arXiv:1907.04502}, 2019.

\bibitem{long2019pde}
Zichao Long, Yiping Lu, and Bin Dong.
\newblock Pde-net 2.0: Learning pdes from data with a numeric-symbolic hybrid
  deep network.
\newblock {\em Journal of Computational Physics}, 399:108925, 2019.

\bibitem{bar2019unsupervised}
Leah Bar and Nir Sochen.
\newblock Unsupervised deep learning algorithm for pde-based forward and
  inverse problems.
\newblock {\em arXiv preprint arXiv:1904.05417}, 2019.

\bibitem{haghighat2020deep}
Ehsan Haghighat, Maziar Raissi, Adrian Moure, Hector Gomez, and Ruben Juanes.
\newblock A deep learning framework for solution and discovery in solid
  mechanics.
\newblock {\em arXiv preprint arXiv:2003.02751}, 2020.

\bibitem{ayensajimenez2020identification}
Jacobo Ayensa-Jiménez, Mohamed~H. Doweidar, Jose~Antonio Sanz-Herrera, and
  Manuel Doblaré.
\newblock Identification of state functions by physically-guided neural
  networks with physically-meaningful internal layers, 2020.

\bibitem{ruas2016}
V.~Ruas.
\newblock {\em Numerical Methods for Partial Differential Equations: An
  Introduction}.
\newblock John Wiley and Sons Ltd., Chichester, West Sussex, United Kingdom,
  2016.

\bibitem{larsson2009}
S.~Larsson and V.~Thomee.
\newblock {\em Partial Differential Equations with Numerical Methods}.
\newblock Springer Verlag, Belin-Heidelberg, 2009.

\bibitem{yoo2015deep}
Hyeon-Joong Yoo.
\newblock Deep convolution neural networks in computer vision: a review.
\newblock {\em IEIE Transactions on Smart Processing \& Computing},
  4(1):35--43, 2015.

\bibitem{rawat2017deep}
Waseem Rawat and Zenghui Wang.
\newblock Deep convolutional neural networks for image classification: A
  comprehensive review.
\newblock {\em Neural computation}, 29(9):2352--2449, 2017.

\bibitem{mccann2017convolutional}
Michael~T McCann, Kyong~Hwan Jin, and Michael Unser.
\newblock Convolutional neural networks for inverse problems in imaging: A
  review.
\newblock {\em IEEE Signal Processing Magazine}, 34(6):85--95, 2017.

\bibitem{anwar2018medical}
Syed~Muhammad Anwar, Muhammad Majid, Adnan Qayyum, Muhammad Awais, Majdi
  Alnowami, and Muhammad~Khurram Khan.
\newblock Medical image analysis using convolutional neural networks: a review.
\newblock {\em Journal of medical systems}, 42(11):226, 2018.

\bibitem{pujari2018}
P.~Pujari, M.~Sewak, and R.~Karim.
\newblock {\em Practical Convolutional Neural Network Models}.
\newblock Packt Publishing, 2018.

\bibitem{bonet2016nonlinear}
Javier Bonet, Antonio~J Gil, and Richard~D Wood.
\newblock {\em Nonlinear solid mechanics for finite element analysis: statics}.
\newblock Cambridge University Press, 2016.

\bibitem{langtangen1999computational}
Hans~Petter Langtangen.
\newblock {\em Computational partial differential equations: numerical methods
  and diffpack programming}, volume~2.
\newblock Springer Berlin, 1999.

\bibitem{zienkiewicz1971finite}
Olgierd~Cecil Zienkiewicz and PB~Morice.
\newblock {\em The finite element method in engineering science}, volume 1977.
\newblock McGraw-Hill London, 1971.

\bibitem{boyd2001chebyshev}
John~P Boyd.
\newblock {\em Chebyshev and Fourier spectral methods}.
\newblock Courier Corporation, 2001.

\bibitem{abadi2016tensorflow}
Mart{\'\i}n Abadi, Ashish Agarwal, Paul Barham, Eugene Brevdo, Zhifeng Chen,
  Craig Citro, Greg~S Corrado, Andy Davis, Jeffrey Dean, Matthieu Devin, et~al.
\newblock Tensorflow: Large-scale machine learning on heterogeneous distributed
  systems.
\newblock {\em arXiv preprint arXiv:1603.04467}, 2016.

\bibitem{cybenko1989approximations}
George Cybenko.
\newblock Approximations by superpositions of a sigmoidal function.
\newblock {\em Mathematics of Control, Signals and Systems}, 2:183--192, 1989.

\bibitem{hornik1991approximation}
Kurt Hornik.
\newblock Approximation capabilities of multilayer feedforward networks.
\newblock {\em Neural networks}, 4(2):251--257, 1991.

\bibitem{lu2017expressive}
Zhou Lu, Hongming Pu, Feicheng Wang, Zhiqiang Hu, and Liwei Wang.
\newblock The expressive power of neural networks: A view from the width.
\newblock In {\em Advances in neural information processing systems}, pages
  6231--6239, 2017.

\bibitem{hanin2017universal}
Boris Hanin.
\newblock Universal function approximation by deep neural nets with bounded
  width and relu activations.
\newblock {\em arXiv preprint arXiv:1708.02691}, 2017.

\bibitem{ros2015nonlocal}
Xavier Ros-Oton.
\newblock Nonlocal elliptic equations in bounded domains: a survey.
\newblock {\em arXiv preprint arXiv:1504.04099}, 2015.

\bibitem{gingold1977smoothed}
Robert~A Gingold and Joseph~J Monaghan.
\newblock Smoothed particle hydrodynamics: theory and application to
  non-spherical stars.
\newblock {\em Monthly notices of the royal astronomical society},
  181(3):375--389, 1977.

\bibitem{nayroles1992generalizing}
B~Nayroles, G~Touzot, and P~Villon.
\newblock Generalizing the finite element method: diffuse approximation and
  diffuse elements.
\newblock {\em Computational mechanics}, 10(5):307--318, 1992.

\bibitem{sukumar1998natural}
Natarajan Sukumar, Brian Moran, and Ted Belytschko.
\newblock The natural element method in solid mechanics.
\newblock {\em International journal for numerical methods in engineering},
  43(5):839--887, 1998.

\bibitem{chinesta2014}
F.~Chinesta, S.~Cescotto, E.~Cueto, and Lorong P.
\newblock {\em Natural Element Method for the Simulation of Structures and
  Processes}.
\newblock John Wiley and Sons Ltd., 2014.

\bibitem{fischer2001filter}
Paul Fischer, Julia Mullen, et~al.
\newblock Filter-based stabilization of spectral element methods.
\newblock {\em Comptes Rendus de l'Academie des Sciences Series I Mathematics},
  332(3):265--270, 2001.

\bibitem{ciaurri2018nonlocal}
{\'O}scar Ciaurri, Luz Roncal, Pablo~Ra{\'u}l Stinga, Jos{\'e}~L Torrea, and
  Juan~Luis Varona.
\newblock Nonlocal discrete diffusion equations and the fractional discrete
  laplacian, regularity and applications.
\newblock {\em Advances in Mathematics}, 330:688--738, 2018.

\bibitem{osserman2013survey}
Robert Osserman.
\newblock {\em A survey of minimal surfaces}.
\newblock Courier Corporation, 2013.

\bibitem{frank2005nonlinear}
Till~Daniel Frank.
\newblock {\em Nonlinear Fokker-Planck equations: fundamentals and
  applications}.
\newblock Springer Science \& Business Media, 2005.

\bibitem{ishizuka2008integral}
R~Ishizuka, S-H Chong, and F~Hirata.
\newblock An integral equation theory for inhomogeneous molecular fluids: The
  reference interaction site model approach.
\newblock {\em The Journal of chemical physics}, 128(3):034504, 2008.

\bibitem{barenblatt1989theory}
Grigory~Isaakovich Barenblatt, Vladimir~Mordukhovich Entov, and
  Viktor~Mikha{\u\i}lovich Ryzhik.
\newblock {\em Theory of fluid flows through natural rocks}.
\newblock Norwell, MA (USA); Kluwer Academic Publishers, 1989.

\bibitem{caffarelli2010nonlinear}
Luis~A Caffarelli and Juan~L Vazquez.
\newblock Nonlinear porous medium flow with fractional potential pressure.
\newblock {\em arXiv preprint arXiv:1001.0410}, 2010.

\bibitem{barles1998option}
Guy Barles and Halil~Mete Soner.
\newblock Option pricing with transaction costs and a nonlinear black-scholes
  equation.
\newblock {\em Finance and Stochastics}, 2(4):369--397, 1998.

\bibitem{ankudinova2008numerical}
Julia Ankudinova and Matthias Ehrhardt.
\newblock On the numerical solution of nonlinear black--scholes equations.
\newblock {\em Computers \& Mathematics with Applications}, 56(3):799--812,
  2008.

\bibitem{kingma2014adam}
Diederik~P Kingma and Jimmy Ba.
\newblock Adam: A method for stochastic optimization.
\newblock {\em arXiv preprint arXiv:1412.6980}, 2014.

\bibitem{levenberg1944method}
Kenneth Levenberg.
\newblock A method for the solution of certain non-linear problems in least
  squares.
\newblock {\em Quarterly of applied mathematics}, 2(2):164--168, 1944.

\bibitem{hou2017convnets}
Le~Hou, Dimitris Samaras, Tahsin~M Kurc, Yi~Gao, and Joel~H Saltz.
\newblock Convnets with smooth adaptive activation functions for regression.
\newblock {\em Proceedings of machine learning research}, 54:430, 2017.

\bibitem{poggio2017and}
Tomaso Poggio, Hrushikesh Mhaskar, Lorenzo Rosasco, Brando Miranda, and Qianli
  Liao.
\newblock Why and when can deep-but not shallow-networks avoid the curse of
  dimensionality: a review.
\newblock {\em International Journal of Automation and Computing},
  14(5):503--519, 2017.

\end{thebibliography}






\end{document}